\documentclass{article}

\usepackage{arxiv}

\usepackage[utf8]{inputenc} 
\usepackage[T1]{fontenc}    
\usepackage{hyperref}       
\usepackage{url}            
\usepackage{booktabs}       
\usepackage{amsfonts}       
\usepackage{nicefrac}       
\usepackage{microtype}      
\usepackage{lipsum}		
\usepackage{graphicx}
\usepackage{doi}

\usepackage{mathtools} 
\usepackage{bbm} 
\usepackage{tikz} 
\usepackage{bigints} 
\usepackage{algorithm}
\usepackage{algorithmic}
\usepackage{dirtree} 
\usepackage{pgfplots}

\usepackage[nameinlink,noabbrev]{cleveref}

\usetikzlibrary{automata, arrows}
\usetikzlibrary{fit,positioning}
\usetikzlibrary{shapes}
\usetikzlibrary{backgrounds}
\usetikzlibrary{bayesnet}
\usetikzlibrary{arrows.meta}
\usetikzlibrary{matrix}

\makeatletter
\def\@seccntformat#1{\@ifundefined{#1@cntformat}%
   {\csname the#1\endcsname\quad}
   {\csname #1@cntformat\endcsname}
}
\makeatother

\title{Hierarchical Blockmodelling for Knowledge Graphs}

\date{} 					

\author{ \href{https://orcid.org/0000-0001-7559-8658}{\includegraphics[scale=0.06]{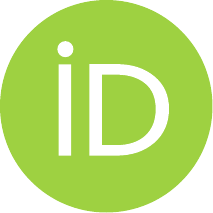}\hspace{1mm}Marcin Pietrasik}\thanks{Corresponding author: \texttt{marcin.pietrasik@maastrichtuniversity.nl} } \\
    \\
    University of Alberta, Edmonton, Canada \\	
    Maastricht University, Maastricht, The Netherlands\\
	\texttt{marcin.pietrasik@maastrichtuniversity.nl} \\
	\And
	\href{https://orcid.org/0000-0003-4783-0717}{\includegraphics[scale=0.06]{orcid.pdf}\hspace{1mm}Marek Reformat} \\
     \\
	University of Alberta, Edmonton Canada\\
	University of Social Sciences, \L{}\'{o}d\'{z}, Poland\\
	\texttt{reformat@ualberta.ca} \\
	 \\
	 \AND
	 \href{https://orcid.org/0000-0002-1989-0301}{\includegraphics[scale=0.06]{orcid.pdf}\hspace{1mm}Anna Wilbik} \\
  \\
	 Maastricht University, Maastricht, The Netherlands \\
	 \texttt{a.wilbik@maastrichtuniversity.nl}\\
}

\renewcommand{\shorttitle}{Hierarchical Blockmodelling for Knowledge Graphs}

\begin{document}
\maketitle

\begin{abstract}
	In this paper, we investigate the use of probabilistic graphical models, specifically stochastic blockmodels, for the purpose of hierarchical entity clustering on knowledge graphs. These models, seldom used in the Semantic Web community, decompose a graph into a set of probability distributions. The parameters of these distributions are then inferred allowing for their subsequent sampling to generate a random graph. In a non-parametric setting, this allows for the induction of hierarchical clusterings without prior constraints on the hierarchy's structure. Specifically, this is achieved by the integration of the Nested Chinese Restaurant Process and the Stick Breaking Process into the generative model. In this regard, we propose a model leveraging such integration and derive a collapsed Gibbs sampling scheme for its inference. To aid in understanding, we describe the steps in this derivation and provide an implementation for the sampler. We evaluate our model on synthetic and real-world datasets and quantitatively compare against benchmark models. We further evaluate our results qualitatively and find that our model is capable of inducing coherent cluster hierarchies in small scale settings. The work presented in this paper provides the first step for the further application of stochastic blockmodels for knowledge graphs on a larger scale. We conclude the paper with potential avenues for future work on more scalable inference schemes.
\end{abstract}


\section{Introduction}\label{sec:intro}

In recent years, using graph structures to model and store data has been garnering an increasing amount of attention among practitioners in sectors ranging from academia to government to industry. Indeed by some measures \cite{dbengines,forbes}, graph database management systems are the fastest growing database type over the past decade. One of the more obvious manifestations of this rise is the recent growth of large scale public graph databases such as DBpedia \cite{lehmann2015dbpedia}, YAGO \cite{pellissier2020yago}, and WikiData \cite{vrandevcic2014wikidata}. The last of these, for instance, contains just over 100 million entities as of 2024, a near seven fold increase over its count in 2014.
The open access to such amounts of graph data has spurred on its use in research related to the Semantic Web, artificial intelligence, and computer science broadly.
One field of research which has received considerable attention is that of mathematically modelling the underlying graph structure that emerges when a knowledge base is populated by information.
The modelling of this structure -- which we refer to as the knowledge graph -- proves useful in its application to solve downstream problems such as link prediction, entity clustering, and hierarchy induction. The last two of these provided the impetus for our work.

Entity clustering refers to the task of grouping together entities in a knowledge graph which share similar properties. The measure by which entities are judged to be similar varies and is one of the key considerations when devising an approach to their clustering. Obtaining an entity clustering allows for the discovery of structures which are implicit in the knowledge graph and provides insight into the number and types of categories which exist in the data. The process operates on unlabelled data and is therefore a type of unsupervised learning. As such, it is one of the first and most useful operations applied to a knowledge graph when performing exploratory analysis.
Another important unsupervised learning task is that of hierarchy induction. The clearest example of a knowledge graph hierarchy is the class taxonomy which organizes a knowledge graph’s classes through superclass-subclass relations. The task of inducing such a taxonomy merely amounts to learning how the classes are organized hierarchically in the knowledge graph.
Similarly, hierarchical clustering of a knowledge graph's entities extends the clustering task described earlier by imposing a hierarchical organization to the clusters themselves. This allows not only to discover which entities are semantically similar as per the clustering but also how entities relate to one another hierarchically.
The motivating factors behind learning knowledge graph hierarchies are various. Perhaps the simplest is that hierarchical structures organize data in a way that is highly intuitive and interpretable to humans. For instance, a hierarchical clustering of knowledge graph entities makes it apparent which entities constitute the broadest concepts in the knowledge graph and how they relate to their descendants. Similarly, a taxonomy of classes reveals implicit relations between entities through its transitive properties. Put plainly, hierarchies induced from knowledge graphs are useful because they are easy to understand.
Indeed, the most widely used knowledge bases -- such as the aforementioned DBpedia, YAGO, and WikiData -- are organized by hierarchical structures, namely trees and directed acyclic graphs. That is to say, these knowledge graphs are hierarchical at their core.
Furthermore, hierarchies are used as components of larger systems to solve common tasks related to knowledge graphs. For instance, hierarchies are used in learning knowledge graph embeddings, both explicitly as an input feature of the model \cite{xie2016representation} and implicitly as a byproduct of the embedding process \cite{zhang2020learning}. As embedding is one of the most common problems in the knowledge graph community, learning accurate hierarchies is therefore desirable.

In this regard, our work proposes a generative model for knowledge graphs which induces a clustering of entities and organizes it hierarchically. Our approach belongs to a class of probabilistic graphical models called stochastic blockmodels. In broad strokes, these models operate by decomposing a knowledge graph into a set of probability distributions which are then sampled from to generate the knowledge graph. As a byproduct of this sampling process, a hierarchical clustering of knowledge graph entities is induced. 
To the best of our knowledge, our approach is the first to apply stochastic blockmodels to knowledge graphs and one of a very few probabilistic graphical models to be used for the purpose of knowledge graph hierarchy induction.
To highlight this, we position our work in the context of existing stochastic blockmodels and hierarchy induction methods in Section \ref{sec:rw} and provide a gentle introduction for their understanding in Section \ref{sec:preliminaries}.
The formal definition of our model that follows in Section \ref{sec:model} results in a joint distribution which is intractable for exact inference. The parameters for our model must therefore be approximated using collapsed Gibbs sampling. To this end, we provide the full derivation of sampling equations as well as the marginalization of collapsed variables. Additional information to supplement Section \ref{sec:model} may be found in Appendices \ref{adx:a} through \ref{adx:level_mixtures}.
Section \ref{sec:conclusion} concludes the paper by summarizing its contributions and providing avenues for future work.

\section{Related Work}\label{sec:rw}

Our proposed model lies at the intersection of two areas in artificial intelligence which deal with modelling graph data: stochastic blockmodelling and hierarchy induction. Due to the limited overlap of these fields, we provide separate summaries of related works for each.

\subsection{Stochastic Blockmodels} \label{subsec:rw_blockmodels}

Stochastic blockmodels are a class of probabilistic graphical models used for generating random graphs with roots in the fields of social science and mathematics. First proposed in 1983 by Holland et al. \cite{holland1983stochastic} for modelling social networks, they have expanded their utility to fields such as biochemistry \cite{wang2022ppisb}, education \cite{sweet2019modeling}, and artificial intelligence \cite{airoldi2008mixed,ho2011multiscale,zhang2022hierarchical} among others. 
In simplest terms, stochastic blockmodels are a type of Bayesian non-parametric graph partition model in that their approach relies on grouping graph entities together via partitions -- often referred to as blocks -- which share similar structural properties.
The generative process by which this partitioning occurs is realized by sampling from a set of probability distributions, giving rise to the stochasticity of stochastic blockmodels. The learning process is then infer the parameters of these distributions using a Bayesian inference scheme. 
We provide a technical introduction to stochastic blockmodels in the subsequent section.

The seminal work in this area is the Stochastic Blockmodel \cite{nowicki2001estimation} which partitions entities into a fixed number of communities and models the interactions between them as those of their communities. Community relations are modelled via a community relations matrix which assigns a degree to all pairwise interactions between the communities in the model. This idea was extended to the infinite case allowing for an a priori unspecified number of communities via the Chinese restaurant process \cite{aldous1985exchangeability} in the Infinite Relational Model \cite{kemp2006learning} and its recent hierarchical counterpart the Hierarchical Infinite Relational Model \cite{saad2021hierarchical}. 
A variant which relaxes the notion of community membership to allow for entities belonging to multiple communities is the aptly named Mixed Membership Stochastic Blockmodel \cite{airoldi2008mixed}. By allowing for mixed membership, the model is better able to capture entities whose belonging to a community is not crisp. For instance, the belonging of tomatoes to the community of fruits is not perfect since it can be considered a vegetable in certain contexts such as in cooking. 
This idea was generalized to the infinite case in the Dynamic Infinite Mixed Membership Stochastic Blockmodel \cite{fan2014dynamic} and the hierarchical case in the Multiscale Community Blockmodel \cite{ho2011multiscale}. The latter of these two is closely related to our model and receives more attention later in the paper. All of the aforementioned models, however, operate on graphs wherein entities are related to one another through the same type of edge, making them unsuitable for application to knowledge graphs without modification.

The underlying structure of a knowledge graph is that of a multilayer graph wherein entities interact with one another through different types of relations, represented as different types of edges in the graph.
These relations may be thought of as separate layers of graphs which share the same entities. 
Multilayer graphs have also received considerable attention in stochastic blockmodelling.
Perhaps the simplest approach is to aggregate the layers in the multilayer graph to a single layer before applying a conventional blockmodelling approach as was done in Berlingerio et al. \cite{berlingerio2011finding}. A closely related approach is to model each layer in the graph independently as done in Barigozzi et al. \cite{barigozzi2011identifying} and aggregate the results afterwards.
These approaches offer limited success as they don't capture the interlayer dependencies in the multilayer graph and treat each layer as equally valuable in its content during modelling, as pointed out by Paul and Chen \cite{paul2016consistent}. To remedy this, the authors propose a multilayer extension of the aforementioned Stochastic Block Model, aptly named the Multi-Layer Stochastic Blockmodel, which modifies the original community relations matrix to a community relations tensor to account for graph multilayeredness.
Analogously, a multilayer extension for the Mixed Membership Stochastic Blockmodel was proposed by De Bacco et al. \cite{de2017community}.
Finally, the Multilayer Neural Blockmodel \cite{pietrasik2021neural} was proposed recently as a way to marry neural networks with the probabilistic approach of stochastic blockmodels for modelling multilayer graphs.
A comprehensive review of stochastic blockmodels and their applications is provided by Lee and Wilkinson \cite{lee2019review}.

\subsection{Hierarchy Induction Models}\label{subsec:rw_hierarchy_induction}

In the context of our work, hierarchy induction refers to the discovery of hierarchical structures which are implicit and otherwise unexpressed in a knowledge graph.
One concrete way this task is formulated is as that of learning subsumption axioms for classes in a knowledge graph, thereby discovering a hierarchical organization of a knowledge graph's entities.
To this end, Statistical Schema Induction \cite{volker2011statistical} uses association rule mining on a knowledge graph's transaction table to generate subsumption axioms with support and confidence values which are then used as the basis for a greedy algorithm for constructing an ontology.
SMICT \cite{pietrasik2020simple} transforms a knowledge graph into a tuple structure wherein entities are annotated by tags and applies a greedy algorithm to learn a taxonomy of classes. This method was extended to perform hierarchical clustering using the Jaccard coefficient \cite{pietrasik2021path}.
In general, by transforming a knowledge graph to a tuple structure, various \cite{heymann2006collaborative, schmitz2006inducing, wang2018hybrid} methods in the area of tag hierarchy induction can be leveraged.
In a related approach, Chen and Reformat \cite{chen2014learning} derive a similarity matrix from a knowledge graph's tuple structure which serves as the clustering metric for hierarchical agglomerative clustering.
Mohamed \cite{mohamed2019unsupervised} takes a similar approach wherein subjects which are described by the same tag pairs are assigned to the same groups. The similarity between these groups is then calculated to construct a hierarchy.
In a method which bears similarity to our own, Zhang et al. \cite{zhang2022hierarchical} use a non-parametric Bayesian approach to induce a hierarchy of topic communities. Despite a similar statistic framework and inference scheme, the hierarchy induced by this work differs significantly from our own. For instance, relations between communities are not modelled and entities are never explicitly assigned to communities.
Along similar lines is GMMSchema \cite{bonifati2022hierarchical} which uses a Gaussian mixture model to generate a schema graph which can be viewed as a hierarchical abstraction of the original knowledge graph.

Another common approach to learning hierarchies from knowledge graphs is via an intermediate representation which lends itself well to existing hierarchy induction methods. To this end, knowledge graph embedding is oftentimes leveraged. This process involves learning a mapping from the discrete knowledge graph to a continuous vector space. The vector representation may then serve as the input to machine and deep learning methods for hierarchy learning.
Translation based methods such as the seminal TransE \cite{bordes2013translating} and its extensions \cite{wang2014knowledge, lin2015learning, ji2015knowledge} treat relations in a knowledge graph as translations between entities. Additive in nature, they operate on the intuition that embeddings of subjects and objects should be proximal when translated by the relation of a valid triple. These embeddings are learned by minimizing an objective function using an optimization method such as stochastic gradient descent.
Bilinear methods \cite{nickel2011three, yang2014embedding, kazemi2018simple, balavzevic2019tucker} operate on the binary adjacency tensor of the knowledge graph and factorize entities and relations into vectors and matrices. Triples are then modelled as their resulting product. These methods tend to perform well on measures of performance compared to translation based methods but suffer from higher training complexity. 
Deep learning models have also been proposed in the context of knowledge graph embeddings. For instance, the Relational Graph Convolution Network \cite{schlichtkrull2018modeling} leverages graph convolutions to learn neighbourhood information of entities, thereby explicitly incorporating structural information into its modelling. Another widely used deep approach, ConvE \cite{dettmers2018convolutional}, stacks subject and predicate embeddings as a matrix and convolves over them in two dimensions using a neural framework. This approach was extended in ConvKB \cite{nguyen2017novel} which incorpoates objects into the convolution process and CapsE \cite{vu2019capsule} which uses a similar architecture with capsule layers to yield scores for triples.
A recent and comprehensive comparative analysis of various embedding methods may be found in \cite{rossi2021knowledge}.

Having obtained an embedded representation of a knowledge graph, hierarchical clustering methods can be applied to induce a hierarchy.
For instance RESCAL \cite{nickel2011three}, a bilinear embedding method, was used in conjunction with OPTICS \cite{ankerst1999optics}, a density based hierarchical clustering algorithm, in Nickel et al. \cite{nickel2012factorizing} to obtain a hierarchical clustering of entities. They found that such an approach achieve more coherent results for concepts which appear at the top of the hierarchy, largely due to data sparsity for descendant concepts.
Along similar lines, TIEmb \cite{ristoski2017large} generates embeddings using RDF2Vec \cite{ristoski2016rdf2vec}, an embedding method based on the skip-gram language model \cite{mikolov2013efficient}, before learning a hierarchical structure based on the proximities of class centroids in the embedded space. The same embedding approach was used in Martel and Zouaq \cite{martel2021taxonomy} wherein the embeddings were then clustered using hierarchical agglomerative clustering and assigned types.
This type of clustering was used in the field of cybersecurity in Ding et al. \cite{ding2021method} wherein a bag-of-words representation of a knowledge graph served as input.

\section{Preliminaries}\label{sec:preliminaries}

Before describing the details of our proposed model, we provide a basic overview of several concepts necessary for its understanding. 
These concepts are described only insofar as to provide readers with the foundation on which the explanation of our model can be built. We implore readers unfamiliar with knowledge graphs or Bayesian nonparametrics to follow the relevant citations provided in each of the subsequent subsections.
To aid in readability we use the following conventions in our notation: lowercase italic Latin letters for iterators and indexers; uppercase italic Latin letters for scalar variables; lowercase boldface Latin letters for vectors; uppercase boldface Latin letters for matrices and tensors; uppercase stylized Latin letters for sets; lowercase Greek letters for hyperparameters; and uppercase Greek letters for functions.

\subsection{Knowledge Graphs}\label{subsec:kgs}

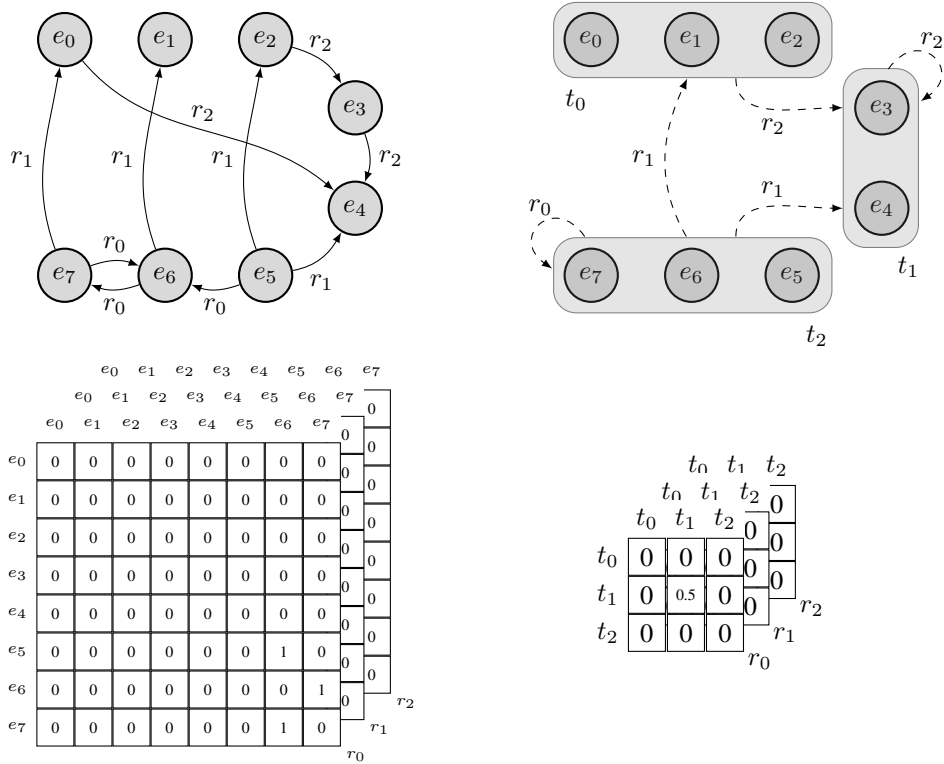
\begin{figure}
\centering
\begin{tikzpicture}[
      base/.style = {draw, align=center},
  entity/.style  ={base,minimum width=2em, shape=circle,thick, black,fill=gray!30,},
  auto matrix/.style={matrix of nodes,
  nodes in empty cells,
  cells={nodes={minimum width=1.4em,minimum height=1.4em,
   draw,anchor=center,fill=white
  }}}]
\node (entity7) [entity] {$e_7$};
\node (entity6) [entity,  right= 0.6cm of entity7] {$e_6$};
\draw [-latex] (entity7) to [out=20,in=160] node[midway, above] {$r_0$} (entity6);
\draw [-latex] (entity6) to [out=200,in=340] node[midway, below] {$r_0$} (entity7);
\node (entity5) [entity,  right= 0.6cm of entity6] {$e_5$};
\draw [-latex] (entity5) to [out=200,in=340] node[midway, below] {$r_0$} (entity6);
\node (entity0) [entity,  above= 2.4cm of entity7] {$e_0$};
\draw [-latex] (entity7) to [out=110,in=260] node[midway, left] {$r_1$} (entity0);
\node (entity1) [entity,  right= 0.6cm of entity0] {$e_1$};
\draw [-latex] (entity6) to [out=110,in=260] node[midway, left] {$r_1$} (entity1);
\node (entity2) [entity,  right= 0.6cm of entity1] {$e_2$};
\draw [-latex] (entity5) to [out=110,in=260] node[midway, left] {$r_1$} (entity2);
\node (dummy1) [above= 0.4cm of entity5] {};
\node (entity4) [entity,  right= 0.7cm of dummy1] {$e_4$};
\node (entity3) [entity,  above= 0.6cm of entity4] {$e_3$};
\draw [-latex] (entity0) to [out=310,in=140] node[midway, above] {$r_2$} (entity4);
\draw [-latex] (entity2) to [out=344,in=120] node[midway, above] {$r_2$} (entity3);
\draw [-latex] (entity5) to [out=10,in=240] node[midway, below = 0.1cm] {$r_1$} (entity4);
\draw [-latex] (entity3) to [out=290,in=70] node[midway, right] {$r_2$} (entity4);
 \matrix[auto matrix=z,xshift=2em,yshift=2em](matz) at (1.4,-4){
  |[draw=none]| & |[draw=none]| \tiny{$e_0$} & |[draw=none]| \tiny{$e_1$} & |[draw=none]| \tiny{$e_2$} & |[draw=none]| \tiny{$e_3$}& |[draw=none]| \tiny{$e_4$} & |[draw=none]| \tiny{$e_5$}& |[draw=none]| \tiny{$e_6$} & |[draw=none]| \tiny{$e_7$} \\
  |[draw=none]| & |[draw=none]| & |[draw=none]| & |[draw=none]| & |[draw=none]| & |[draw=none]| & |[draw=none]| & |[draw=none]| & \tiny{0}   \\
  |[draw=none]|  & |[draw=none]| & |[draw=none]| & |[draw=none]| & |[draw=none]| & |[draw=none]| & |[draw=none]| & |[draw=none]| & \tiny{0}  \\
  |[draw=none]|  & |[draw=none]| & |[draw=none]| & |[draw=none]| & |[draw=none]| & |[draw=none]| & |[draw=none]| & |[draw=none]| & \tiny{0}   \\
  |[draw=none]|  & |[draw=none]| & |[draw=none]| & |[draw=none]| & |[draw=none]| & |[draw=none]| & |[draw=none]| & |[draw=none]| & \tiny{0}  \\
  |[draw=none]|  & |[draw=none]| & |[draw=none]| & |[draw=none]| & |[draw=none]| & |[draw=none]| & |[draw=none]| & |[draw=none]| & \tiny{0}  \\
  |[draw=none]|   & |[draw=none]| & |[draw=none]| & |[draw=none]| & |[draw=none]| & |[draw=none]| & |[draw=none]| & |[draw=none]| & \tiny{0}   \\
  |[draw=none]|  & |[draw=none]| & |[draw=none]| & |[draw=none]| & |[draw=none]| & |[draw=none]| & |[draw=none]| & |[draw=none]| & \tiny{0} \\
  |[draw=none]| & |[draw=none]| & |[draw=none]| & |[draw=none]| & |[draw=none]| & |[draw=none]| & |[draw=none]| & |[draw=none]| & \tiny{0}  \\
 };
 \matrix[auto matrix=y,xshift=1em,yshift=1em] at (1.4,-4) (maty){
  |[draw=none]| & |[draw=none]| \tiny{$e_0$} & |[draw=none]| \tiny{$e_1$} & |[draw=none]| \tiny{$e_2$} & |[draw=none]| \tiny{$e_3$}& |[draw=none]| \tiny{$e_4$} & |[draw=none]| \tiny{$e_5$}& |[draw=none]| \tiny{$e_6$} & |[draw=none]| \tiny{$e_7$} \\
  |[draw=none]| & |[draw=none]| & |[draw=none]| & |[draw=none]| & |[draw=none]| & |[draw=none]| & |[draw=none]| & |[draw=none]| & \tiny{0}   \\
  |[draw=none]|  & |[draw=none]| & |[draw=none]| & |[draw=none]| & |[draw=none]| & |[draw=none]| & |[draw=none]| & |[draw=none]| & \tiny{0}  \\
  |[draw=none]|  & |[draw=none]| & |[draw=none]| & |[draw=none]| & |[draw=none]| & |[draw=none]| & |[draw=none]| & |[draw=none]| & \tiny{0}   \\
  |[draw=none]|  & |[draw=none]| & |[draw=none]| & |[draw=none]| & |[draw=none]| & |[draw=none]| & |[draw=none]| & |[draw=none]| & \tiny{0}  \\
  |[draw=none]|  & |[draw=none]| & |[draw=none]| & |[draw=none]| & |[draw=none]| & |[draw=none]| & |[draw=none]| & |[draw=none]| & \tiny{0}  \\
  |[draw=none]|   & |[draw=none]| & |[draw=none]| & |[draw=none]| & |[draw=none]| & |[draw=none]| & |[draw=none]| & |[draw=none]| & \tiny{0}   \\
  |[draw=none]|  & |[draw=none]| & |[draw=none]| & |[draw=none]| & |[draw=none]| & |[draw=none]| & |[draw=none]| & |[draw=none]| & \tiny{0} \\
  |[draw=none]| & |[draw=none]| & |[draw=none]| & |[draw=none]| & |[draw=none]| & |[draw=none]| & |[draw=none]| & |[draw=none]| & \tiny{0}  \\
 };
 \matrix[auto matrix=x] at (1.4,-4) (matx){
  |[draw=none]| & |[draw=none]| \tiny{$e_0$} & |[draw=none]| \tiny{$e_1$} & |[draw=none]| \tiny{$e_2$} & |[draw=none]| \tiny{$e_3$}& |[draw=none]| \tiny{$e_4$} & |[draw=none]| \tiny{$e_5$}& |[draw=none]| \tiny{$e_6$} & |[draw=none]| \tiny{$e_7$} \\
  |[draw=none]|  \tiny{$e_0$} & \tiny{0} & \tiny{0} & \tiny{0} & \tiny{0} & \tiny{0} & \tiny{0} & \tiny{0} & \tiny{0}   \\
  |[draw=none]|  \tiny{$e_1$} & \tiny{0} & \tiny{0} & \tiny{0} & \tiny{0} & \tiny{0} & \tiny{0} & \tiny{0} & \tiny{0}  \\
  |[draw=none]|  \tiny{$e_2$} & \tiny{0} & \tiny{0} & \tiny{0} & \tiny{0} & \tiny{0} & \tiny{0} & \tiny{0} & \tiny{0}   \\
  |[draw=none]|  \tiny{$e_3$} & \tiny{0} & \tiny{0} & \tiny{0} & \tiny{0} & \tiny{0} & \tiny{0} & \tiny{0} & \tiny{0}  \\
  |[draw=none]|  \tiny{$e_4$} & \tiny{0} & \tiny{0} & \tiny{0} & \tiny{0} & \tiny{0} & \tiny{0} & \tiny{0} & \tiny{0}  \\
  |[draw=none]|  \tiny{$e_5$} & \tiny{0} & \tiny{0} & \tiny{0} & \tiny{0} & \tiny{0} & \tiny{0} & \tiny{1} & \tiny{0}   \\
  |[draw=none]|  \tiny{$e_6$} & \tiny{0} & \tiny{0} & \tiny{0} & \tiny{0} & \tiny{0} & \tiny{0} & \tiny{0} & \tiny{1} \\
  |[draw=none]|  \tiny{$e_7$} & \tiny{0} & \tiny{0} & \tiny{0} & \tiny{0} & \tiny{0} & \tiny{0} & \tiny{1} & \tiny{0}  \\
 };
 \node[] at ([xshift=0.5ex]matz.south east) {\tiny{$r_2$}};
 \node[] at ([xshift=0.5ex]maty.south east) {\tiny{$r_1$}};
 \node[] at ([xshift=0.5ex]matx.south east) {\tiny{$r_0$}};
 \node (entity7_1) at (7,0) [entity] {$e_7$};
 \node (entity6_1) [entity,  right= 0.6cm of entity7_1] {$e_6$};
\node (entity5_1) [entity,  right= 0.6cm of entity6_1] {$e_5$};
\node (entity0_1) [entity,  above= 2.4cm of entity7_1] {$e_0$};
\node (entity1_1) [entity,  right= 0.6cm of entity0_1] {$e_1$};
\node (entity2_1) [entity,  right= 0.6cm of entity1_1] {$e_2$};
\node (dummy1_1) [above= 0.4cm of entity5_1] {};
\node (entity4_1) [entity,  right= 0.7cm of dummy1_1] {$e_4$};
\node (entity3_1) [entity,  above= 0.6cm of entity4_1] {$e_3$};
\draw [rounded corners=3mm,solid,gray, fill=gray,fill opacity=0.2] (6.5,0.5)--(10.2,0.5)--(10.2,-0.5)--(6.5,-0.5)--cycle;
\draw [rounded corners=3mm,solid,gray, fill=gray,fill opacity=0.2] (6.5,3.63)--(10.2,3.63)--(10.2,2.63)--(6.5,2.63)--cycle;
\draw [rounded corners=3mm,solid,gray, fill=gray,fill opacity=0.2] (10.35,2.75)--(11.35,2.75)--(11.35,0.37)--(10.35,0.37)--cycle;
\node (dummy2_1) [above= -0.1cm of entity6_1] {};
\node (dummy3_1) [below= -0.1cm of entity1_1] {};
\draw [-latex, dashed] (dummy2_1) to [out=120,in=245] node[midway, left] {$r_1$} (dummy3_1);
\node (dummy4_1) [above= -0.1cm of entity7_1] {};
\node (dummy5_1) [left= -0.1cm of entity7_1] {};
\draw [-latex, dashed] (dummy4_1) to [out=90,in=180, bend right=100, looseness=3] node[midway, above] {$r_0$} (dummy5_1);
\node (dummy6_1) [below= -0.1cm of entity2_1] {};
\node (dummy8_1) [left= 0.5cm of dummy6_1] {};
\node (dummy7_1) [left= -0.1cm of entity3_1] {};
\draw [-latex, dashed] (dummy8_1) to [out=270,in=180] node[midway, below] {$r_2$} (dummy7_1);
\node (dummy9_1) [above= -0.1cm of entity5_1] {};
\node (dummy10_1) [left= 0.5cm of dummy9_1] {};
\node (dummy11_1) [left= -0.1cm of entity4_1] {};
\draw [-latex, dashed] (dummy10_1) to [out=90,in=180] node[midway, above] {$r_1$} (dummy11_1);.
\node (dummy12_1) [above= -0.1cm of entity3_1] {};
\node (dummy13_1) [right= -0.1cm of entity3_1] {};
\draw [-latex, dashed] (dummy12_1) to [out=90,in=0, bend left=100, looseness=3] node[midway, above] {$r_2$} (dummy13_1);
\node[] at (6.8,2.33) {{$t_0$}};
\node[] at (10,-0.8) {{$t_2$}};
\node[] at (11.22,0.11) {{$t_1$}};
 \matrix[auto matrix=z,xshift=2em,yshift=2em](matc) at (8,-4){
  |[draw=none]| & |[draw=none]| {$t_0$} & |[draw=none]| {$t_1$} & |[draw=none]| {$t_2$}  \\
  |[draw=none]|  & |[draw=none]| & |[draw=none]| & 0 \\
  |[draw=none]|   &  &  & 0 \\
  |[draw=none]|  &  &  & 0 \\
 };
 \matrix[auto matrix=y,xshift=1em,yshift=1em] at (8,-4) (matb){
  |[draw=none]| & |[draw=none]| {$t_0$} & |[draw=none]| {$t_1$} & |[draw=none]| {$t_2$}  \\
  |[draw=none]|  & |[draw=none]| & |[draw=none]| & 0 \\
  |[draw=none]|   & 0 & 0 & 0 \\
  |[draw=none]|  & 0 & 0 & 0 \\
 };
 \matrix[auto matrix=x] at (8,-4) (mata){
  |[draw=none]| & |[draw=none]| {$t_0$} & |[draw=none]| {$t_1$} & |[draw=none]| {$t_2$}  \\
  |[draw=none]|  {$t_0$} & 0 & 0 & 0 \\
  |[draw=none]|  {$t_1$} & 0 & \tiny{0.5} & 0 \\
  |[draw=none]|  {$t_2$} & 0 & 0 & 0 \\
 };
 \node[] at ([xshift=0.5ex]matc.south east) {$r_2$};
 \node[] at ([xshift=0.5ex]matb.south east) {$r_1$};
 \node[] at ([xshift=0.5ex]mata.south east) {$r_0$};
\end{tikzpicture}
\caption{Toy example of a knowledge graph and how it may be modelled by a stochastic blockmodel. Starting from top left quadrant and proceeding clockwise: graphical representation of a knowledge graph with entities $e_0$ through $e_7$ and predicates $r_0$ through $r_2$; graphical representation of aforementioned knowledge graph as modelled by a stochastic blockmodel with communities $t_0$ through $t_2$; potential community relations tensor induced by stochastic blockmodel; adjacency tensor of knowledge graph above it.}
\label{fig:toy_kg_and_sb}
\end{figure}

We refer to Hogan et al. \cite{hogan2021knowledge} for their definition of knowledge graphs as \textit{``a graph of data intended to accumulate and convey knowledge of the real world, whose nodes represent entities of interest and whose edges represent potentially different relations between these entities.''}
Concretely, information is stored as a collection of triples wherein each triple relates a subject entity, $e_i$, to an object entity, $e_j$, via a predicate, $p_r$. 
Formally, we define a knowledge graph, $\mathcal{G}$, as a set such that $\mathcal{G} = \{\langle e_i, r_p, e_j \rangle \in \mathcal{E} \times \mathcal{R} \times \mathcal{E} \}$ where $\langle e_i, r_p, e_j \rangle$ is a triple, $\mathcal{E}$ is the set of entities in $\mathcal{G}$, and $\mathcal{R}$ is the set of predicates in $\mathcal{G}$. When put together, the triples form a directed graph with nodes corresponding to entities and edges corresponding to predicates.
Each triple in a knowledge graph describes one piece of information or fact. For instance, $\langle$\texttt{Henry Ford, }\texttt{occupation, }\texttt{Engineer}$\rangle$ relates the subject \texttt{Henry Ford} to the object \texttt{Engineer} through the predicate \texttt{occupation} and states, in plain English, that Henry Ford's occupation is an engineer. 
Notice that this definition of knowledge graphs allows for cycles and entity self-relations to exist.
This is made clear when analyzing a knowledge graph's binary adjacency tensor which may be asymmetric and containing non-zero values in its main diagonal.
Knowledge graphs are oftentimes represented in their tensor form as it allows for easier numerical operation and thus opens the door to various tools and methods in artificial intelligence.
A binary adjacency tensor is obtained from a knowledge graph by ordering its entities and predicates along an $|\mathcal{E}| \times |\mathcal{E}| \times |\mathcal{R}|$ tensor, $\mathbf{G}$, that takes on values $g_{ijr} = 1$ if there exists a triple in $\mathcal{G}$ from entity $e_i$ to entity $e_j$ on predicate $r_p$ and $g_{ijr} = 0$ otherwise. 
This representation is used in stochastic blockmodelling and is the one we will use in this paper henceforth. The left half of Figure \ref{fig:toy_kg_and_sb} depicts a simple knowledge graph along with its adjacency tensor representation.
A comprehensive introduction to knowledge graphs is provided by Gutierrez and Sequeda \cite{gutierrez2021knowledge}.

\subsection{Stochastic Blockmodels}\label{subsec:blockmodelling}

Stochastic blockmodels are a heterogeneous collection of generative models united in their adoption of two characteristics: stochasticity in the generative process and the partitioning of nodes into communities. Describing them by referring to a concrete instance is thus bound to include definitions which do not apply to all members of the class. With this in mind, our introduction to stochastic blockmodels draws on their key characteristics to motivate a toy stochastic blockmodel for generating a knowledge graph.
All stochastic blockmodels are defined by a set of probability distributions from which samples are obtained to generate the adjacency tensor of the knowledge graph, $\mathbf{G}$. 
In order to perform this generation, the knowledge graph's entities must first be assigned to one of the model's communities. This is done by sampling the model's variables responsible for this assignment.
Let $\mathbf{A}$ be a tensor representing these variables with a corresponding hyperparameter $\alpha$ responsible for parameterizing their prior distribution.
In stochastic blockmodels, the probability of an interaction between two entities is modelled as the degree of interaction between their respective communities. It is necessary, therefore, to capture these community relations by sampling their corresponding model variables.
Let $\mathbf{B}$ be a tensor representing this subset of variables with a prior hyperparameter $\beta$.
The joint distribution of this model is obtained by applying the chain rule of probability as follows:
\begin{align}
    \mathbbm{P}(\mathbf{G},\mathbf{A}, \mathbf{B} \ | \ \alpha, \beta) = \prod_{g_{ijr} \in \mathbf{G}} \mathbbm{P}(g_{ijr}|\mathbf{A}_{ijr}, \mathbf{B}_{ijr}, \alpha, \beta) \mathbbm{P}(\mathbf{B}_{ijr} \ | \ \mathbf{A}_{ijr},  \beta) \mathbbm{P}(\mathbf{A}_{ijr} \ | \ \alpha) \label{eqn:toy_joint}
\end{align}
where $\mathbf{A}_{ijr}$ and $\mathbf{B}_{ijr}$ indicate the latent variables in $\mathbf{A}$ and $\mathbf{B}$ associated with sampling $g_{ijr}$. 
Notice that the probability of drawing a value in the knowledge graph's adjacency tensor, $\mathbbm{P}(g_{ijr}|\mathbf{A}_{ijr}, \mathbf{B}_{ijr}, \alpha, \beta)$, is conditioned on $\mathbf{A}$ and $\mathbf{B}$. Thus, in order to generate the knowledge graph, it's necessary to first infer the values of $\mathbf{A}$ and $\mathbf{B}$. This inference process is analogous to the training phase of other machine and deep learning models.
In most cases, the solution is intractable for exact inference and must be approximated using an inference scheme.
Perhaps the simplest inference scheme used in stochastic blockmodelling is Gibbs sampling, a Markov chain Monte Carlo method which can be used for sampling from a joint distribution.
Gibbs sampling approximates this distribution by iteratively sampling from its variables' full conditional distributions. This iterative sampling creates a Markov chain of samples wherein its stationary distribution approximates the joint distribution of the model. 
Continuing the example above, to infer the blockmodel's parameters for $g_{ijr}$, namely $\mathbf{A}_{ijr}$ and $\mathbf{B}_{ijr}$, inference is performed on their conditional distributions $\mathbbm{P}(\mathbf{A}_{ijr} \ | \ \mathbf{G, \mathbf{B}}, \alpha)$ and $\mathbbm{P}(\mathbf{B}_{ijr} \ | \ \mathbf{G}, \mathbf{A}, \beta)$, respectively.  
We apply Bayes' theorem to obtain these distributions. Recall that by this theorem the posterior distribution is proportional to the product of the likelihood and the prior. We can therefore express the conditionals of $\mathbf{A}_{ijr}$ and $\mathbf{B}_{ijr}$ as follows:
\begin{align}
    \mathbbm{P}(\mathbf{A}_{ijr} \ | \ \mathbf{G, \mathbf{B}}, \alpha) \propto \mathbbm{P}(\mathbf{G} \ | \ \mathbf{A}_{ijr}, \mathbf{B}) \mathbbm{P}(\mathbf{A} \ | \ \alpha) \label{eqn:posterior_a}  \\
    \mathbbm{P}(\mathbf{B}_{ijr} \ | \ \mathbf{G}, \mathbf{A}, \beta) \propto \mathbbm{P}(\mathbf{G} \ | \ \mathbf{B}_{ijr}, \mathbf{A}) \mathbbm{P}(\mathbf{B} \ | \ \beta) \label{eqn:posterior_b} 
\end{align}
Where $\mathbbm{P}(\mathbf{G} \ | \ \mathbf{A}, \mathbf{B})$ and $\mathbbm{P}(\mathbf{G} \ | \ \mathbf{B}, \mathbf{A})$ are the likelihoods, and $\mathbbm{P}(\mathbf{A}_{ijr} \ | \ \alpha)$ and $\mathbbm{P}(\mathbf{B}_{ijr} \ | \ \beta)$ are the priors of $\mathbf{A}_{ijr}$ and $\mathbf{B}_{ijr}$, respectively.
The likelihood may be understood as the chance observing the data given the model parameters. In Equations \ref{eqn:posterior_a} and \ref{eqn:posterior_b}, it is the likelihood of drawing $\mathbf{G}$ from our model with parameters $\mathbf{A}$ and $\mathbf{B}$.
The prior represents the assumptions about a variable before any data is taken into account. They are oftentimes chosen in order to leverage a conjugacy with their dependant variables. Priors are parameterized by hyperparameters which must be specified a priori. The choice of these hyperparameters influences the density of the prior and can thus change the output of the model.
Gibbs sampling draws from the variables' full conditional distributions iteratively for a predetermined number of iterations, $iters$. To highlight this, the superscript $iter$ is added to denote the value of a variable at the corresponding iteration. The Gibbs sampling process may be summarized as follows:
\begin{enumerate}
    \item Initialize $\mathbf{A}_{ijr}^0$ and $\mathbf{B}_{ijr}^0$ for each $g_{ijr} \in \mathbf{G}$ \label{algo:toy_gibbs_init}
    \item For iteration $iter$ in $1, 2, ..., iters$ \label{algo:toy_gibbs_loop}
    \begin{enumerate}
        \item Sample $\mathbf{A}_{ijr}^{iter} \sim \mathbbm{P}(\mathbf{A}_{ijr}^{iter} \ | \ \mathbf{G}, \mathbf{B}^{iter-1}, \alpha)$ for each $g_{ijr} \in \mathbf{G}$ using Equation \ref{eqn:posterior_a}
        \item Sample $\mathbf{B}_{ijr}^{iter} \sim \mathbbm{P}(\mathbf{B}_{ijr}^{iter} \ | \ \mathbf{G}, \mathbf{A}^{iter-1}, \alpha)$ for each $g_{ijr} \in \mathbf{G}$ using Equation \ref{eqn:posterior_b}
    \end{enumerate}
\end{enumerate}
In step \ref{algo:toy_gibbs_init}, variables can be initialized by sampling from their prior distributions or specified explicitly if a priori evidence to suggest their true values exists.
Step \ref{algo:toy_gibbs_loop} depicts the iterative sampling of model variables from their full conditionals. We note that samples obtained early in this process may be drawn from a distribution distant to that of the desired stationary distribution. As such it is necessary to discard the samples obtained before this distribution has been reached. This process is commonly referred to as burning in the Gibbs sampler and the number of discarded iterations as the burn in iterations. Furthermore, as successive samples in this process are autocorrelated, there may be a lag period applied in obtaining results such that samples in during the lag period are also discarded. Thus, if our toy example performs 1000 iterations with a burn in of 900 and a lag of 10, only 9 samples will be obtained as the output of the Gibbs sampler. These 9 samples are then aggregated over to account for the stochasticity in sampling from the posterior and arrive at a final result. The process by which these samples are aggregated are model specific and may be as simple as merely taking the sampled mode. An introduction to Gibbs sampling and related sampling schemes is covered by Mackay \cite{mackay2003information} and a thorough discussion of stochastic blockmodels along with their concrete examples is provided by Abbe \cite{abbe2017community}.

\subsection{The Chinese Restaurant Process}\label{subsec:crp}

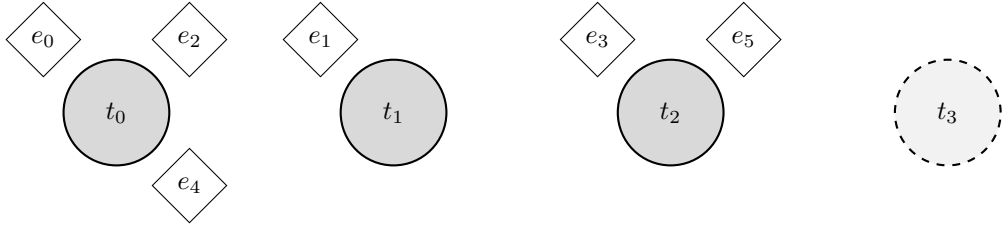
\begin{figure}
\centering
\begin{tikzpicture}[
      base/.style = {draw, align=center},
  patron/.style = {base, diamond, fill=white!30, minimum width=5mm, minimum height=5mm},
  every edge quotes/.style = {auto=right},
  circlenode/.style  ={base,minimum width=4em, shape=circle,thick, black,fill=gray!30,}]
\node (table0) [circlenode] {$t_0$};
\node (patron0) [patron, above left= 0.3cm of table0] {$e_0$};
\node (patron2) [patron, above right= 0.3cm of table0] {$e_2$};
\node (patron4) [patron, below right= 0.3cm of table0] {$e_4$};
\node (table1) [circlenode, right= 2.25cm of table0] {$t_1$};
\node (patron1) [patron, above left= 0.3cm of table1] {$e_1$};
\node (table2) [circlenode, right= 2.25cm of table1] {$t_2$};
\node (patron3) [patron, above left= 0.3cm of table2] {$e_3$};
\node (patron5) [patron, above right= 0.3cm of table2] {$e_5$};
\node (table3) [circlenode, right= 2.25cm of table2, fill=gray!10, dashed] {$t_3$};
\end{tikzpicture}
\caption{Toy example of the CRP after sitting patrons $e_0$ through $e_5$. Tables $t_0$ through $t_2$ are occupied and table $t_3$ is the next unoccupied table. We illustrate Equation \ref{eqn:crp} by calculating the probabilities of sitting patron $p_6$ at tables $t_0$ and $t_3$: $\mathbbm{P}(e_6 = t_0) = \frac{3}{6 + \gamma}$ and $\mathbbm{P}(e_6 = t_2) = \frac{\gamma}{6 + \gamma}$.}
\label{fig:crp}
\end{figure}

The Chinese restaurant process (CRP) \cite{aldous1985exchangeability} is a discrete stochastic process that yields a probability distribution in accordance with the preferential attachment principle.
In this view, it is both a Dirichlet process \cite{ferguson1973bayesian} as it generates a probability distribution and a preferential attachment process \cite{barabasi1999emergence} as the distribution is generated such that probabilities are proportional to past draws.
The process is explained through a metaphor of sitting patrons at a Chinese restaurant. Consider this restaurant as containing an infinite number of tables with each table having the capacity to seat an infinite number of patrons. Patrons are seated sequentially, such that the first patron is seated at the first table and every subsequent patron may be seated at an occupied table or the first unoccupied table. The probability of being seated at an occupied table is proportional to the number of patrons already seated at it. 
This process is illustrated through the toy example in Figure \ref{fig:crp} which shows a potential state of the CRP after sitting six patrons along with the sample probabilities of sitting the seventh.
Formally, the probability of seating patron $e_i$\footnote{We shall see later that patrons in this analogy are equivalent to entities in our model, hence the double use of $e$ to represent entities and patrons. This same principle applies to the use of $t$ to represent both tables and communities.} at a table $t_m$ in a restaurant where $\mathcal{T}_i$ is the set of occupied tables when patron $e_i$ arrives is:
\begin{equation}
{
 \mathbbm{P}(e_i = t_m \ | \ e_0, e_1, ..., e_{i-1}, \gamma) = \begin{dcases} 
          \dfrac{\#_i^m}{i  + \gamma} & t_m \in \mathcal{T}_i \\
          \dfrac{\gamma}{i  + \gamma} & t_m \notin\mathcal{T}_i \\ 
       \end{dcases}
}\label{eqn:crp}
\end{equation}
Here, $\#_i^m$ is the number of patrons seated at table $t_m$ when patron $e_i$ arrives and $\gamma > 0$ is a hyperparameter of the CRP responsible for controlling the probability that an incoming patron is seated at an unoccupied table such that increasing $\gamma$ increases this probability. Thus, increasing $\gamma$ values will yield results with an increasing number of occupied tables. Specifically, the expected number of occupied tables grows logarithmically with respect to the number of seated patrons:
\begin{equation}
    \mathbbm{E}\Bigg[\sum\limits_{t_m \in \mathcal{T}_i} \mathbbm{I}( \#_i^m > 0 )  \ | \ \gamma \Bigg] = \mathcal{O}(\gamma \log{i})
\end{equation}
Where $\mathbbm{I}$ is the indicator function which returns 1 if the condition is met and 0 otherwise. Big-O notation is leveraged with $\mathcal{O}$ to indicate the asymptotic upper bound of the expectation. This principle becomes relevant when controlling for the branching factor of the induced tree as we will see later on.
The realization of the CRP yields a partition of patrons over the infinitely many tables in the restaurant. If we consider each table to be a community, we can leverage this process to obtain a probability distribution over an infinite number of communities. Indeed this is the main utility of the CRP, namely to serve as a conjugate prior to infinite non-parametric discrete distributions. 
While this approach allows for the modelling of flat communities, it does not account for hierarchical relations between them. To remedy this, the CRP must be extended to its nested variant. 

\subsection{The Nested Chinese Restaurant Process}\label{subsec:ncrp}

\begin{figure}
\centering
\begin{tikzpicture}[
      base/.style = {draw, align=center},
  patron/.style = {base, rectangle, fill=white!30, minimum width=5mm, minimum height=5mm},
  every edge quotes/.style = {auto=right},
  circlenode/.style  ={base,minimum width=4em, shape=circle,thick, black,fill=gray!30,}]
\node (table2) [circlenode] {$t_2$};
\node (table6) [circlenode,  right= 0.6cm of table2, fill=gray!10, dashed] {$t_6$};
\node (table4) [circlenode,  right= 0.6cm of table6] {$t_4$};
\node (table5) [circlenode,  right= 0.6cm of table4] {$t_5$};
\node (table7) [circlenode,  right= 0.6cm of table5, fill=gray!10, dashed] {$t_7$};
\node (table9) [circlenode,  right= 0.6cm of table7, fill=gray!10, dashed] {$t_9$};
\node (dummy1) [above= 1.5cm of table2] {};
\node (table1) [circlenode, right= 0.2cm of dummy1] {$t_1$};
\node (table3) [circlenode, above= 0.9cm of table5] {$t_3$};
\node (table8) [circlenode, above= 0.9cm of table9, fill=gray!10, dashed] {$t_8$};
\node (table0) [circlenode, above= 0.9cm of table3] {$t_0$};
\node (patron0) [patron, below = 0.1cm of table2] {$e_0$};
\node (patron1) [patron, below = 0.1cm of table4] {$e_1$};
\node (patron2) [patron, below = 0.1cm of patron1] {$e_2$};
\node (patron3) [patron, below = 0.1cm of patron2] {$e_3$};
\node (patron4) [patron, below = 0.1cm of patron0] {$e_4$};
\node (patron5) [patron, below = 0.1cm of table5] {$e_5$};
\draw (table0) edge (table1);
\draw (table1) edge (table2);
\draw (table1) edge[dashed] (table6);
\draw (table0) edge (table3);
\draw (table3) edge (table4);
\draw (table3) edge (table5);
\draw (table3) edge[dashed] (table7);
\draw (table0) edge[dashed] (table8);
\draw (table8) edge[dashed] (table9);
\end{tikzpicture}
\caption{Toy example of a nCRP truncated to a depth of $L = 2$ after assigning patrons $e_0$ through $e_5$. Solid lines indicate paths which have been taken by patrons and thus exist in the tree whereas dashed lines indicate indicate potential paths. We illustrate Equation \ref{eqn:ncrp} by calculating the probability of a patron taking a path through communities $t_2$ and $t_9$: $\mathbbm{P}(e_6 = t_2) = (\frac{2}{2 + \gamma})( \frac{2}{6 + \gamma})$ and $\mathbbm{P}(e_6 = t_9) = \frac{\gamma}{6 + \gamma}$.}
\label{fig:toy_ncrp}
\end{figure}
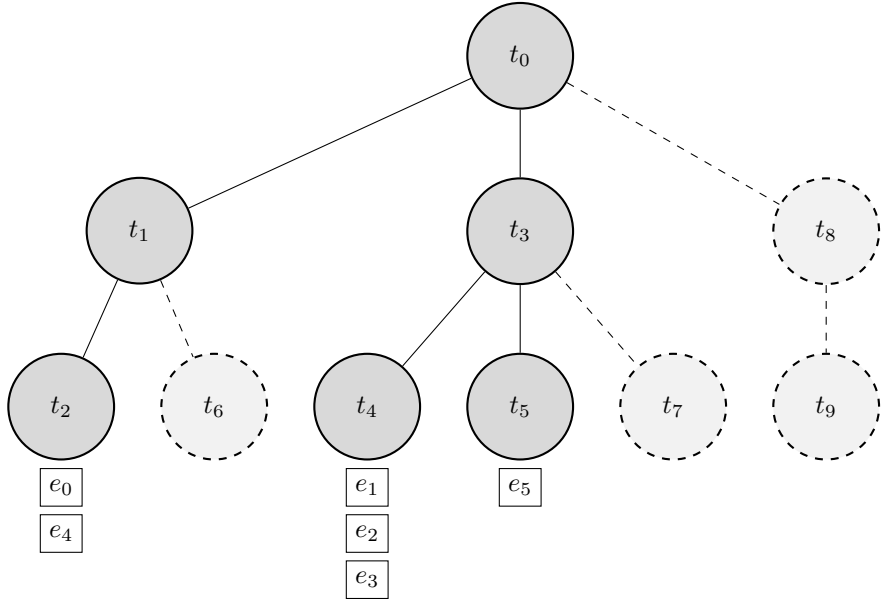

The nested Chinese restaurant process (nCRP) \cite{griffiths2003hierarchical,blei2010nested} is an extension of the CRP formulated to account for hierarchical relations between the generated communities. 
The realization of this process is an infinitely deep and infinitely branching tree of communities defined by a set of paths, $\mathcal{P}$, taken from the root community to a leaf community. In principle, the tree is unbounded in depth, however, we limit our discussion to a nCRP bounded to a depth of $L$. 
As in the case of the CRP, the allocation of paths along the tree is consistent with the preferential attachment principle.
The tree is generated stochastically by sampling a path at each level in the tree via the CRP such that drawing a table is analogous to taking a path at that level.
To extend the metaphor of seating patrons at a Chinese restaurant, consider the scenario of an infinite number of restaurants with an infinite number of infinite seat tables. When patrons are seated at these restaurants they are not served food but rather a table specific reference to another restaurant to which they must go. One of these restaurants is designated a root restaurant with no reference and all other restaurants are referenced exactly once. The seating of patrons at these restaurants is performed as in the CRP. We can see how realizing this process yields a tree by examining the paths taken by patrons. They first arrive at the root restaurant before being sent off to one of the root restaurant's descendant restaurants. At this restaurant, the patron is sent off to another descendant restaurant and this process is repeated until $L$ restaurants have been visited in the bounded case. The paths taken by patrons generate the tree as illustrated in the toy example in Figure \ref{fig:toy_ncrp}. 
As before, we extend this analogy of patrons and tables to entities and communities, respectively.
Thus, when drawing path $\mathbf{p}_i$ for entity $e_i$, the process starts by initializing the path at the top level to the root community, namely $p_i^0 = t_0$ where the superscript in path $\mathbf{p}_i$ indexes into the path vector to obtain the community at the corresponding level and $t_0$ is the root community.
The process then continues by drawing a descendant community according to the CRP.
Recall that this draw results in a community which either has or hasn't been visited before by a previous entity. The latter case corresponds to branching out a new path in the tree at the descendant level. This process is repeated $L$ times at which point the specified depth has been reached.
We can formalize this process by extending the previously defined notation. Specifically, let $\mathcal{T}_i$ be the set of communities in the tree before entity $e_i$ has its path sampled and $\mathcal{C}_i^q$ be the set of children communities for community $t_q$ at this time as well.
The sampling process is then expressed as follows: when entity $e_i$ arrives at community $t_q$ on the $(l-1)^{\text{th}}$ level in the tree, the probability of selecting an existing community, $p_i^l \in \mathcal{C}_i^q$ or creating a new community, ${p}_i^l \notin \mathcal{T}_i$, is:
\begin{align}
\mathbbm{P}(p_i^l = t_c \ | \ \mathbf{p}_{0}, \mathbf{p}_1, ..., \mathbf{p}_{i-1}, \mathbf{p}_i^{1 \colon l-1}, \gamma) & = \begin{dcases} 
  \dfrac{\#_i^{t_c}}{\#_i^{t_q} + \gamma} & t_c \in \mathcal{C}_i^c \\
  \dfrac{\gamma}{\#_i^{t_q} + \gamma} & t_c \notin \mathcal{T}_i \\
\end{dcases} \label{eqn:ncrp}
\end{align}
Where $\#_i^{t_q}$ and $\#_i^{t_c}$ is the number of entities that have passed through communities $t_q$ and $t_c$ before entity $e_i$ started its path. 
The superscript in $\mathbf{p}_i^{1:l-1}$ indicates that the probability distribution for sampling $p_i^l$ is conditioned on the path taken by entity $e_i$ up until level $l$.
The hyperparameter $\gamma$ serves a similar function as in the CRP, namely controlling the branching factor of the tree such that higher $\gamma$ values yield trees with more branches.
The use of the CRP in the path decision process ensures that probability mass will be pulled towards drawing paths which have been more frequently drawn before. The resulting distribution allows us to use the nCRP as a non-parametric prior over a tree structure in our model. In drawing paths, we not only generate a hierarchy but also define a subset of communities to which an entity can belong to, namely those along the path. 
This highlights an important difference between the CRP and the nCRP. While the CRP is sufficient for drawing a community for an entity, the nCRP must be used alongside another stochastic process to determine the level along the path that the entity belongs to. This provides a segue to one such process, specifically the stick breaking process.

\subsection{The Stick Breaking Process}\label{subsec:stick}

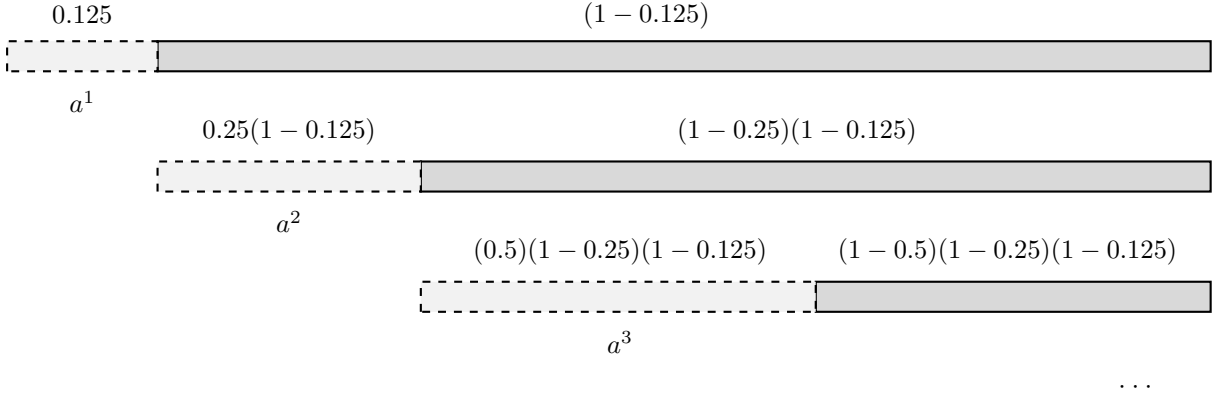
\begin{figure}[t]
\begin{tikzpicture}
\node[] at (1,-0.05) {$0.125$};
\node[] at (8.5,-0.05) {$(1 - 0.125)$};
\node[] at (1,-1.2) {$a^1$};
\draw[ fill=gray!30, thick] (2,-0.8) rectangle ++(14,0.4);
\draw[fill=gray!10, thick, dashed] (0,-0.8) rectangle ++(2,0.4);
\node[] at (3.75,-1.6) {$0.25(1-0.125)$};
\node[] at (10.5,-1.6) {$(1 - 0.25)(1 - 0.125)$};
\node[] at (3.75,-2.8) {$a^2$};
\draw[draw=black, fill=gray!30,thick] (5.5,-2.4) rectangle ++(10.5,0.4);
\draw[draw=black, fill=gray!10,thick, dashed] (2,-2.4) rectangle ++(3.5,0.4);
\node[] at (8.15,-3.2) {$(0.5)(1 - 0.25)(1 - 0.125)$};
\node[] at (13.3,-3.2) {$(1 - 0.5)(1 - 0.25)(1 - 0.125)$};
\node[] at (8.15,-4.4) {$a^3$};
\draw[draw=black, fill=gray!30,thick] (10.75,-4) rectangle ++(5.25,0.4);
\draw[draw=black, fill=gray!10,thick, dashed] (5.5,-4) rectangle ++(5.25,0.4);
\node[] at (15,-5) {$\cdot\cdot\cdot$};

\end{tikzpicture}
\caption{Toy example of the stick breaking process with values $v^1 = 0.125$ $v^2 = 0.25$ $v^3 = 0.5$. Starting at the top of the figure, a unit length stick is broken at $v^1$. The remainder is then iteratively broken proportionally to draws from the Beta distribution.}
\label{fig:toy_stick}
\end{figure}

The stick breaking process \cite{sethuraman1994constructive} is -- like the CRP and nCRP -- a Dirichlet process that draws its name from a metaphor which describes it. The metaphor starts by breaking a stick of unit length into two fragments at a point in the interval from 0 to 1 as drawn from the Beta distribution. One of the two fragments is preserved and the other fragment is broken again, analogously to the initial stick. This process is repeated an infinite number of times to yield an infinite number of fragments whose combined length is that of the initial stick. These fragments may be viewed as a probability distribution over the infinite sequence of discrete time-steps used to generate them. In other words, the stick breaking process is an infinite extension of the Dirichlet distribution insofar as while the Dirichlet distribution yields a probability distribution over $L$ categories, the stick breaking process yields a probability distribution over an \textit{infinite} number of categories.
Formally, let the draw from the Beta distribution at the $l^{\text{th}}$ iteration of the stick breaking process be denoted as $v^l \sim \text{Beta}(\mu\sigma, (1-\mu)\sigma)$. Thus, the lengths of the first fragment, denoted $a^1$, and its remainder are $v^1$ and $1 - v^1$, respectively. To obtain the length of the second fragment, $a^2$, draw $v^1$ and break off that fragment from what remains of the stick, namely $a^2 = v^1(1-v^1)$. We define this process for an arbitrary $l^\text{th}$ time-step as follows:
\begin{equation}
    a^l = v^l \prod\limits_{k=1}^{l-1} (1 - v^k)
\end{equation}
A concrete example involving the application of this rule is illustrated in Figure \ref{fig:toy_stick} which demonstrates the first three breaks of the stick along with the respective values of the broken fragments and their remainders. The realized stick fragments form a probability distribution in that $\sum_{l=1}^\infty a^l = 1$. We can thus define the probability mass function of the stick breaking process, denoted Stick$(\mu,\sigma)$, as follows:
\begin{align}
    \text{Stick}(\mu,\sigma) & =  \sum\limits_{l=1}^\infty a^l  \nonumber \\
    & = \sum\limits_{l=1}^\infty v^l \prod\limits_{k=1}^{l-1} (1 - v^k)  \label{eqn:stick}
\end{align}
The stick breaking process is a generalization of the Griffiths-Engen-McCloskey distribution \cite{blei2010nested,pitman2002combinatorial} which may be seen as a special case where $\mu\sigma = 1$. The hyperparameters, $1 > \mu > 0$ and $\sigma > 0$, control the mean and variance of the distribution, respectively. Specifically, increasing $\mu$ values will pull the mean towards fragments broken later in the process and increasing $\sigma$ values will increase the variance of the distribution.
The resulting distribution can be used in conjunction with the nCRP to obtain a community for an entity given its sampled path. This is because by sampling the stick breaking distribution an index is obtained which can correspond to the level on the path that the entity belongs to.
This motivates the use of the stick breaking process in our model.
Namely, we use the stick breaking process as a prior over the levels in the induced hierarchy. We explain this in detail in the subsequent section.

\section{Proposed Model}\label{sec:model}

In describing our proposed model, we will adopt the notations used in the previous section to indicate the connection with the ideas discussed in the preliminaries.
To aid in understanding, we first provide a summary of the components of our model before defining the generative process. This is followed by a formalization of the Gibbs sampling procedure and derivation of sampling equations.

\subsection{Model Description}\label{sec:desc}

Like all stochastic blockmodels, our model is defined as a set of probability distributions such that when these distributions are sampled from, they generate the adjacency tensor of the knowledge graph. The choice of these distributions makes assumptions about the underlying structure that governs the graph's interactions. 
In devising our model, we assume a hierarchy of entity communities which are captured in the form of a tree. 
The entities in these communities interact with one another as a function of their membership to a community. In other words, interactions are modelled at the community level and extended downwards to their constituent entities.
Unlike most stochastic blockmodels, these community relations are modelled with respect to a predicate in the knowledge graph. This allows the model to capture structures extending beyond those implied by mere interaction density.
Thus, in order to generate the knowledge graph's adjacency tensor, we need to know its hierarchical community structure, its entities' memberships to communities, and the interactions between its communities. 
The induction of these components, which may be seen as a byproduct of the generative process, is the objective of our model.
We note that the communities' constituent entities do not conform to is-a relationships as would be implied by the hierarchy. This is because the hierarchy is imposed on the communities themselves as opposed to their constituent entities. An example of this is highlighted in Figure \ref{fig:toy_hierarchy} where the entity \texttt{Canada} is a descendant of the entity \texttt{Pacific Ocean}. Of course, \texttt{Canada} is not a \texttt{Pacific Ocean} however the concept modelled by community $t_5$, namely countries, is an instance of the concept modelled by community $t_2$, namely locations.

\subsubsection{Community Memberships}

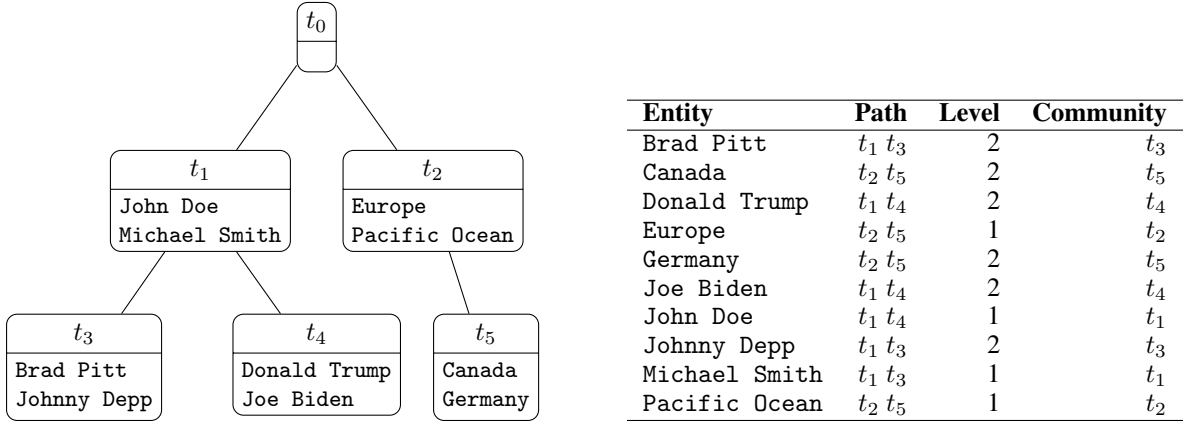
\begin{figure}
  \begin{minipage}[b]{0.50\linewidth}

\begin{tikzpicture}[scale = 1, sibling distance=8.8em,
  every node/.style = {shape=rectangle, rounded corners,
    draw, align=justify,
    top color=white, bottom color=white}]]
  \node [rectangle split, rectangle split, rectangle split parts=2,
         text ragged, font=\normalsize\selectfont, anchor=north] {\textbf{$t_0$}}
    child {  node [rectangle split, rectangle split, rectangle split parts=2,
         text ragged, font=\normalsize\selectfont, anchor=north] {
            \textbf{$t_1$}
                  \nodepart{second}
                   \small{\texttt{John Doe}} \\ \small{\texttt{Michael Smith}}
        }
        child { node [rectangle split, rectangle split, rectangle split parts=2,
         text ragged, text ragged, font=\normalsize\selectfont,anchor = north] {
            \textbf{$t_3$}
                  \nodepart{second}
                   \small{\texttt{Brad Pitt}} \\
                   \small{\texttt{Johnny Depp}}
        } }
        child { node [rectangle split, rectangle split, rectangle split parts=2,
         text ragged, text ragged, font=\normalsize\selectfont, anchor = north] {
            \textbf{$t_4$}
                  \nodepart{second}
                   \small{$\texttt{Donald Trump}$}  \\ \small{\texttt{Joe Biden}}
        }}
    }
     child {  node [rectangle split, rectangle split, rectangle split parts=2,
         text ragged, text ragged, font=\normalsize\selectfont,  anchor = north] {
            \textbf{$t_2$}
                  \nodepart{second}
                   \small{\texttt{Europe}} \\ \small{\texttt{Pacific Ocean}} 
        }
        child { node [rectangle split, rectangle split, rectangle split parts=2,
         text ragged, text ragged, font=\normalsize\selectfont,anchor = north west] {
            \textbf{$t_5$}
                  \nodepart{second}
                   \small{\texttt{Canada}} \\
                   \small{\texttt{Germany}}
        } }
    }
      ;
\end{tikzpicture}
  \end{minipage}%
  \begin{minipage}[t]{0.50\linewidth}
    \begin{tabular}[b]{l r r r}
    \hline
    \textbf{Entity} & \textbf{Path} & \textbf{Level} & \textbf{Community} \\ 
    \hline
    \texttt{Brad Pitt} & $t_1$ $t_3$ & 2 & $t_3$ \\
    \texttt{Canada} & $t_2$ $t_5$ & 2 & $t_5$ \\
    \texttt{Donald Trump} & $t_1$ $t_4$ & 2 & $t_4$ \\
    \texttt{Europe} & $t_2$ $t_5$ & 1 & $t_2$ \\
    \texttt{Germany} & $t_2$ $t_5$ & 2 & $t_5$ \\
    \texttt{Joe Biden} & $t_1$ $t_4$ & 2 & $t_4$ \\
    \texttt{John Doe} & $t_1$ $t_4$ & 1 & $t_1$  \\ 
    \texttt{Johnny Depp} & $t_1$ $t_3$ & 2 & $t_3$  \\
    \texttt{Michael Smith} & $t_1$ $t_3$ & 1 & $t_1$ \\ 
    \texttt{Pacific Ocean} & $t_2$ $t_5$ & 1 & $t_2$ \\ 
    \hline
    \end{tabular}
    \end{minipage}
\caption{Toy example depicting a potential hierarchy induced by our model. The table on the right side captures the path and level sampled for each entity in the knowledge graph as well as its corresponding community. The left side provides a visualization of this hierarchy.}
\label{fig:toy_hierarchy}
\end{figure}

Entities are assigned to communities through the conjunction of two variables: entity paths and level indicators.
Paths define the tree structure over the community hierarchy by sampling from the nCRP as described in the previous section.
We thus denote an entity path as $\mathbf{p}_i$ for entity $e_i$, such that $\mathbf{p}_i := [ p^1_i, p^2_i, ..., p^L_i ]$ where $p^l_i$ represents the community at level $l$. We draw attention to the fact that this definition omits the root community from the path, namely $p_i^0$, since all entities must pass through it. It also allows a hierarchy with a depth of $L$ to have entity path vectors of dimension $L$, simplifying the notation.
Entity paths are drawn from the nCRP, denoted as $\mathbf{p}_i \sim \text{nCRP}(\gamma)$.
Thus, all the entity paths sampled in the model form a $|\mathcal{E}| \times L$ matrix which we denote as $\mathbf{P}$.
$\gamma$ is the aforementioned hyperparameter of the nCRP and is responsible for controlling the probability of generating a new branch in the hierarchy as the path is being sampled. When a new branch is generated at level $l$ such that $l > L$, $L - l$ new communities are also generated and populated solely by the sampling entity. Furthermore, if a path is resampled such that its corresponding entity obtains a new path which leaves behind empty communities, those empty communities and removed from the hierarchy.
As such, the number of communities in the hierarchy is subject to constant change throughout the sampling process. 
Having sampled entity paths, in order for entities to be assigned to communities, their levels must be obtained. Entity levels are modelled by two variables in our approach: level memberships and level indicators. Level memberships, denoted $\mathbf{a}_i$ for entity $e_i$, capture the probability of the entity's belonging to each of the $L$ levels. As such, all the level memberships in our model form a $| \mathcal{E} | \times L$ matrix, $\mathbf{A}$. 
This is similar to the mixed-membership property of the Mixed Membership Stochastic Blockmodel wherein an entity has a membership distribution over all communities. The difference, as pointed out by Ho et al. \cite{ho2011multiscale}, is that in hierarchical models this distribution is restricted to communities along the entity's sampled path, otherwise the process of obtaining paths, and indeed the hierarchy itself, would lose its meaning. Level memberships are drawn from the stick breaking process, $\mathbf{a}_i \sim \text{Stick}(\mu,\sigma)$ with hyperparameters $\mu$ and $\sigma$. Recall that this process yields an infinite distribution and must therefore be truncated to a dimension of $L$ to correspond with the depth of the tree. 
The truncation is performed by removing all probabilities at levels greater than $L$ and renormalizing. 
The distribution captured by an entity's level membership is used to sample its level indicator. The level indicator indicates the level to which an entity belongs and thus, in conjunction with its path, assigns it to a community.
Level indicators are drawn in the context of an interaction between two entities.
Specifically, when modelling the probability of an interaction from entity $e_i$ to entity $e_j$ we draw two level indicators, one for the sender entity and one for the receiver entity denoted as ${z}_{i \rightarrow j}$ and ${z}_{i \leftarrow j}$, respectively. The sender and receiver level indicators correspond to the levels of entities $e_i$ and $e_j$ in the context of their pairwise interaction. Thus, our model samples $|\mathcal{E}|^2$ sender and receiver level indicators each leading to two $|\mathcal{E}| \times |\mathcal{E}|$ matrices $\mathbf{Z}_\rightarrow$ and $\mathbf{Z}_\leftarrow$ for all the senders and receivers, respectively.
To simplify notation in our inference procedure, we concatenate these matrices to form a $|\mathcal{E}| \times |\mathcal{E}| \times 2$ level indicator tensor, $\mathbf{Z}$.
Since level memberships are themselves probability distributions, they may be sampled from directly to indicate an entity's level.
Specifically, level indicators are drawn from multinomial distributions, namely ${z}_{i \rightarrow j} \sim \text{Multinomial}(\mathbf{a}_i)$ and ${z}_{i \leftarrow j} \sim \text{Multinomial}(\mathbf{a}_j)$, which yield one of the $L$ levels in the hierarchy.
The interplay between paths and levels when assigning entities to communities may be summarized as follows: paths identify a hierarchy of candidate communities and level indicators select one of the candidates for the entity. This dynamic is captured in the toy example in Figure \ref{fig:toy_hierarchy}.

\subsubsection{Community Relations}

\begin{figure}
    \begin{minipage}[b]{0.50\linewidth}

\begin{tikzpicture}[scale = 1, sibling distance=8.8em,
  every node/.style = {shape=rectangle, rounded corners,
    draw, align=justify,
    top color=white, bottom color=white}]]
  \node [rectangle split, rectangle split, rectangle split parts=2,
         text ragged, font=\normalsize\selectfont, anchor=north] {\textbf{$t_0$}}
    child {  node [rectangle split, rectangle split, rectangle split parts=2,
         text ragged, font=\normalsize\selectfont, anchor=north] {
            \textbf{$t_1$}
                  \nodepart{second}
                   \small{\texttt{John Doe}} \\ \small{\texttt{Michael Smith}}
        }
        child { node [rectangle split, rectangle split, rectangle split parts=2,
         text ragged, text ragged, font=\normalsize\selectfont,anchor = north] {
            \textbf{$t_3$}
                  \nodepart{second}
                   \small{\texttt{Brad Pitt}} \\
                   \small{\texttt{Johnny Depp}}
        } }
        child { node [rectangle split, rectangle split, rectangle split parts=2,
         text ragged, text ragged, font=\normalsize\selectfont, anchor = north] {
            \textbf{$t_4$}
                  \nodepart{second}
                   \small{\texttt{Donald Trump}}  \\ \small{\texttt{Joe Biden}}
        }}
    }
     child {  node [rectangle split, rectangle split, rectangle split parts=2,
         text ragged, text ragged, font=\normalsize\selectfont,  anchor = north] {
            \textbf{$t_2$}
                  \nodepart{second}
                   \small{\texttt{Europe}} \\ \small{\texttt{Pacific Ocean}} 
        }
        child { node [rectangle split, rectangle split, rectangle split parts=2,
         text ragged, text ragged, font=\normalsize\selectfont,anchor = north west] {
            \textbf{$t_5$}
                  \nodepart{second}
                   \small{\texttt{Canada}} \\
                   \small{\texttt{Germany}}
        } }
    }
      ;
\end{tikzpicture}
  \end{minipage}%
  \begin{minipage}[t]{0.50\linewidth}
    \begin{tikzpicture}[auto matrix/.style={matrix of nodes,
  nodes in empty cells,
  cells={nodes={minimum width=1.9em,minimum height=1.9em,
   draw,anchor=center,fill=white
  }}}]
 \matrix[auto matrix=z,xshift=2.8em,yshift=2.8em](matz) at (0,0){
  |[draw=none]| & |[draw=none]|  $t_1$ & |[draw=none]|  $t_2$ \\
  |[draw=none]|  $t_1$ & |[draw=none]|  & 0.2  \\
  |[draw=none]|  $t_2$ & |[draw=none]|  & 0  \\
 };
 \matrix[auto matrix=y,xshift=1.4em,yshift=1.4em] at (0,0) (maty){
  |[draw=none]| & |[draw=none]|  $t_1$ & |[draw=none]|  $t_2$ \\
  |[draw=none]|  $t_1$ & |[draw=none]|  & 0  \\
  |[draw=none]|  $t_2$ & |[draw=none]|  & 0.3  \\
 };
 \matrix[auto matrix=x] at (0,0) (matx){
  |[draw=none]| & |[draw=none]|  $t_1$ & |[draw=none]| $t_2$ \\
  |[draw=none]|  $t_1$ & 0.3 & 0  \\
  |[draw=none]|  $t_2$ &  0 & 0  \\
 };
 \matrix[auto matrix=z,xshift=2.8em,yshift=2.8em](matc) at (-2,-3.1){
  |[draw=none]| & |[draw=none]|  $t_3$ & |[draw=none]|  $t_4$ \\
  |[draw=none]|  $t_3$ & |[draw=none]|  & 0  \\
  |[draw=none]|  $t_4$ & |[draw=none]|  & 0 \\
 };
 \matrix[auto matrix=y,xshift=1.4em,yshift=1.4em] at (-2,-3.1) (matb){
  |[draw=none]| & |[draw=none]|  $t_3$ & |[draw=none]|  $t_4$ \\
  |[draw=none]|  $t_3$ & |[draw=none]|  & 0  \\
  |[draw=none]|  $t_4$ & |[draw=none]|  & 0  \\
 };
 \matrix[auto matrix=a] at (-2,-3.1) (mata){
  |[draw=none]| & |[draw=none]|  $t_3$ & |[draw=none]| $t_4$ \\
  |[draw=none]|  $t_3$ & 0.8 & 0.6  \\
  |[draw=none]|  $t_4$ & 0.7  & 0.9  \\
  };  
\matrix[auto matrix=y,xshift=2.8em,yshift=2.8em] at (2,-3.1) (matf){
  |[draw=none]| & |[draw=none]|  $t_5$ \\
  |[draw=none]|  $t_5$ & 0 \\
  };
\matrix[auto matrix=y,xshift=1.4em,yshift=1.4em] at (2,-3.1) (mate){
  |[draw=none]| & |[draw=none]|  $t_5$ \\
  |[draw=none]|  $t_5$ & 0.1 \\
  };
  \matrix[auto matrix=a] at (2,-3.1) (matd){
  |[draw=none]| & |[draw=none]|  $t_5$ \\
  |[draw=none]|  $t_5$ & 0 \\
  };
 \node[] at ([xshift=0.5ex]matz.south east) {\scriptsize{\texttt{bornIn}}};
 \node[] at ([xshift=0.5ex]maty.south east) {\scriptsize{\texttt{locatedIn}}};
 \node[] at ([xshift=0.5ex]matx.south east) {\scriptsize{\texttt{knows}}};
 \node[] at ([xshift=0.5ex]matc.south east) {\scriptsize{\texttt{bornIn}}};
 \node[] at ([xshift=0.5ex]matb.south east) {\scriptsize{\texttt{locatedIn}}};
 \node[] at ([xshift=0.5ex]mata.south east) {\scriptsize{\texttt{knows}}};
  \node[] at ([xshift=0.5ex]matf.south east) {\scriptsize{\texttt{bornIn}}};
 \node[] at ([xshift=0.5ex]mate.south east) {\scriptsize{\texttt{locatedIn}}};
 \node[] at ([xshift=0.5ex]matd.south east) {\scriptsize{\texttt{knows}}};
\end{tikzpicture}
\end{minipage}
    \caption{The potential community relations induced by our model on the toy example introduced earlier. The hierarchy on the left of the figure has three sibling groups and three predicates: $\texttt{knows}$, $\texttt{locatedIn}$, and $\texttt{bornIn}$. The three tensors on the right correspond to the community relations of the three sibling groups.}
    \label{fig:toy_example_communities}
\end{figure}
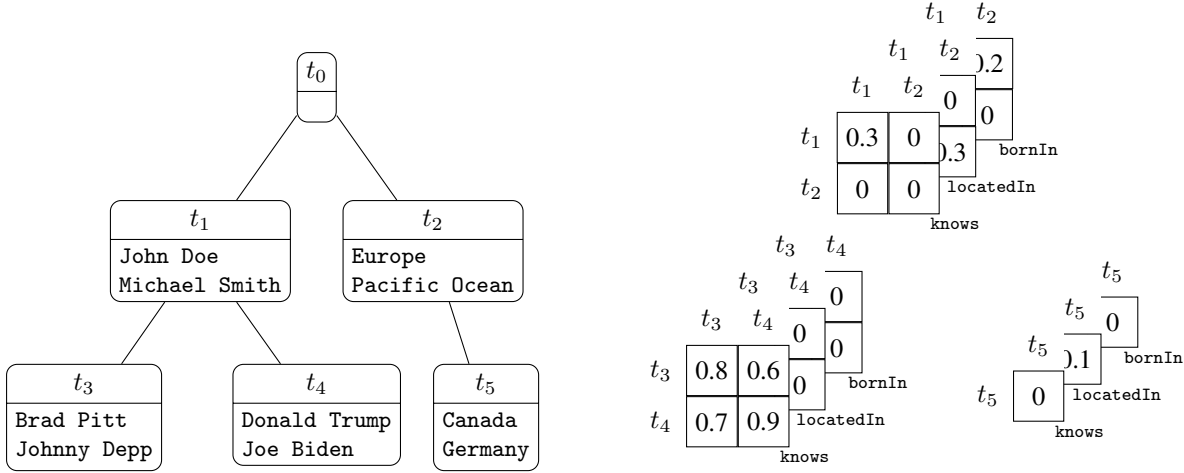

Community relations describe the degree to which entities in any two communities are likely to interact with one another through a specific predicate. In other words, they model the probability of observing a value of one in the knowledge graph's adjacency tensor.
These interactions are captured by a $\mathcal{T} \times \mathcal{T} \times \mathcal{R}$ tensor, denoted $\mathbf{C}$, where $\mathcal{T}$ is the set of all communities in the hierarchy.
We note that because the communities in $\mathcal{T}$ are a result of sampling from the nCRP and are thus subject to change with each successive sample, the dimensionality of $\mathbf{C}$ is also subject to change in the sampling process. This presents a challenge to our sampling scheme since it is possible to sample communities via the nCRP for which there are no community relation values. We overcome this issue through the marginalization of community relations as discussed in the subsequent subsection.
The community relation $c_{pqr}$ is an entry in $\mathbf{C}$ and captures the probability of interaction between entities in community $t_p$ with entities in community $t_q$ through predicate $r_r$. As such, the value of $c_{pqr}$ is bounded to $1 \geq c_{pqr} \geq 0$.
In order to preserve the hierarchical structure that was induced by sampling paths and levels, the community relations must be limited to take on non-zero values only when interacting with communities which are proximal to them in the hierarchy. This restriction is vital as allowing for interaction between any two communities in the hierarchy would render it meaningless and our model would be reduced to a fixed size mixed membership stochastic blockmodel such as the ones described in Section \ref{sec:rw}.

In restricting the values of community relations we take an approach similar to that of the Multiscale Community Blockmodel. Specifically, we borrow the concept of a sibling group which refers to a set of communities that share the same parent in the hierarchy.
Only the community relations between communities in the same sibling group are modelled in our approach. Thus, when obtaining the interaction degree of two entities whose communities have the same parent, it's sufficient to merely access the corresponding value in $\mathbf{C}$. When their communities do not share the same parent, a coarsening procedure is applied to obtain an interaction degree. The coarsening procedure traverses the paths of the two entities to find the deepest pair of communities which are in the same sibling group.
Formally, to obtain the community relation degree from entity $e_i$ to entity $e_j$ on predicate $r_r$, we define the function $\Psi(i,j,r)$ as follows:
\begin{align}
{  
\Psi(i,j,r) = \begin{dcases} 
  c_{p_i^{z_{i \rightarrow j}}p_j^{z_{i \leftarrow j}}r} &  p_i^{z_{i \rightarrow j} - 1} = p_j^{z_{i \leftarrow j} - 1}\\
  c_{\Phi(i \ | \ j)\Phi(j \ | \ i)r} &  p_i^{z_{i \rightarrow j} - 1} \neq p_j^{z_{i \leftarrow j} - 1} \\
\end{dcases}
}
\end{align}
Wherein $\Phi(i \ | \ j)$ and $\Phi(j \ | \ i)$ are functions that find the ancestor communities of entities $e_i$ and $e_j$ which share the same sibling group, respectively. These values are obtained by indexing the entities' paths on the level at which they diverge from their interaction partner. This process is made clear in their definitions:
\begin{align}
    \Phi(i \ | \ j) = p_i^{min(\{l \ : \ p_i^l \ \neq \ p_j^l\})} \nonumber \\
    \Phi(j \ | \ i) = p_j^{min(\{l \ : \ p_i^l \ \neq \ p_j^l\})} \label{eqn:phi}
\end{align}
This approach differs from the Multiscale Community Blockmodel in that while there are restricted entries in $\mathbf{C}$, these values are never accessed. Instead, all communities are coarsened to an ancestor which allows for their interaction to take on a non-restricted value.

Community relations are drawn from the Beta distribution parameterized by $\lambda > 0$ and $\eta > 0$, and denoted as $c_{pqr} \sim \text{Beta}(\lambda, \eta)$. This ensures that community relations take on probability values which can be used in conjunction with the Bernoulli distribution. $\lambda$ and $\eta$ are hyperparameters of our model and determine the density of the generated knowledge graph such that increasing $\lambda$ values with respect to $\eta$ yields denser results. Figure \ref{fig:toy_example_communities} provides a visualization of a potential sampling of community relations. We note that the three tensors correspond to the three sibling groups for which community relations take on non-restricted values. The diagonal and off-diagonal values in these tensors represent the intra and inter community relations, respectively. Thus, based on these values, there is a probability of 0.8 that \texttt{Brad Pitt} \texttt{knows} \texttt{Johnny Depp} and a probability of 0.6 that \texttt{Brad Pitt} \texttt{knows} \texttt{Donald Trump}.
We provide an exploration of the recovered community relations on real-world data in Section \ref{sec:evaluation}.

\subsubsection{Generative Process}

\begin{figure}
\centering
\begin{tikzpicture}[auto,thick,node distance=4.3em, scale=0.80, transform shape]
\tikzset{
  circlenode/.style  ={draw,minimum width=2.8em, shape=circle,thick, black,fill=white},
}
\tikzstyle{line} = [draw, -latex']
    \node[circlenode,] (gamma) {$\gamma$};
    \node[below of = gamma,circle,fill,inner sep=0.4pt] (point1) {.};
    \node[circlenode, right = 1cm of  gamma] (c1) {$p_i^0$};
    \node[circlenode, below of = c1] (c2) {$p_i^1$};
    \node[circlenode, below of = c2, draw = none, scale = 0.7] (dots1) {$...$};
    \node[circlenode, below of = dots1] (cl) {$p_i^L$};
    \node[left = 1.4cm of cl,circle,fill,inner sep=0.4pt] (point2) {.};
    \node[right of = c1,circle,fill,inner sep=0.4pt] (point3) {.};
    \node[right of = c2,circle,fill,inner sep=0.4pt] (point4) {.};
    \node[circlenode, right of = cl] (pi) {$\mathbf{p}_i$};
    \node[right of = dots1,circle,fill,inner sep=0.4pt] (point5) {.};
    \node[circlenode, right of = point5] (zij) {$z_{i \rightarrow j}$};
    \node[circlenode, right = 1.25cm of pi] (gij) {$g_{ijr}$};
    \node[circlenode, right of = point3] (mu) {$\mu$};
    \node[circlenode, right of = mu] (sigma) {$\sigma$};
    \node[circlenode, right of = zij] (zji) {$z_{i \leftarrow j}$};
    \node[circlenode, right = 1.68cm of point4] (ai) {$\mathbf{a}_i$};
    \node[circlenode, right = 2.05cm of gij] (c) {$c_{pqr}$};
    \node[circlenode, right of = zji] (lambda1) {$\lambda$};
    \node[circlenode, right of = lambda1] (lambda2) {$\eta$};

    \path (gamma) edge (point1);
    \draw [-latex] (gamma) to (c1);
    \draw [-latex] (point1) to (c2);
    \draw [-latex] (c1) to (c2);
    \draw [-latex] (c2) to (dots1);
    \draw [-latex] (dots1) to (cl);
    \draw [-latex] (c1) to [out=225,in=135] (cl);
    \draw [-latex] (c2) to [out=225,in=115] (cl);
    \draw (point1) edge (point2);
    \draw [-latex] (point2) to (cl);
    \draw (c1) edge (point3);
    \draw (c2) edge (point4);
    \draw (point3) edge (point4);
    \draw (point4) edge (point5);
    \draw (point5) edge (pi);
    \draw (cl) edge (pi);
    \draw [-latex] (pi) to (gij);
    \draw [-latex] (zij) to (gij);
    \draw [-latex] (zji) to (gij);
    \draw [-latex] (ai) to (zij);
    \draw [-latex] (ai) to (zji);
    \draw [-latex] (mu) to (ai);
    \draw [-latex] (sigma) to (ai);
    \draw [-latex] (lambda1) to (c);
    \draw [-latex] (lambda2) to (c);
    \draw [-latex] (c) to (gij);
\end{tikzpicture}
\caption{Plate diagram for our model.}
\label{fig:plate}
\end{figure}
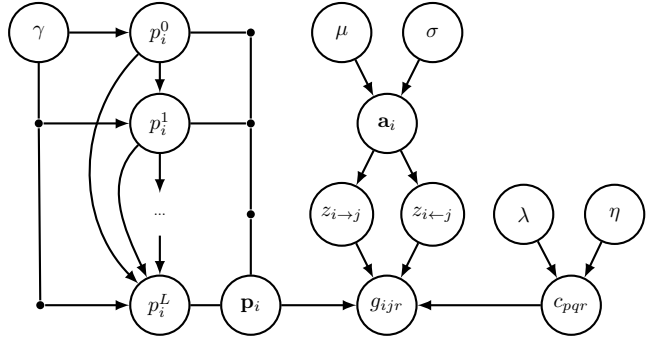

The generative process of our model refers to the sequential sampling of components which allows for the generation of the target knowledge graph. In other words, the goal is to draw a binary value for each $g_{ijr} \in \mathbf{G}$ such that it equals the knowledge graph's adjacency tensor. But before this can be done, it's necessary to sample the variables it is dependent on.
The first components sampled in the generative process are the paths and level memberships for each entity in the knowledge graph from the nCRP and stick distributions, respectively. 
Having drawn the paths, we now have the set of communities in the hierarchy and can draw community relations from the Beta distribution.
At this point in the generative process, entities are not yet assigned to communities. The community memberships for these entities have been drawn, however, allowing for the sampling of community levels for each pair of entities in the knowledge graph from the multinomial distribution.
With the community levels drawn, all the components for generating the knowledge graph are in place. The binary value for the interaction from entity $e_i$ to entity $e_j$ on predicate $r_r$ is drawn from the Bernoulli distribution using each entity's respective community's interactions, namely $g_{ijr} \sim \text{Bernoulli}(\Psi(i, j, r))$.
The plate diagram for this process is illustrated in Figure \ref{fig:plate} and the formal definition is as follows:
\begin{itemize}
    \item For each entity in the knowledge graph; $e_i \in \mathcal{E}$
    \begin{itemize}
        \item $\mathbf{p}_i \sim \text{nCRP}(\gamma)$
        \item $\mathbf{a}_i \sim  \text{Stick}(\mu,\sigma)$
    \end{itemize}
    \item For each sender community in the hierarchy; $t_p \in \mathcal{T}$
    \begin{itemize}
        \item For each receiver community in the hierarchy; $t_q \in \mathcal{T}$
        \begin{itemize}
            \item For each predicate in the knowledge graph; $r_r \in \mathcal{R}$
            \begin{itemize}
                \item $c_{pqr} \sim \text{Beta}(\lambda, \eta)$ 
            \end{itemize}
        \end{itemize}
    \end{itemize}
    \item For each sender entity in the knowledge graph; $e_i \in \mathcal{E}$
    \begin{itemize}
        \item For each receiver entity in the knowledge graph; $e_j \in \mathcal{E}$
        \begin{itemize}
            \item $z_{i \rightarrow j} \sim \text{Multinomial}(\mathbf{a}_i)$
            \item $z_{i \leftarrow j} \sim \text{Multinomial}(\mathbf{a}_j)$
            \item For each predicate in the knowledge graph; $r_r \in \mathcal{R}$
            \begin{itemize}
                \item $g_{ijr} \sim \text{Bernoulli}(\Psi(i, j, r))$
            \end{itemize}
        \end{itemize}
    \end{itemize}
\end{itemize}
We note that this process is unsupervised and does not impose any assumptions about the partition of entities to communities or the structure of the hierarchy other than to limit its depth. In fact, the depth is the only constraint imposed on the generative process. The other hyperparameters which must be specified a priori -- namely $\gamma$, $\mu$, $\sigma$, $\lambda$, and $\eta$ -- merely influence the prior distributions of our model. They may pull the latent variables in the assumed direction but only insofar as the data allows it. This, recall, is due to the sampling of latent variables from their posterior distribution which is conditioned on the data. As a result, with a strong enough likelihood, the effects of the hyperparameters and the prior relatively diminish. As with most stochastic blockmodels, the exact inference for our model is intractable and must be approximated using an inference scheme. For this we adopt collapsed Gibbs sampling, an extension of the aforementioned Gibbs sampling.

\subsection{Collapsed Gibbs Sampling}\label{subsec:gibbs}

Collapsed Gibbs sampling refers to an extension of Gibbs sampling in which a subset of model variables are marginalized over and therefore do not need to be sampled directly. These variables are said to be collapsed out of the Gibbs sampler. Collapsing of these variables is done analytically via integration and ensures a faster mixing process. This is because the calculation of probability distributions for sampling is generally computationally expensive. Having fewer variables then leads to a faster arrival at the desired stationary distribution.
Furthermore, the calculation of probability distributions which have not been collapsed out of the sampling process is generally faster in collapsed Gibbs sampling. This is because in regular Gibbs sampling draws are made from the full conditionals of variables. In collapsed Gibbs sampling, collapsed variables have been integrated out of the process and the remaining variables are conditioned on a lower-dimensional space.
Collapsing of variables is usually tractable when they are the conjugate prior of their dependent variables. In our model, community relations and level memberships are both conjugate priors of their dependant variables, namely level indicators and entity relations, respectively. We leverage these conjugacies to marginalize over these two variables in our sampling process.
After marginalization, the sampling equations may be derived for the remaining variables.

\subsubsection{Marginalizing Community Relations}\label{subsubsec:marginalizing_community_interactions}

In order to marginalize out community relations, it is necessary to find a closed form solution which allows for integration during path sampling. To this end, we can leverage the Bernoulli-Beta conjugacy which ensures that given a Bernoulli likelihood and Beta prior, the posterior will also be drawn from the Beta distribution. Employing this conjugacy is possible due to the formulation of our model in which entity relations are drawn from the Bernoulli distribution and community relations assume a Beta prior.
We see this explicitly when applying Bayes' theorem to obtain the posterior as follows:
\begin{equation}
    \mathbbm{P}(c_{pqr} \ | \ \mathbf{C}_{-(pqr)}, \mathbf{G}, \mathbf{P}, \mathbf{Z}, \lambda, \eta) = \dfrac{\mathbbm{P}(\mathbf{G} \ | \ \mathbf{C}, \mathbf{P}, \mathbf{Z}, \lambda, \eta) \mathbbm{P}(c_{pqr} \ | \ \mathbf{C}_{-(pqr)}, \lambda, \eta)}{\bigintssss_{c_{pqr}} \mathbbm{P}(\mathbf{G} \ | \ \mathbf{C}, \mathbf{P}, \mathbf{Z}, \lambda, \eta) \mathbbm{P}(c_{pqr} \ | \ \mathbf{C}_{-(pqr)}, \lambda, \eta) \ \mathrm{d}c_{pqr}} \label{eqn:community_interactions_posterior}
\end{equation}
Where $\mathbbm{P}(\mathbf{G} \ | \ \mathbf{C}, \mathbf{P}, \mathbf{Z}, \lambda, \eta)$ is the likelihood of generating entity relations and $\mathbbm{P}(c_{pqr} \ | \ \mathbf{C}_{-(pqr)}, \lambda, \eta)$ is the prior placed on community relations. $\mathbf{C}_{-(pqr)}$ indicates the community relations tensor $\mathbf{C}$ without $c_{pqr}$.
Before proceeding we introduce helper variables $\#^{c_{pqr}=1}$ and $\#^{c_{pqr}=0}$ to indicate the number of existing and non-existing interactions between entities from community $t_p$ to community $t_q$ on predicate $r_r$, respectively:
\begin{align}
    \#^{c_{pqr}=1} = \Big|\Big\{g_{xyz} \in \mathbf{G} \  : \ \Psi(x,y,z) = c_{pqr} \land g_{xyz} = 1 \Big\}\Big| \nonumber \\
    \#^{c_{pqr}=0} = \Big|\Big\{g_{xyz} \in \mathbf{G} \  : \  \Psi(x,y,z) = c_{pqr} \land g_{xyz} = 0 \Big\}\Big|
\end{align}
We can now derive a closed-form solution for the posterior of community relations by applying the distributions defined in our model:
\begin{align}
        \mathbbm{P}(c_{pqr} \ | \ \mathbf{C}_{-(pqr)}, \mathbf{G}, \mathbf{P}, \mathbf{Z}, \lambda, \eta) & \overset{\mathrm{(1)}}{=} \dfrac{\Big( \prod_{g_{xyz} \in \mathbf{G}} \text{Bernoulli}(c_{pqr}, 1 - c_{pqr}) \Big) \Big( \text{Beta}(\lambda, \eta) \Big)}{\bigintss_{c_{pqr}} \Big( \prod_{g_{xyz} \in \mathbf{G}} \text{Bernoulli}(c_{pqr}, 1 - c_{pqr}) \Big) \Big( \text{Beta}(\lambda, \eta) \Big) \ \text{d}c_{pqr}}  \nonumber \\
        & \overset{\mathrm{(2)}}{=} \dfrac{\Bigg(c_{pqr}^{\#^{c_{pqr}=1}}(1-c_{pqr})^{\#^{c_{pqr}=0}}\Bigg) \Bigg( \dfrac{c_{pqr}^{\lambda - 1}(1-c_{pqr})^{\eta - 1}}{\text{B}(\lambda,\eta)}\Bigg)}{\bigints_{\ c_{pqr}} \Bigg( c_{pqr}^{\#^{c_{pqr}=1}}(1-c_{pqr})^{\#^{c_{pqr}=0}} \Bigg) \Bigg( \dfrac{c_{pqr}^{\lambda - 1}(1-c_{pqr})^{\eta - 1}}{\text{B}(\lambda,\eta)} \Bigg) \ \text{d}c_{pqr}} \nonumber \\
        & \overset{\mathrm{(3)}}{=} \dfrac{c_{pqr}^{\#^{c_{pqr}=1} + \lambda - 1}(1-c_{pqr})^{\#^{c_{pqr}=0} + \eta - 1}}{\bigintss_{\ c_{pqr}} c_{pqr}^{\#^{c_{pqr}=1} + \lambda - 1}(1-c_{pqr})^{\#^{c_{pqr}=0} + \eta - 1} \ \text{d}c_{pqr}} \nonumber \\
        & \overset{\mathrm{(4)}}{=} \dfrac{c_{pqr}^{\#^{c_{pqr}=1} + \lambda - 1}(1-c_{pqr})^{\#^{c_{pqr}=0} + \eta - 1}}{\text{B}(\#^{c_{pqr}=1} + \lambda, \#^{c_{pqr}=0} + \eta)} \nonumber \\
        & = \text{Beta}(\#^{c_{pqr}=1} + \lambda, \#^{c_{pqr}=0} + \eta) \label{eqn:marginalized_community_interactions}
\end{align}
Such that in the derivation above: (1) is the Bayes' theorem definition as per Equation \ref{eqn:community_interactions_posterior} using the probability distributions defined in our model; (2) uses the probability masses and densities of the Bernoulli and Beta distributions as per Equations \ref{eqn:bernoulli} and \ref{eqn:beta}; (3) is obtained by applying power rules and dividing out the Beta function which is constant with respect to $c_{pqr}$ in the integral; and (4) utilizes the integral form of the Beta function as derived in Equation \ref{eqn:normalizingconstant}. The posterior as defined in Equation \ref{eqn:marginalized_community_interactions} allows for community relations to be integrated out when sampling paths. As such they are not sampled directly in the inference process. 

\subsubsection{Marginalizing Level Memberships}

There are two ways in which to approach marginalizing level memberships in our model. Firstly, Sethuraman \cite{sethuraman1994constructive} showed that the realization of the stick breaking process follows the Dirichlet distribution. We can leverage this because, in practice, the dimensionality of the level memberships gets bounded to the depth of the tree, $L$. It is therefore possible to model level memberships with an $L$ dimensional Dirichlet distribution. As discussed in Ho et al. \cite{ho2011multiscale}, this prior has the disadvantage of either being too expressive or not expressive enough depending on its parameterization. Regardless, we show the marginalization of this case in Appendix \ref{adx:level_mixtures}.
In this subsection, however, we focus on the infinite case using the stick breaking process as defined in our model. To this end, we use the multinomial-stick conjugacy to obtain a stick breaking posterior which is used as the prior for the level indicators later on. The posterior is defined as follows:
\begin{align}
    \mathbbm{P}(\mathbf{a}_i \ | \ \mathbf{A}_{-i}, \mathbf{Z}, \mu, \sigma) & = \dfrac{\mathbbm{P}(\mathbf{Z} \ | \ \mathbf{A}, \mu, \sigma) \mathbbm{P}(\mathbf{a}_i \ | \ \mathbf{A}_{-i}, \mu, \sigma)}{\bigintssss_{\mathbf{a}_i} \mathbbm{P}(\mathbf{Z} \ | \ \mathbf{A}, \mu, \sigma) \mathbbm{P}(\mathbf{a}_i \ | \ \mathbf{A}_{-i}, \mu, \sigma) \ \text{d}\mathbf{a}_i} \nonumber \\
    & = \dfrac{\text{Multinomial}(\mathbf{a}_i)\text{Stick}(\mu, \sigma)}{\bigintssss_{\mathbf{a}_i} \text{Multinomial}(\mathbf{a}_i)\text{Stick}(\mu, \sigma) \ \text{d}\mathbf{a}_i } \label{eqn:level_mixtures}
\end{align}
Where $\mathbbm{P}(\mathbf{Z} \ | \ \mathbf{A}, \mu, \sigma)$ and $\mathbbm{P}(\mathbf{a}_i \ | \ \mathbf{A}_{-i}, \mu, \sigma)$ are the likelihood and prior of level memberships, respectively. We use the definitions from our model and replace these with the multinomial and stick breaking distributions. Before proceeding, we define
$\mathbf{z}_{i*} = \{ z_{x \leftrightarrow y} \in \mathbf{Z} : x = i \lor y = i\}$ representing all level indicators for entity $e_i$. In this notation $z_{i \leftrightarrow j} := z_{i \rightarrow j} \lor z_{i \leftarrow j}$ is used as shorthand for any level indicator relating entity $e_i$ with entity $e_j$ regardless of which entities are taking on the sender and receiver roles.
This allows for defining two helper variables, $\#^{\mathbf{z}_{i*}=l}$ and $\#^{\mathbf{z}_{i*}>l}$, to indicate the number of indicators in $\mathbf{z}_{i*}$ at and below level $l$ in the hierarchy, respectively:
\begin{align}
    \#^{\mathbf{z}_{i*}=l} = \Big|\Big\{ z_{i \leftrightarrow j} \in \mathbf{z}_{i*} : z_{i \leftrightarrow j} = l \Big\}\Big| \nonumber \\
    \#^{\mathbf{z}_{i*}>l} = \Big|\Big\{ z_{i \leftrightarrow j} \in \mathbf{z}_{i*} : z_{i \leftrightarrow j} > l \Big\}\Big| \label{eqn:level_membership_marginalization_definitions}
\end{align}
With these definitions in place, we can derive the stick breaking posterior of level memberships:
\begin{align}
\mathbbm{P}&( \mathbf{a}_i \ | \ \mathbf{A}_{-i}, \mathbf{Z}, \mu, \sigma) \nonumber \\
    & \overset{\mathrm{(1)}}{=} \dfrac{\Bigg( \dfrac{\Gamma(\sum\limits_{l=1}^{\infty} \#^{\mathbf{z}_{i*}=l} + 1))}{\prod\limits_{l=1}^{\infty}\Gamma(\#^{\mathbf{z}_{i*}=l} + 1))} \prod\limits_{l=1}^\infty a_i^l \Bigg) \Bigg( \sum\limits_{l=1}^\infty  v_l  \prod\limits_{k=1}^{l-1} (1 - v_k) \mathbbm{I}(a_i^l = l) \Bigg)}{\bigint_{\mathbf{V}} \Bigg( \dfrac{\Gamma(\sum\limits_{l=1}^{\infty} \#^{\mathbf{z}_{i*}=l} + 1))}{\prod\limits_{l=1}^{\infty}\Gamma(\#^{\mathbf{z}_{i*}=l} + 1))} \prod\limits_{l=1}^\infty a_i^l \Bigg) \Bigg( \sum\limits_{l=1}^\infty  v_l  \prod\limits_{k=1}^{l-1} (1 - v_k) \mathbbm{I}(a_i^l = l) \Bigg) \text{d}^\infty \mathbf{V}}\nonumber \\
    & \overset{\mathrm{(2)}}{=} \dfrac{\Bigg( \dfrac{\Gamma(\sum\limits_{l=1}^{\infty} \#^{\mathbf{z}_{i*}=l} + 1))}{\prod\limits_{l=1}^{\infty}\Gamma(\#^{\mathbf{z}_{i*}=l} + 1))}  \prod\limits_{l=1}^\infty \big(v_l \prod\limits_{k=1}^{l-1}(1-v_k)\big)^{\#^{\mathbf{z}_{i*}=l}} \Bigg) \Bigg(  \text{Beta}(\mu\sigma, (1-\mu)\sigma) \prod\limits_{k=1}^{l-1} \text{Beta}((1-\mu)\sigma, \mu\sigma) \Bigg)}{\bigint_{\mathbf{V}} \Bigg( \dfrac{\Gamma(\sum\limits_{l=1}^{\infty} \#^{\mathbf{z}_{i*}=l} + 1))}{\prod\limits_{l=1}^{\infty}\Gamma(\#^{\mathbf{z}_{i*}=l} + 1))}  \prod\limits_{l=1}^\infty \big(v_l \prod\limits_{k=1}^{l-1}(1-v_k)\big)^{\#^{\mathbf{z}_{i*}=l}} \Bigg) \Bigg(  \text{Beta}(\mu\sigma, (1-\mu)\sigma) \prod\limits_{k=1}^{l-1} \text{Beta}((1-\mu)\sigma, \mu\sigma) \Bigg) \text{d}^\infty \mathbf{V}}\nonumber \\
    & \overset{\mathrm{(3)}}{=} \dfrac{\Bigg( \dfrac{\Gamma(\sum\limits_{l=1}^{\infty} \#^{\mathbf{z}_{i*}=l} + 1))}{\prod\limits_{l=1}^{\infty}\Gamma(\#^{\mathbf{z}_{i*}=l} + 1))} v_l^{\#^{\mathbf{z}_{i*}=l}} (1 - v_l)^{\#^{\mathbf{z}_{i*}>l}} \prod\limits_{k=1}^{l-1}v_k^{\#^{\mathbf{z}_{i*}>k}}(1 - v_k)^{\#^{\mathbf{z}_{i*}=k}} \Bigg)}{\bigint_{\mathbf{V}} \Bigg( \dfrac{\Gamma(\sum\limits_{l=1}^{\infty} \#^{\mathbf{z}_{i*}=l} + 1))}{\prod\limits_{l=1}^{\infty}\Gamma(\#^{\mathbf{z}_{i*}=l} + 1))} v_l^{\#^{\mathbf{z}_{i*}=l}} (1 - v_l)^{\#^{\mathbf{z}_{i*}>l}} \prod\limits_{k=1}^{l-1}v_k^{\#^{\mathbf{z}_{i*}>k}}(1 - v_k)^{\#^{\mathbf{z}_{i*}=k}} \Bigg) }\nonumber \\
    & \hspace{6.8cm} \dfrac{\Bigg( \dfrac{v_l^{\mu\sigma - 1}(1 - v_l)^{(1-\mu)\sigma - 1}}{\text{B}(\mu\sigma , (1 - \mu) \sigma)} \prod\limits_{k=1}^{l-1} \dfrac{v_k^{(1 - \mu)\sigma - 1}(1 - v_k)^{\mu\sigma - 1}}{\text{B}((1 - \mu)\sigma, \mu\sigma)}  \Bigg)}{\Bigg( \dfrac{v_l^{\mu\sigma - 1}(1 - v_l)^{(1-\mu)\sigma - 1}}{\text{B}(\mu\sigma , (1 - \mu) \sigma)} \prod\limits_{k=1}^{l-1} \dfrac{v_k^{(1 - \mu)\sigma - 1}(1 - v_k)^{\mu\sigma - 1}}{\text{B}((1 - \mu)\sigma, \mu\sigma)} \Bigg) \text{d}^\infty \mathbf{V}} \nonumber \\
    & \overset{\mathrm{(4)}}{=} \dfrac{  v_l^{\#^{\mathbf{z}_{i*}=l} + \mu\sigma - 1}(1 - v_l)^{\#^{\mathbf{z}_{i*}>l} + (1 -\mu)\sigma - 1} \prod\limits_{k=1}^{l-1} v_k^{\#^{\mathbf{z}_{i*}>k} + (1 - \mu)\sigma - 1}(1 - v_k)^{\#^{\mathbf{z}_{k*}=k} + \mu\sigma - 1}}{\bigintss_{\mathbf{V}}  v_l^{\#^{\mathbf{z}_{i*}=l} + \mu\sigma - 1}(1 - v_l)^{\#^{\mathbf{z}_{i*}>l} + (1 -\mu)\sigma - 1} \prod\limits_{k=1}^{l-1} v_k^{\#^{\mathbf{z}_{i*}>k} + (1 - \mu)\sigma - 1}(1 - v_k)^{\#^{\mathbf{z}_{k*}=k} + \mu\sigma - 1} \text{d}^\infty \mathbf{V}} \nonumber \\
    & \overset{\mathrm{(5)}}{=} \dfrac{  v_l^{\#^{\mathbf{z}_{i*}=l} + \mu\sigma - 1}(1 - v_l)^{\#^{\mathbf{z}_{i*}>l} + (1 -\mu)\sigma - 1} \prod\limits_{k=1}^{l-1} v_k^{\#^{\mathbf{z}_{i*}>k} + (1 - \mu)\sigma - 1}(1 - v_k)^{\#^{\mathbf{z}_{k*}=k} + \mu\sigma - 1}}{\text{B}(\#^{\mathbf{z}_{i*}=l} + \mu\sigma, \#^{\mathbf{z}_{i*}>l} + (1 -\mu)\sigma) \prod\limits_{k=1}^{l-1} \text{B}(\#^{\mathbf{z}_{i*}=k} + (1 -\mu)\sigma, \#^{\mathbf{z}_{k*}=k} + \mu\sigma - 1)} \nonumber \\
    & = \text{Beta}(\#^{\mathbf{z}_{i*}=l} + \mu\sigma, \#^{\mathbf{z}_{i*}>l} + (1 - \mu)\sigma) \prod\limits_{k=1}^{l-1} \text{Beta}( \#^{\mathbf{z}_{i*}>k} + (1 - \mu)\sigma,\#^{\mathbf{z}_{i*}=k} + \mu\sigma) \nonumber \\
    & =  \sum\limits_{l=1}^\infty v_l \prod\limits_{k=1}^{l-1} (1 - v_k) \ | \ \mathbf{z}_{i*} \nonumber \\
    & = \text{Stick}(\mu\sigma, (1 - \mu)\sigma) \ | \ \mathbf{z}_{i*} \label{eqn:marginalized_level_mixtures}
\end{align}
Where (1) is an application of the definitions of the Multinomial and stick breaking distributions as per Equations \ref{eqn:multinomial} and \ref{eqn:stick} to the posterior as per Equation \ref{eqn:level_mixtures}. (2) redefines the likelihood in terms of the Beta samples of the stick breaking process and the prior leverages the mirror symmetry property of the Beta function. The summation and indicator functions are both removed at this stage to enhance readability. (3) rearranges the likelihood and substitutes the probability density function of the Beta distribution as per Equation \ref{eqn:beta}. (4) involves cancelling out terms in the numerator and denominator which is a constant with respect to the integration. (5) utilizes the integral form of the Beta function as per Equation \ref{eqn:normalizingconstant} and the remainder of the derivation merely reverses the definitions applied earlier to arrive at the definition of the stick breaking process. We use the notation $ | \ \mathbf{z}_{i*}$ to denote that the distribution preceding the notation is conditioned on $\mathbf{z}_{i*}$.

\subsubsection{Sampling Entity Paths}\label{subsubsec:paths}

Entity paths are one of the two variables which remain after collapsing the Gibbs sampler and must therefore be sampled directly.
To sample a path for entity $e_i$, it must first be removed from the hierarchy, thereby allowing for its full conditional distribution to be obtained. The set of paths after having removed path $\mathbf{p}_i$ is denoted as $\mathbf{P}_{-i}$. We derive the posterior distribution of $\mathbf{p}_{i}$ by applying Bayes' theorem: 
\begin{align}
    \mathbbm{P}(\mathbf{p}_{i} \ | \ \mathbf{P}_{-i} , \mathbf{G} , \mathbf{Z} , \gamma , \lambda , \eta ) & = \dfrac{\mathbbm{P}(\mathbf{G}_{i*} \ | \ \mathbf{G}_{-(i*)} , \mathbf{P}, \mathbf{Z}, \gamma , \lambda , \eta ) \mathbbm{P}(\mathbf{p}_i \ | \ \mathbf{P}_{-i} , \gamma )}{\bigintssss_{\mathbf{p}_{i}} \mathbbm{P}(\mathbf{G}_{i*} \ | \ \mathbf{G}_{-(i*)} , \mathbf{P}, \mathbf{Z}, \gamma ,  \lambda , \eta ) \mathbbm{P}(\mathbf{p}_i \ | \ \mathbf{P}_{-i} , \gamma )  \ \text{d}\mathbf{p}_i} \nonumber \\
    & \propto \mathbbm{P}(\mathbf{G}_{i*} \ | \ \mathbf{G}_{-(i*)} , \mathbf{P}, \mathbf{Z}, \gamma ,  \lambda , \eta ) \mathbbm{P}(\mathbf{p}_i \ | \ \mathbf{P}_{-i} , \gamma ) \label{eqn:path_posterior}
\end{align}
Where $\mathbf{G}_{i*} = \{ g_{xyz} \in \mathbf{G} : i = x \ \lor \ i = y \}$ denotes all the triples in the knowledge graph that depend on path $\mathbf{p}_i$ and $\mathbf{G}_{-(i*)} = \mathbf{G} \setminus \mathbf{G}_{i*}$ is its complement. The integral form of the marginal distribution for generating the data is a normalizing constant for the posterior distribution. Calculating this integral is not necessary and we can instead sample paths from its proportional distribution as per Equation \ref{eqn:path_posterior}.
The prior for sampling an entity path, $\mathbbm{P}(\mathbf{p}_i \ | \ \mathbf{P}_{-i} , \gamma )$, is obtained from the nCRP. We note that due to the iterative nature of the Gibbs sampler, a path for entity $e_i$ may already exist in the hierarchy from a previous iteration. As such, it must first be removed, hence the conditioning on $\mathbf{P}_{-i}$.
We use $\mathbf{p}_i = t_q$ as a shorthand to indicate the path that terminates at community $t_q$, in other words $\mathbf{p}_i = t_q$ if $p^L_i = t_q$. Thus, the prior is calculated as follows:
\begin{align}
    \mathbbm{P}(\mathbf{p}_i = t_q \ | \ \mathbf{P}_{-i} , \gamma) & = \mathbbm{P}(p^L_i = t_q  \ | \ \mathbf{P}_{-i} , \gamma ) \nonumber \\
    & = \mathbbm{E} \Big[ (\mathbbm{I}(p^L_i = t_q) \ \big| \ \mathbf{P}_{-i} , \gamma ) \Big] \nonumber \\  
    & =  \mathbbm{P}(p^L_i = t_q \ | \ \mathbf{p}^{1 \ : \ L-1}_i, \mathbf{P}_{-i}, \gamma) \prod_{l = 1}^{L - 1} \mathbbm{P}(p^l_i = t_q^l \ | \ \mathbf{p}^{1 \ : \ l-1}_{i}, \mathbf{P}_{-i} , \gamma) \label{eqn:paths_prior}
\end{align}
Where $t_q^l$ is the ancestor community of community $t_q$ at level $l$. Equation \ref{eqn:paths_prior} requires the distribution for sampling a community conditioned on a partially sampled path. Recall that this is defined by the nCRP in Equation \ref{eqn:ncrp} and can be adapted here.
Specifically, we calculate the probability of taking community $t_q$ on level $l$ having already sampled its path up to level $l - 1$ as:
\begin{align}
\mathbbm{P}(p^l_i = t_q \ | \  \mathbf{p}^{1 \ : \ l-1}_i, \mathbf{P}_{-i}, \gamma) & = \begin{dcases} 
  \dfrac{\#_{-i}^{t_q}}{\#_{-i}^{t^{l-1}_q} + \gamma} & t_q \in \mathcal{T}_{-i} \\
  \dfrac{\gamma}{\#_{-i}^{t^{l-1}_q} + \gamma} & t_q \notin \mathcal{T}_{-i} \\
\end{dcases} 
\end{align}
Where $\#_{-i}^{t_q}$ extends the notation defined earlier to indicate the number of entities that have gone through community $t_q$ in the hierarchy with path $\mathbf{p}_i$ removed.
$\mathcal{T}_{-i}$ indicates all the communities in the hierarchy after path $\mathbf{p}_i$ has been removed. 
We note that this process requires the sampling of a path to start at the root community and proceed sequentially to a leaf community.
Having obtained the prior, it's necessary to update the belief about the posterior with the data via the likelihood.
The likelihood given a sampled path, $\mathbbm{P}(\mathbf{G}_{i*} \ | \ \mathbf{G}_{-(i*)}, \mathbf{P}, \mathbf{Z}, \gamma, \lambda, \eta )$, is defined with the help of the following helper variables: 
\begin{align}
    \mathbf{C}_{i*} & = \Big\{ c_{pqr} \in \mathbf{C} \ : \ (\exists g_{xyz} \in \mathbf{G}_{i*} : \Psi(x,y,z) = c_{pqr}) \Big\} \nonumber \\ 
    \#_{-i}^{c_{pqr}=1} & = \Big|\Big\{g_{xyz} \in \mathbf{G}_{-(i*)} : \Psi(x,y,z) = c_{pqr} \land g_{xyz} = 1 \Big\} \Big| \nonumber \\ 
    \#_{-i}^{c_{pqr}=0} & = \Big|\Big\{g_{xyz} \in \mathbf{G}_{-(i*)} : \Psi(x,y,z) = c_{pqr} \land g_{xyz} = 0 \Big\}\Big| \nonumber \\ 
    \#_{i}^{c_{pqr}=1} & = \Big|\Big\{g_{xyz} \in \mathbf{G}_{i*} : \Psi(x,y,z) = c_{pqr} \land g_{xyz} = 1 \Big\}\Big| \nonumber \\ 
    \#_{i}^{c_{pqr}=0} & = \Big|\Big\{g_{xyz} \in \mathbf{G}_{i*} : \Psi(x,y,z) = c_{pqr} \land g_{xyz} = 0 \Big\}\Big| \label{eqn:path_helper}
\end{align}
The definitions above capture the following: $\mathbf{C}_{i*}$ is the set of communities dependant on an interaction in $\mathbf{G}_{i*}$;
$\#_{-i}^{c_{pqr}=1}$ and $\#_{i}^{c_{pqr}=1}$ are the counts of existing entity relations from communities $t_p$ to $t_q$ in $\mathbf{G}_{-(i)}$ and $\mathbf{G}_{i}$, respectively;
and $\#_{-i}^{c_{pqr}=0}$ and $\#_{i}^{c_{pqr}=0}$ are the counts of non-existing entity relations from communities $t_p$ to $t_q$ in $\mathbf{G}_{-(i)}$ and $\mathbf{G}_{i}$, respectively.
In the discrete space, the likelihood is understood as the joint probability of generating the data as per a probability mass function. In our model, the data is obtained by drawing from the Bernoulli distribution conditioned on community relations. Recall that these parameters are marginalized out of our model and thus never sampled directly. As such, in order to calculate the likelihood we must integrate with respect to the community relations. This is possible by leveraging Equation \ref{eqn:marginalized_community_interactions} which allows us to obtain a closed form solution:
\begin{align}
    \mathbbm{P}( & \mathbf{G}_{i*} \ |  \ \mathbf{G}_{-(i*)} , \mathbf{P}, \mathbf{Z}, \gamma , \lambda , \eta ) \nonumber \\
     & = \prod_{c_{pqr} \in \mathbf{C}_{i*}} \bigintssss_{c_{pqr}} \mathbbm{P}(\mathbf{G}_{i*} \ | \ \mathbf{G}_{-(i*)} , \mathbf{C}, \mathbf{P}, \mathbf{Z}, \gamma , \mu , \sigma , \lambda , \eta ) \mathbbm{P}(c_{pqr} \ | \ \mathbf{C}_{-(pqr)}, \mathbf{G}_{-(i*)}, \mathbf{P}, \mathbf{Z}, \lambda, \eta) \ \text{d}c_{pqr} \nonumber \\
    & \overset{\mathrm{(1)}}{=} \prod_{c_{pqr} \in \mathbf{C}_{i*}} \bigintssss_{c_{pqr}} \Bigg( \prod_{g_{xyz} \in \mathbf{G}_{i*}} \text{Bernoulli}(c_{pqr}, 1 - c_{pqr}) \Bigg) \Bigg( \text{Beta}(\#_{-i}^{c_{pqr}=1} + \lambda, \#_{-i}^{c_{pqr}=0} + \eta) \Bigg) \ \text{d}c_{pqr} \nonumber \\
    & \overset{\mathrm{(2)}}{=} \prod_{c_{pqr} \in \mathbf{C}_{i*}} \bigintssss_{c_{pqr}} \Bigg( c^{\#_{i}^{c_{pqr}=1}}_{pqr}(1 - c_{pqr})^{\#_{i}^{c_{pqr}=0}} \Bigg) \Bigg(  \dfrac{c_{pqr}^{\#_{-i}^{c_{pqr}=1} + \lambda - 1}(1-c_{pqr})^{\#_{-i}^{c_{pqr}=0} + \eta - 1}}{\text{B}(\#_{-i}^{c_{pqr}=1} + \lambda, \#_{-i}^{c_{pqr}=0} + \eta)} \Bigg)  \ \text{d}c_{pqr} \nonumber \\ 
    & = \prod_{c_{pqr} \in \mathbf{C}_{i*}} \dfrac{1}{\text{B}(\#_{-i}^{c_{pqr}=1} + \lambda, \#_{-i}^{c_{pqr}=0} + \eta)} \bigintssss_{c_{pqr}} c^{\#_{i}^{c_{pqr}=1} + \#_{-i}^{c_{pqr}=1} + \lambda - 1}_{pqr}(1 - c_{pqr})^{\#_{i}^{c_{pqr}=0} + \#_{-i}^{c_{pqr}=0} + \eta - 1}  \ \text{d}c_{pqr} \nonumber \\
    & \overset{\mathrm{(3)}}{=} \prod_{c_{pqr} \in \mathbf{C}_{i*}} \dfrac{\text{B}(\#_{i}^{c_{pqr}=1} + \#_{-i}^{c_{pqr}=1} + \lambda, \#_{i}^{c_{pqr}=0} + \#_{-i}^{c_{pqr}=0} + \eta )}{\text{B}(\#_{-i}^{c_{pqr}=1} + \lambda, \#_{-i}^{c_{pqr}=0} + \eta)} \nonumber \\
    & \overset{\mathrm{(4)}}{=} \prod_{c_{pqr} \in \mathbf{C}_{i*}} \Bigg( \dfrac{\Gamma(\#_{i}^{c_{pqr}=1} + \#_{-i}^{c_{pqr}=1} + \lambda)\Gamma(\#_{i}^{c_{pqr}=0} + \#_{-i}^{c_{pqr}=0} + \eta)}{\Gamma(\#_{i}^{c_{pqr}=1} + \#_{-i}^{c_{pqr}=1} + \#_{i}^{c_{pqr}=0} + \#_{-i}^{c_{pqr}=0} + \lambda + \eta)} \Bigg) \Bigg( \dfrac{\Gamma(\#_{-i}^{c_{pqr}=1} + \#_{-i}^{c_{pqr}=0} + \lambda + \eta)}{\Gamma(\#_{-i}^{c_{pqr}=1} + \lambda) \Gamma(\#_{-i}^{c_{pqr}=0} + \eta)} \Bigg) \label{eqn:paths_likelihood}
\end{align}
In the derivation above: (1) the prior probability of drawing $c_{pqr}$ is obtained from Equation \ref{eqn:marginalized_community_interactions}; (2) utilizes the definitions as per Equations \ref{eqn:bernoulli} and \ref{eqn:beta} as well as the helper variables introduced in Equation \ref{eqn:path_helper}; (3) leverages the integral form of the Beta function as per Equation \ref{eqn:normalizingconstant}; and (4) expands the Beta function to its Gamma formulation as per Equation \ref{eqn:beta}. Having derived the prior and likelihood in Equations \ref{eqn:paths_prior} and \ref{eqn:paths_likelihood}, respectively, it is possible to sample from Equation \ref{eqn:path_posterior} to obtain entity paths in our model. The time complexity of sampling one such path is $\mathcal{O}(|\mathcal{E}|^2 |\mathcal{R}| L)$. This is due to the fact that it's necessary to obtain a sampling probability for all potential paths in the hierarchy, which has a bound of $|\mathcal{E}| L$ in the case where each entity takes a unique path. For each of these potential paths, the iteration through all $|\mathcal{E}|$ entities and $|\mathcal{R}|$ predicates is required to determine the effect on the likelihood selecting such a path would have.

\subsubsection{Sampling Level Indicators}\label{subsubsec:levels}

Level indicators are drawn from the multinomial distribution conditioned on level memberships. Recall that level memberships were marginalized over in our inference scheme using the multinomial-stick conjugacy and are thus never sampled directly. Nevertheless, we draw them indirectly when computing the prior for level indicators. As with sampling paths, we obtain the distribution proportional to that of level indicators by Bayes' rule. In what follows, we provide the derivation for the posterior of $z_{i \rightarrow j}$ and note that given its structural symmetry, $z_{i \leftarrow j}$ is derived analogously. The posterior distribution of $z_{i \rightarrow j}$ is expressed as:
\begin{align}
    \mathbb{P}(z_{i \rightarrow j} \ | \ \mathbf{Z}_{-(i \rightarrow j)}, \mathbf{G}, \mathbf{P}, \gamma, \mu, \sigma, \lambda, \eta) & =  \dfrac{\mathbb{P}(\mathbf{g}_{ij*} \ | \ \mathbf{G}_{-(ij*)}, \mathbf{P}, \mathbf{Z}, \lambda, \eta)\mathbbm{P}(z_{i \rightarrow j} \ | \ \mathbf{Z}_{-(i \rightarrow j)}, \mu, \sigma)}{\bigintssss_{z_{i \rightarrow j}} \mathbb{P}(\mathbf{g}_{ij*} \ | \ \mathbf{G}_{-(ij*)}, \mathbf{P}, \mathbf{Z}, \lambda, \eta)\mathbbm{P}(z_{i \rightarrow j} \ | \ \mathbf{Z}_{-(i \rightarrow j)}, \mu, \sigma)  \  \ \text{d}z_{i \rightarrow j}}  \nonumber \\
    & \propto  \mathbb{P}(\mathbf{g}_{ij*} \ | \ \mathbf{G}_{-(ij*)}, \mathbf{P}, \mathbf{Z}, \lambda, \eta) \mathbbm{P}(z_{i \rightarrow j} \ | \ \mathbf{Z}_{-(i \rightarrow j)}, \mu, \sigma) \label{eqn:level_indicators_sample}
\end{align}
Where $\mathbf{g}_{ij*} = \{g_{xyz} \in \mathbf{G} : i = x \land j = y \}$ denotes the vector of relations in $\mathbf{G}$ from entity $e_i$ to $e_j$ across all predicates, $\mathbf{G}_{-(ij*)} = \mathbf{G} \setminus \mathbf{g}_{ij*}$ is its complement, and $\mathbf{Z}_{-(i \rightarrow j)}$ is all the level indicators excluding $z_{i \rightarrow j}$.
The prior probability for sampling levels, $\mathbbm{P}(z_{i \rightarrow j} \ | \ \mathbf{Z}_{-(i \rightarrow j)}, \mu, \sigma)$, is drawn from the derived posterior of level memberships in Equation \ref{eqn:marginalized_level_mixtures}. 
We follow Blei et al. \cite{blei2010nested} and use the law of total expectations to obtain the probability of $z_{i \rightarrow j}$ realizing level $l$ as the expectation of the size of the stick broken off at the $l^{th}$ break. To do this we define two variables which may be seen as the directed extensions of those introduced in Equation \ref{eqn:level_membership_marginalization_definitions}:
\begin{align}
    \#^{\mathbf{Z}_{-(i \rightarrow j)} = l} = \Big|\Big\{ z_{i \rightarrow j} \in \mathbf{Z}_{-(i \rightarrow j)} : z_{i \rightarrow j} = l \Big\}\Big| \nonumber \\
    \#^{\mathbf{Z}_{-(i \rightarrow j)} > l} = \Big|\Big\{ z_{i \rightarrow j} \in \mathbf{Z}_{-(i \rightarrow j)} : z_{i \rightarrow j} > l \Big\}\Big| \label{eqn:level_indicator_variables}
\end{align}
With these variables in place, we obtain the prior distribution as follows:
\begin{align}
    \mathbbm{P}(z_{i \rightarrow j} = l \ | \ \mathbf{Z}_{-(i \rightarrow j)}, \mathbf{G}_{-(ij*)}, \mathbf{P}, \mu, \sigma) & =  \mathbbm{E} \Big[ \mathbbm{I} ( z_{i \rightarrow j} = l ) \ \big| \ \mathbf{Z}_{-(i \rightarrow j)}, \mu, \sigma \Big] \nonumber \\
    & \overset{\mathrm{(1)}}{=} \mathbbm{E} \Bigg[ \mathbbm{E} \Bigg[ \mathbbm{I} ( z_{i \rightarrow j} = l ) \ \big| \  v_1, v_2, ..., v_l, \mathbf{Z}_{-(i \rightarrow j)}, \mu, \sigma \Bigg] \Bigg] \nonumber \\ 
     & \overset{\mathrm{(2)}}{=} \mathbbm{E} \Bigg[ \sum\limits_{m=1}^\infty v_l \prod\limits_{k=1}^{l-1} (1 - v_k) \mathbbm{I}(m = l)  \ \big| \  \mathbf{Z}_{-(i \rightarrow j)}, \mu, \sigma \Bigg]  \nonumber \\
    & = \mathbbm{E} \Big[ v^l  \ | \  \mathbf{Z}_{-(i \rightarrow j)}, \mu, \sigma \Big]  \prod_{k = 1}^{l - 1} \mathbbm{E} \Big[ 1 - v^k \ \big| \ \mathbf{Z}_{-(i \rightarrow j)}, \mu, \sigma \Big] \nonumber \\
    & \overset{\mathrm{(3)}}{=} \dfrac{\mu\sigma + \#^{\mathbf{Z}_{-(i \rightarrow j)} = l}}{\sigma + \#^{\mathbf{Z}_{-(i \rightarrow j)} = l} + \#^{\mathbf{Z}_{-(i \rightarrow j)} > l}} \prod\limits_{k=1}^{l-1}\dfrac{(1 - \mu)\sigma + \#^{\mathbf{Z}_{-(i \rightarrow j)} > k}}{\sigma + \#^{\mathbf{Z}_{-(i \rightarrow j)} = k} + \#^{\mathbf{Z}_{-(i \rightarrow j)} > k}} \label{eqn:level_indicator_prior}
\end{align}
Where (1) is derived by the application of the law of total expectation; (2) is obtained from the probability of drawing level $l$ from the stick breaking process conditioned on the successive draws from the Beta distribution, denoted $v_1, v_2, ..., v_l$, as per Equation \ref{eqn:stick}; and (3) is the expected value of drawing from the Beta distribution conditioned on $\mathbf{Z}_{-(i \rightarrow j)}$ as per the level membership posterior obtained in Equation \ref{eqn:marginalized_level_mixtures}.
The likelihood, $\mathbb{P}(\mathbf{g}_{ij*} \ | \ \mathbf{G}_{-(ij*)}, \mathbf{P}, \mathbf{Z}, \lambda, \eta)$, is obtained analogously to entity paths. To aid in this derivation, we define two constants as follows: 
\begin{align}
    \#^{c_{pqr}=1}_{-(ijr)} = \Big|\Big\{g_{xyz} \in \mathbf{G}_{-(ijr)} \  : \ \Psi(x,y,z) = c_{pqr} \land g_{xyz} = 1 \Big\}\Big| \nonumber \\
    \#^{c_{pqr}=0}_{-(ijr)} = \Big|\Big\{g_{xyz} \in \mathbf{G}_{-(ijr)} \  : \  \Psi(x,y,z) = c_{pqr} \land g_{xyz} = 0 \Big\}\Big|
\end{align}
Such that $\#^{c_{pqr}=1}_{-(ijr)}$ and $\#^{c_{pqr}=0}_{-(ijr)}$ capture the number of existing and non-existing relations from communities $t_p$ to $t_q$ not including $g_{ijr}$. With these in place, we can derive the level indicator likelihood. This process is analogous to the one for entity paths in that we use the Bernoulli distribution for model output and integrate over community relations: 
\begin{align}
    \mathbb{P}(\mathbf{g}_{ij*} &] \ | \  \mathbf{G}_{-(ij*)},  \mathbf{P}, \mathbf{Z}, \lambda, \eta) \nonumber \\
    & = \bigintssss_{c_{pqr}} \mathbbm{P}(\mathbf{g}_{ij*} \ | \ \mathbf{G}_{-(ij*)} , \mathbf{C}, \mathbf{P}, \mathbf{Z}, \gamma , \mu , \sigma , \lambda , \eta ) \mathbbm{P}(c_{pqr} \ | \ \mathbf{C}_{-(pqr)}, \mathbf{G}_{-(ij*)}, \mathbf{P}, \mathbf{Z}, \lambda, \eta) \ \text{d}c_{pqr} \nonumber \\
    & \overset{\mathrm{(1)}}{=} \bigintssss_{c_{pqr}} \prod_{g_{ijr} \in \mathbf{g}_{ij*}} \mathbbm{P}(g_{ijr} \ | \ \mathbf{G}_{-(ijr)} , \mathbf{C}, \mathbf{P}, \mathbf{Z}, \gamma , \mu , \sigma , \lambda , \eta ) \mathbbm{P}(c_{pqr} \ | \ \mathbf{C}_{-(pqr)}, \mathbf{G}_{-(ijr)}, \mathbf{P}, \mathbf{Z}, \lambda, \eta) \ \text{d}c_{pqr} \nonumber \\
    & \overset{\mathrm{(2)}}{=} \bigintssss_{c_{pqr}} \prod_{g_{ijr} \in \mathbf{g}_{ij*}} \Bigg( \text{Bernoulli}(c_{pqr}, 1 - c_{pqr}) \Bigg) \Bigg( \text{Beta}(\#^{c_{pqr}=1}_{-(ijr)} + \lambda, \#^{c_{pqr}=0}_{-(ijr)} + \eta) \Bigg) \ \text{d}c_{pqr} \nonumber \\
    & = \bigintssss_{c_{pqr}} \prod_{g_{ijr} \in \mathbf{g}_{ij*}} \Bigg( c^{g_{ijr}}_{pqr}(1 - c_{pqr})^{1 - g_{ijr}} \Bigg) \Bigg(  \dfrac{c_{pqr}^{\#^{c_{pqr}=1}_{-(ijr)} + \lambda - 1}(1-c_{pqr})^{\#^{c_{pqr}=0}_{-(ijr)} + \eta - 1}}{\text{B}(\#^{c_{pqr}=1}_{-(ijr)} + \lambda, \#^{c_{pqr}=0}_{-(ijr)} + \eta)} \Bigg)  \ \text{d}c_{pqr} \nonumber \\ 
    & = \prod_{g_{ijr} \in \mathbf{g}_{ij*}} \dfrac{1}{\text{B}(\#^{c_{pqr}=1}_{-(ijr)}+ \lambda, \#^{c_{pqr}=0}_{-(ijr)} + \eta)} \bigintssss_{c_{pqr}} c^{g_{ijr} + \#^{c_{pqr}=1}_{-(ijr)}+ \lambda - 1}_{pqr}(1 - c_{pqr})^{(1 - g_{ijr}) + \#^{c_{pqr}=0}_{-(ijr)} + \eta - 1}  \ \text{d}c_{pqr} \nonumber \\
    & \overset{\mathrm{(3)}}{=} \prod_{g_{ijr} \in \mathbf{g}_{ij*}} \dfrac{\text{B}(\#^{c_{pqr}=1}_{-(ijr)} + g_{ijr} + \lambda, \#^{c_{pqr}=0}_{-(ijr)} + (1 - g_{ijr}) + \eta)}{\text{B}(\#^{c_{pqr}=1}_{-(ijr)} + \lambda, \#^{c_{pqr}=0}_{-(ijr)} + \eta)} \nonumber \\
    & \overset{\mathrm{(4)}}{=} \prod_{g_{ijr} \in \mathbf{g}_{ij*}}  \dfrac{\Gamma(\#^{c_{pqr}=1}_{-(ijr)}  + g_{ijr} + \lambda)\Gamma(\#^{c_{pqr}=0}_{-(ijr)} + (1 - g_{ijr}) + \eta )\Gamma(\#^{c_{pqr}=1}_{-(ijr)}  + \#^{c_{pqr}=0}_{-(ijr)} + \lambda + \eta)}{\Gamma(\#^{c_{pqr}=1}_{-(ijr)}  + \#^{c_{pqr}=0}_{-(ijr)} + 1 + \lambda + \eta)\Gamma(\#^{c_{pqr}=1}_{-(ijr)}  + \lambda)\Gamma(\#^{c_{pqr}=0}_{-(ijr)} + \eta)}  \nonumber \\
     & \overset{\mathrm{(5)}}{=} \prod_{g_{ijr} \in \mathbf{g}_{ij*}} \dfrac{g_{ijr}(\#^{c_{pqr}=1}_{-(ijr)} + \lambda) + (1-g_{ijr})(\#^{c_{pqr}=0}_{-(ijr)} + \eta)}{\#^{c_{pqr}=1}_{-(ijr)} + \#^{c_{pqr}=0}_{-(ijr)} + \lambda + \eta} \label{eqn:zlikelihood}
\end{align}
Where (1) applies the chain rule of probability; (2) utilizes the prior for $c_{pqr}$ obtained in Equation \ref{eqn:marginalized_community_interactions}; (3) and (4) leverage the integral and Gamma forms the Beta function as per Equations \ref{eqn:normalizingconstant} and \ref{eqn:beta}, respectively; and (5) simplifies the preceding equation for computational reason by eliminating the Gamma function as shown in Appendix \ref{adx:level_likelihood}.
With the prior and likelihood derived in closed form as per Equations \ref{eqn:level_indicator_prior} and \ref{eqn:zlikelihood}, respectively, it's possible to sample level indicators via Equation \ref{eqn:level_indicators_sample}. The time complexity of sampling a level indicator is $\mathcal{O}(|\mathcal{R}|L)$ due to the $|\mathcal{R}|$ calculations that need to be performed at each of the $L$ levels in the hierarchy.

\subsubsection{Sampling Procedure}

Having marginalized out community relations and level memberships as well as derived the sampling equations for entity paths and level indicators, it is possible it perform collapsed Gibbs sampling by iteratively sampling from the remaining variables' full conditional distributions. This process has a time complexity of $\mathcal{O}(|\mathcal{E}|^2|\mathcal{R}|L + |\mathcal{E}|^3|\mathcal{R}|L)$ for each iteration of the sampler where the former and latter terms are derived from the sampling complexities of the level indicators and entity paths, respectively. This makes the inference scheme infeasible for large-scale datasets.
We respond to this issue by modifying one of the characteristics of collapsed Gibbs sampling, namely that samples are obtained in equal proportions. In its original formulation, one iteration of the sampler samples $|\mathcal{E}|^2$ level indicators and $|\mathcal{E}|$ entity paths. One of the assumptions underlying this process is that the relative importance of all samples is the same.
Such an assumption may be ill-adapted for knowledge graphs which are oftentimes sparse in their adjacency tensors and whose entities exhibit highly imbalanced relation densities. 
In this regard, the placement of highly connected entities will have a disproportionate effect on model likelihood and therefore the induced hierarchy as well. Preferentially sampling these entities may result in faster arrival at a distribution from which we can obtain output samples.
Consider, for instance, a knowledge graph with the entities \texttt{Thing} and \texttt{Henry Ford}. Assuming that \texttt{Thing} has a higher relation density than \texttt{Henry Ford}, its proper placement in the hierarchy may be more critical for model output than \texttt{Henry Ford}.
With this in mind, we propose a stochastic sampling scheme in which samples are drawn for an entity in proportion to their probability of interacting with other entities. Specifically, we introduce a sampling probability, denoted $s_i$ for entity $e_i$, which specifies the chance of sampling a variable for the corresponding entity in an iteration of the collapsed Gibbs sampler. This probability is calculated for each entity as the fraction of entities in the knowledge graph which have fewer relations than itself. Such as formulation ensures that $1 \geq s_i > 0$ which allows $s_i$ to serve as the parameter of a Bernoulli distribution to indicate whether a variable will get sampled in the current iteration of the Gibbs sampler.

After the Gibbs sampler has been burned in, it is necessary to obtain final samples to obtain a hierarchical clustering. We take multiple samples to account for the spread in the posterior distribution. A consequence of this is that samples may differ and need to be aggregated to produce a final result. In this regard, we take the mode over the final samples to arrive at a final hierarchy. The Gibbs sampling procedure is summarized in Algorithm \ref{alg:hb_algorithm}.

\begin{algorithm}[t!]
\caption{Collapsed Gibbs Sampling Procedure for Model Inference}
\label{alg:hb_algorithm}
\textbf{Input:} Knowledge graph adjacency tensor, $\mathbf{G}$; model hyperparameters, $\gamma$, $\mu$, $\sigma$, $\lambda$, and $\eta$; number of iterations, $iters$ 

\textbf{Output:} Paths $\mathbf{P}$; level indicators $\mathbf{Z}$; \newline community relations $\mathbf{C}$ 
\begin{algorithmic}[1]
\STATE Initialize level indicators using Equation \ref{eqn:level_indicator_prior}
\STATE Initialize paths using Equation \ref{eqn:paths_prior}
\FOR {$iter = 1, 2, ..., iters$}
    \STATE Update level indicators using Equation \ref{eqn:level_indicators_sample} if Bernoulli($s_i$)
    \STATE Update paths using Equation \ref{eqn:path_posterior} if Bernoulli($s_i$)
\ENDFOR
\STATE Obtain final level indicators using \ref{eqn:level_indicators_sample}
\STATE Obtain final paths using \ref{eqn:path_posterior}
\end{algorithmic}
\end{algorithm}

\section{Evaluation}\label{sec:evaluation}

The evaluation of our model is split into two parts: quantitative and qualitative. The quantitative evaluation provides objective measures of model performance whereas the qualitative evaluation assesses our model through illustrations and subjective analysis of the results.
For both types of evaluations, our model first had to be inferred before final samples could be drawn. In this regard, we trained our model on three datasets using 200 burn-in samples using hyperparameters chosen by assessing the model's log likelihood. After burn-in, ten final samples were obtained by discarding all but the third of successive samples to account for autocorrelation between samples.  All models we trained to a depth of $L = 4$. Furthermore, the model was trained five times for each dataset to account for stochasticity in the inference process. The implementation of our method as well as the datasets necessary to recreate our evaluation has been made publicly available on GitHub \footnote{\url{www.github.com/mpietrasik/hb}}.

\subsection{Datasets}\label{subsec:hb_datasets}

Our model was evaluated on three datasets: Synthetic Binary Tree, FB15k-237, and WikiData.
What follows is a brief description of each dataset as well as how it was generated.

\subsubsection{Synthetic Binary Tree}

The Synthetic Binary Tree (SBT) dataset was synthetically generated to capture our model's ability to separate communities at the lowest level in the hierarchy.
The generative process first constructed a binary tree with a depth of four, assigned entities to communities, and sampled relations for each entity pair.
All entities were assigned uniformly to communities on the lowest level of the hierarchy, resulting in 25 entities per leaf community.
The sampling probability for each entity pair was determined by the level of their lowest common ancestor. Specifically, sampling probabilities of 0, 0.1, 0.4, and 0.6 were used for levels 0, 1, 2, and 3, respectively. Two entities belonging to the same community have a sampling probability of 1 and are thus always related.
The dataset was generated for two predicates which shared the aforementioned sampling probabilities. We note that even though these probabilities are identical, they do not result in a dataset in which entity relations are identical across predicates.
The generative process yielded a dataset of 55880 triples, 400 entities, and 2 predicates.

\subsubsection{FB15k-237}

The FB15k-237 dataset \cite{toutanova2015observed} is a subset of the FB15k dataset \cite{bordes2013translating}, created by removing redundant and inverse triples.
The original FB15k dataset is in turn a subset of a 2013 version of Freebase, from which triples were queried.
The FB15k-237 dataset is comprised of 272115 triples, 14541 entities, and 237 predicates thus presenting a computation challenge to our model if modelled in whole.
To address this issue, we generated a subset of the data and derived ground truth community labels in an approach inspired by Jain et al. \cite{jain2021embeddings}.
Specifically, entities were mapped to the WordNet taxonomy \cite{miller1995wordnet} through the $sameAs$ predicate, which relates entities from Freebase and YAGO. Triples were then extracted to contain subjects from the sets provided in Zhang et al. \cite{zhang2022hierarchical}. This process yielded a subset of the data containing 103550 triples, 10018 entities, and 190 predicates.
Finally, the subset was reduced even further by extracting only the triples relating to footballers, pianists, journalists, politicians, and scientists as per the identifiers \texttt{/m/05vyk}, \texttt{/m/06q2q}, \texttt{/m/0gl2ny2}, \texttt{/m/0fj9f}, \texttt{/m/0d8qb} on the predicate \texttt{/people/person/profession}.
This final step resulted in a dataset with 2499 triples, 1142 entities, and 79 predicates.

\subsubsection{WikiData}

The WikiData dataset was generated by querying Wikidata for triples relating to people and locations. Specifically, artists and footballers corresponding to WikiData identifiers \texttt{wd:Q1028181} and \texttt{wd:Q937857} respectively were extracted. These entities were then filtered to having been born in cities in four countries: Germany, the United Kingdom, Canada, and the United States of America. Furthermore, the knowledge graph was reduced to the following predicates: instance of, place of birth, citizen of, occupation, country, and located in which are represented by the identifiers \texttt{wdt:P31}, \texttt{wdt:P19}, \texttt{wdt:P27}, \texttt{wdt:P106},  \texttt{wdt:P17}, and \texttt{wdt:P131}, respectively. Finally, the tripleset was further reduced to yield 2525 triples, 716 entities, and 6 predicates.

\subsection{Quantitative Evaluation}\label{subsec:hb_quantitative_evaluation}

\begin{table*}[t!]
\centering
  \caption[Hierarchical blockmodelling ARI and NMI scores on three datasets.]{ARI and MNI scores (mean $\pm$ standard deviation) of our model on the SBT, FB15k-237 and WikiData datasets.}
  \begin{tabular}{l r r r r r r}
    \toprule
     \multicolumn{1}{c}{} & \multicolumn{2}{c}{\textbf{SBT}} & \multicolumn{2}{c}{\textbf{FB15k-237}} & \multicolumn{2}{c}{\textbf{WikiData}}  \\
    \textbf{Method} & \multicolumn{1}{c}{ARI} & \multicolumn{1}{c}{NMI} & \multicolumn{1}{c}{ARI} & \multicolumn{1}{c}{NMI} & \multicolumn{1}{c}{ARI} & \multicolumn{1}{c}{NMI}  \\
    \midrule
   Level 1 & 0.3055 & 0.4855 & 0.5326 & 0.6646 & 0.8411 & 0.7991 \\
    & $\pm 0.0685$ & $\pm 0.1013$ & $\pm 0.1308$ & $\pm 0.0702$ & $\pm 0.2980$ & $\pm 0.1581$  \\
   Level 2 & 0.5895 & 0.7826 & 0.3492 & 0.5083 & $0.8057$  & $0.7232$   \\ 
   & $\pm 0.2826$ & $\pm 0.1434$ & $\pm 0.2044$ & $\pm 0.1175$ &  $\pm 0.2839$ & $\pm 0.1410$ \\
   Level 3 & 0.7279 & 0.8882 & 0.2851 & 0.4329 &  $0.4255$  & $0.5880$   \\ 
   & $\pm0.1656$ & $\pm 0.0621$ & $\pm 0.1993$ & $\pm 0.1030$ &  $\pm 0.2749$ & $\pm 0.1367$ \\
   Level 4 & 0.8337 & 0.9319 & 0.1964 & 0.5334 & $0.3812$  & $0.4980$   \\
   & $\pm 0.1032$ & $\pm 0.0357$ & $\pm 0.0438$ & $\pm 0.0288$ &  $\pm 0.2500$ & $\pm 0.1309$ \\
    \midrule
    Overall & 0.6141 & 0.7721 & 0.3408 & 0.5348 &  $0.6134$  & $0.6521$   \\ 
    & $\pm 0.2577$ & $\pm 0.1988$ & $\pm 0.1867$ & $\pm 0.1145$ & $\pm 0.3341$ & $\pm 0.1770$ \\
    \bottomrule
  \end{tabular}
  \label{tab:hb_evaluation}
\end{table*}

\begin{table*}[t!]
\centering
  \caption[Hierarchical blockmodelling ARI and NMI scores on three datasets against baselines.]{ARI and MNI scores (mean $\pm$ standard deviation) of our model on the SBT, FB15k-237 and WikiData datasets as compared with baseline approaches.}
  \begin{tabular}{l r r r r r r}
    \toprule
     \multicolumn{1}{c}{} & \multicolumn{2}{c}{\textbf{SBT}} & \multicolumn{2}{c}{\textbf{FB15k-237}} & \multicolumn{2}{c}{\textbf{WikiData}}  \\
    \textbf{Method} & \multicolumn{1}{c}{ARI} & \multicolumn{1}{c}{NMI} & \multicolumn{1}{c}{ARI} & \multicolumn{1}{c}{NMI} & \multicolumn{1}{c}{ARI} & \multicolumn{1}{c}{NMI}  \\
    \midrule
   RDF2VEC &  &  &  &  & & \\ 
   \hspace{0.2cm} $k$-means & 0.8060 & 0.8928 & 0.0109 & 0.1402 & 0.2672 & 0.2918 \\
    & $\pm 0.1845$ & $\pm0.0707$ & $\pm0.0929$ & $\pm0.1052$ & $\pm0.1582$ & $\pm0.1040$ \\
   \hspace{0.2cm} Agglomerative & 0.8750 & 0.9317 & 0.0461 & 0.1532 & 0.4674 & 0.5287 \\
    & $\pm 0.1254$ & $\pm0.0575$ & $\pm0.0860$ & $\pm0.1435$ & $\pm0.3281$ & $\pm0.2052$ \\
    \hspace{0.2cm} DBSCAN & 0.5549 & 0.6904 & 0.1468 & 0.2293 & 0.3831 & 0.3698 \\
    & $\pm 0.4576$ & $\pm0.3032$ & $\pm0.1291$ & $\pm0.0561$ & $\pm0.2343$ & $\pm0.0935$ \\
    \hspace{0.2cm} Spectral & 0.6175 & 0.7590 & -0.0014 & 0.0347 & 0.0918 & 0.1021 \\
    & $\pm 0.3540$ & $\pm0.2924$ & $\pm0.0082$ & $\pm0.03129$ & $\pm0.0636$ & $0.0297$ \\
    \midrule
    TransE &  &  &  &  & & \\ 
   \hspace{0.2cm} $k$-means & 0.9851 & 0.9958 & \textbf{0.3559} & 0.4334 & \textbf{0.7427} & 0.6504 \\
    & $\pm 0.0334$ & $\pm0.0066$ & $\pm0.0776$ & $\pm0.1096$ & $\pm0.1953$ & $\pm0.2468$ \\
   \hspace{0.2cm} Agglomerative & \textbf{1.0000} & \textbf{1.0000} & 0.1362 & 0.3107 & 0.3799 & 0.3650 \\
    & $\pm 0.0000$ & $\pm0.0000$ & $\pm0.1379$ & $\pm0.1104$ & $\pm0.4037$ & $\pm0.3780$ \\
    \hspace{0.2cm} DBSCAN & 0.8899 & 0.9665 & 0.2768 & 0.2582 & 0.2418 & 0.3128 \\
    & $\pm 0.1213$ & $\pm0.03829$ & $\pm0.1616$ & $\pm0.0728$ & $\pm0.1355$ & $\pm0.1143$ \\
    \hspace{0.2cm} Spectral & \textbf{1.0000} & \textbf{1.0000} & 0.1400 & 0.2509 & 0.2778 & 0.3296 \\
    & $\pm 0.0000$ & $\pm0.0000$ & $\pm0.1920$ & $\pm0.2418$ & $\pm0.1541$ & $\pm0.0153$ \\
    \midrule
    Our method & 0.6141 & 0.7721 & 0.3408 & \textbf{0.5348} &  0.6134  & \textbf{0.6521}   \\ 
    & $\pm 0.2577$ & $\pm 0.1988$ & $\pm 0.1867$ & $\pm 0.1145$ & $\pm 0.3341$ & $\pm 0.1770$ \\
    \bottomrule
  \end{tabular}
  \label{tab:hb_evaluation_comparison}
\end{table*}

In our quantitative evaluation, we first analyzed the quality of our learned hierarchical clustering by calculating two clustering quality metrics at each level of the hierarchy: the Adjusted Rand Index (ARI) \cite{hubert1985comparing} and Normalized Mutual Information (NMI) \cite{shannon1948mathematical}.
This type of evaluation jointly assesses the quality of the learned community hierarchy as well as the membership of entities to communities.
The ARI is an adjustment to the commonly used Rand Index (RI) \cite{rand1971objective}, corrected to account for chance. Specifically, chance is factored in by calculating the expected RI given a random clustering and measuring the obtained clustering's deviation. Specifically, given an obtained entity clustering $\mathcal{C} = \{\mathcal{C}_1, \mathcal{C}_2, \ldots,  \mathcal{C}_o \}$ and the ground truth clustering $\mathcal{C}^* = \{ \mathcal{C}^*_1, \mathcal{C}^*_2, \ldots,  \mathcal{C}^*_t \}$ , the ARI is calculated as follows:
\begin{align}
    ARI & = \dfrac{\sum_{\mathcal{C}_i \in \mathcal{C}} \sum_{\mathcal{C}^*_j \in \mathcal{C}^*} {\#_{ij} \choose 2} - {| \mathcal{E} | \choose 2}^{-1} \Big( \sum_{\mathcal{C}_i \in \mathcal{C}} {\#_{i} \choose 2} \sum_{\mathcal{C}^*_j \in \mathcal{C}^*} {\#^*_{j} \choose 2} \Big) }{2^{-1} \Big( \sum_{\mathcal{C}_i \in \mathcal{C}} {\#_{i} \choose 2} + \sum_{\mathcal{C}^*_j \in \mathcal{C}^*} {\#^*_{j} \choose 2} \Big) - {| \mathcal{E} | \choose 2}^{-1} \Big( \sum_{\mathcal{C}_i \in \mathcal{C}} {\#_{i} \choose 2} \sum_{\mathcal{C}^*_j \in \mathcal{C}^*} {\#^*_{j} \choose 2} \Big)}
\end{align}
where $\#_{ij} = | \mathcal{C}_i \cap \mathcal{C}^*_j |$ is the number of entities in common between a ground truth and obtained cluster pair; $\#_i = \sum_{\mathcal{C}^*_j \in \mathcal{C}^*} | \mathcal{C}_i \cap \mathcal{C}^*_j |$ is the total number of entities in obtained cluster $\mathcal{C}_i$; and $\#^*_j = \sum_{\mathcal{C}_i \in \mathcal{C}} | \mathcal{C}_i \cap \mathcal{C}^*_j |$ is the total number of entities in ground truth cluster $\mathcal{C}^*_j$.
The NMI is a normalized extension of the Mutual Information (MI) score which quantifies the information gained about the obtained clustering given the ground truth clusters. The normalization of the MI score ensures the result is in the range $[0,1]$ thereby allowing for its comparison against clusterings of different sizes.
Utilizing the notation defined earlier, we define MI and NMI as follows:
\begin{align}
    NMI & =  \dfrac{\sum_{\mathcal{C}_i \in \mathcal{C}} \sum_{\mathcal{C}^*_j \in \mathcal{C}^*} \dfrac{| \mathcal{C}_i \cap \mathcal{C}^*_j|}{|\mathcal{E}|} \log \Bigg( \dfrac{|\mathcal{E}| | \mathcal{C}_i \cap \mathcal{C}^*_j|}{ | \mathcal{C}_i | | \mathcal{C}^*_j|} \Bigg)}{\text{mean}\Bigg( - \sum_{\mathcal{C}_i \in \mathcal{C}} \dfrac{| \mathcal{C}_i |}{ | \mathcal{E} |} \log \Bigg( \dfrac{| \mathcal{C}_i |}{ | \mathcal{E} |} \Bigg) ,  - \sum_{\mathcal{C}^*_j \in \mathcal{C}^*} \dfrac{| \mathcal{C}^*_j |}{ | \mathcal{E} |} \log \Bigg( \dfrac{| \mathcal{C}^*_j |}{ | \mathcal{E} |} \Bigg) \Bigg)}
\end{align}
For both the ARI and NMI, higher scores indicate a clustering of higher quality. We summarize the results of our clustering as per these two metrics in Table \ref{tab:hb_evaluation}.

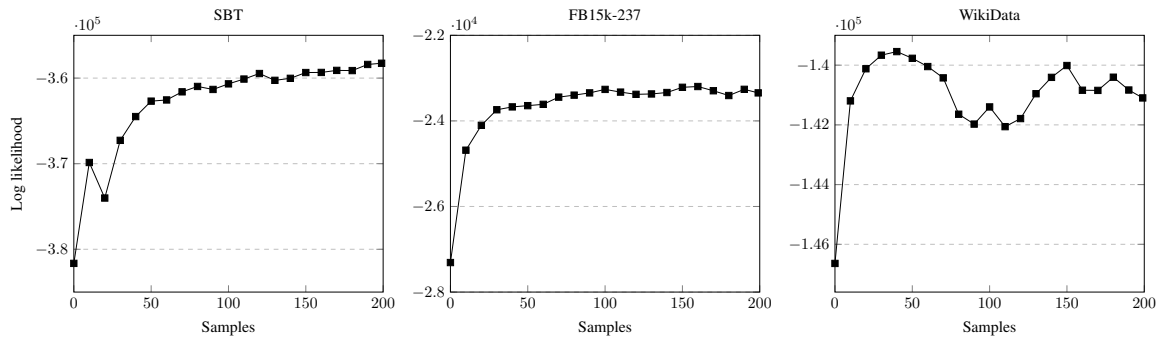
\begin{figure}[t!]
\centering
\begin{tikzpicture}[scale=0.60]
\begin{axis}[
    title = {SBT},
    xlabel={Samples},
    ylabel={Log likelihood},
    xmin=0, xmax=200,
    ymin=-385000, ymax=-355000,
    log ticks with fixed point,
    xtick={},
    ytick={},
    ymajorgrids=true,
    grid style=dashed,
]
\addplot[color=black, mark=square*,]
    coordinates {(0,-381646.97)(10,-369860.77)(20,-374013.09)(30,-367273.00)(40,-364493.96)(50,-362692.02)(60,-362552.84)(70,-361605.10)(80,-360975.55)(90,-361325.40)(100,-360673.71)(110,-360112.10)(120,-359466.05)(130,-360261.55)(140,-360026.51)(150,-359350.19)(160,-359337.59)(170,-359098.99)(180,-359121.50)(190,-358405.33)(199,-358265.08)};
\end{axis}
\end{tikzpicture}
\begin{tikzpicture}[scale=0.60]
\begin{axis}[
    title = {FB15k-237},
    xlabel={Samples},
    ylabel={},
    xmin=0, xmax=200,
    ymin=-28000, ymax=-22000,
    log ticks with fixed point,
    xtick={},
    ytick={},
    ymajorgrids=true,
    grid style=dashed,
]
    \addplot[color=black, mark=square*,]
    coordinates {(0,-27310.20)(10,-24684.31)(20,-24103.33)(30,-23739.67)(40,-23672.71)(50,-23643.59)(60,-23612.85)(70,-23444.69)(80,-23398.12)(90,-23345.55)(100,-23263.91)(110,-23327.32)(120,-23379.14)(130,-23370.35)(140,-23336.82)(150,-23214.62)(160,-23196.77)(170,-23295.57)(180,-23407.17)(190,-23262.15)(199,-23345.30)};
\end{axis}
\end{tikzpicture}
\begin{tikzpicture}[scale=0.60]
\begin{axis}[
    title = {WikiData},
    xlabel={Samples},
    ylabel={},
    xmin=0, xmax=200,
    ymin=-147600, ymax=-139000,
    log ticks with fixed point,
    xtick={},
    ytick={},
    legend pos=south east,
    ymajorgrids=true,
    grid style=dashed,
]
    \addplot[color=black, mark=square*,]
    coordinates {(0,-146640.63)(10,-141196.62)(20,-140120.67)(30,-139665.65)(40,-139545.62)(50,-139770.03)(60,-140046.82)(70,-140425.76)(80,-141643.25)(90,-141973.90)(100,-141399.14)(110,-142059.67)(120,-141788.88)(130,-140956.79)(140,-140408.64)(150,-140012.08)(160,-140839.35)(170,-140845.07)(180,-140403.11)(190,-140832.64)(199,-141097.54)};
\end{axis}
\end{tikzpicture}
\caption[Log likelihood plots for hierarchical blockmodelling]{Plots of the average log likelihood of our model across burn in samples on three datasets.}
\label{fig:hb_likelihoods}
\end{figure}

In general, the results indicate that our model is capable of learning a coherent community hierarchy on each of the three datasets tested. Perhaps unsurprisingly, communities at higher levels in the hierarchy are judged as of higher quality as per the two evaluation metrics. This is because the task of clustering entities at higher levels is simpler as the communities are less fine-grained. For instance, on the FB15k-237 dataset, clustering at level 1 requires the distinction between \texttt{Place} and \texttt{Person} whereas level 4 requires the distinction between \texttt{AmericanFootballPlayer} and \texttt{IceHockeyPlayer}.
We note that the SBT dataset is an exception to this. This is likely due to the nature of the dataset wherein entity relations are drawn at higher proportions between neighbouring communities at lower levels of the hierarchy. In this sense, the claim made before gets inverted and it's easier to assign communities at lower levels in the hierarchy.  
We also compared our model against embedding and clustering methods used in conjunction. Specifically, we first embedded each of the knowledge graphs using the RDF2VEC and TransE embedding methods. Afterwards, we applied four clustering methods: $k$-means, Agglomerative, DBSCAN, and Spectral. These results are summarized in Table \ref{tab:hb_evaluation_comparison} and indicate comparable or superior performance to baselines.

We can also analyze the results of the complete log likelihood as a function of the number of Gibbs samples taken in the inference process. Indeed, while this does not provide us with information about the quality of the obtained results, it does verify the inference process itself.
Specifically, we expect to see the log likelihood of our model to rise given more burn-in samples of the Gibbs sampler. This suggests that the likelihood of generating the knowledge graph given the current state of the sampler is increasing and learning is taking place.
We can see this rise in Figure \ref{fig:hb_likelihoods} which plots the complete log likelihoods of our model across Gibbs samples for the three datasets.
We note dips in log likelihoods on the SBT and WikiData datasets. This is likely due to the sampler being temporarily stuck in a local minimum before leaving that area in the sample space.

\subsection{Qualitative Evaluation}\label{subsec:hb_qualitative_evaluation}

In our qualitative evaluation, we leverage the qualitative attributes possessed by a useful taxonomy as identified by Nickerson et al. \cite{nickerson2013method}. Although these attributes were proposed in the context of taxonomy development, we make use of them in our work as the task of hierarchical clustering shares many of the underlying evaluation principles. Indeed, a taxonomy is implicitly induced using our method, although never explicitly labelled.
The proposed taxonomy attributes are as follows: concise, robust, comprehensive, extendable, and explanatory. For each of these attributes, we provide a brief description extended to hierarchical clustering and use it to evaluate our model.
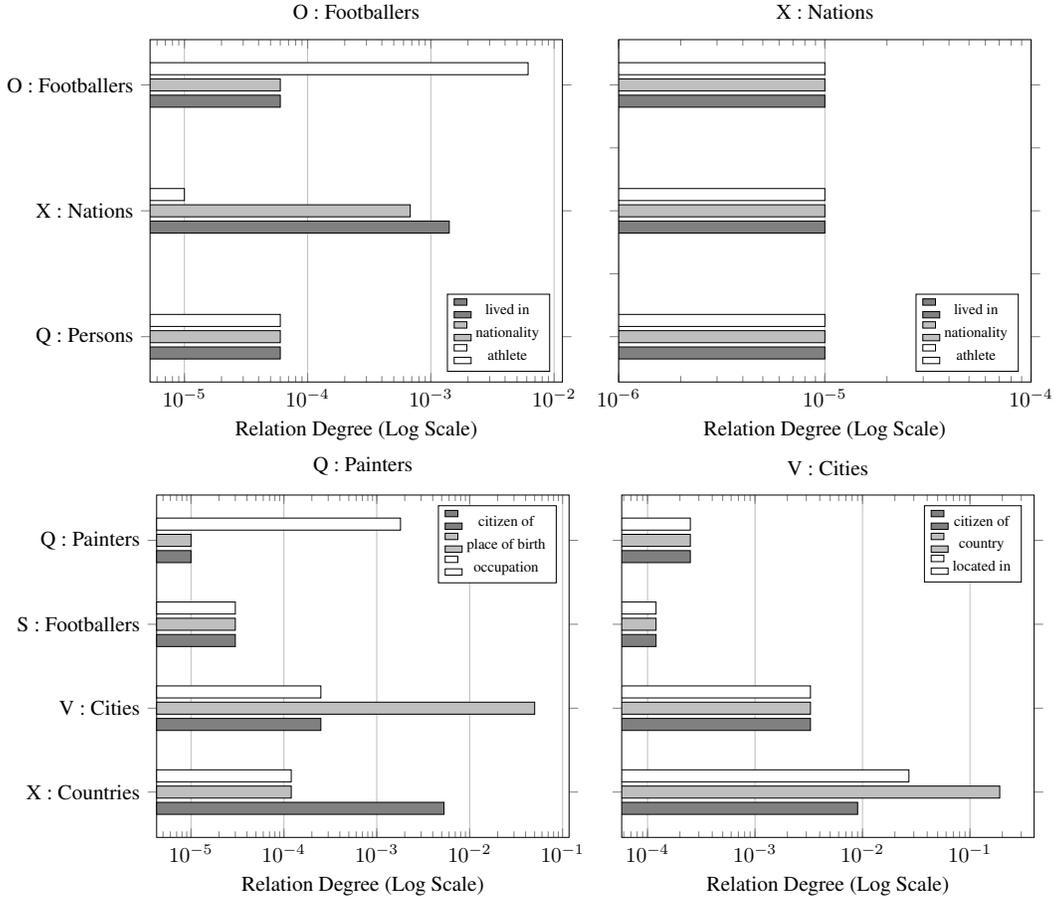
\begin{figure}[t!]
\centering
\begin{tikzpicture}[scale=0.8]
\centering
  \begin{axis}[
        xbar,
        xmode=log, log origin=infty,
        bar width=2mm, 
        xmajorgrids = true,
        enlarge y limits=0.18,
        xlabel={Relation Degree (Log Scale)},
        symbolic y coords={Q : Persons,X : Nations,O : Footballers},
        ytick=data,
        nodes near coords
         align={horizontal},
         legend pos = south east,
         title = O : Footballers
  ]
    \addplot [black,fill=gray] coordinates{(0.00006,O : Footballers)(0.00141,X : Nations)(0.00006,Q : Persons)};
    \addplot [black,fill=gray!50] coordinates{(0.00006,O : Footballers)(0.00068,X : Nations)(0.00006,Q : Persons)};
    \addplot [black,fill=white] coordinates{(0.00615,O : Footballers)(0.00001,X : Nations)(0.00006,Q : Persons)};
    \legend {\scriptsize{lived in}, \scriptsize{nationality}, \scriptsize{athlete}};
  \end{axis}
\end{tikzpicture}
\begin{tikzpicture}[scale=0.8]
\centering
  \begin{axis}[
        xbar,
        xmode=log, log origin=infty,
        bar width=2mm, 
        xmajorgrids = true,
        enlarge y limits=0.18,
        xlabel={Relation Degree (Log Scale)},
        symbolic y coords={Q : Persons,X : Nations,O : Footballers},
        yticklabel=\empty,
        nodes near coords
         align={horizontal},
         legend pos = south east,
         title = X : Nations
  ]
    \addplot [black,fill=gray] coordinates{(0.00001,O : Footballers)(0.00001,X : Nations)(0.00001,Q : Persons)};
    \addplot [black,fill=gray!50] coordinates{(0.00001,O : Footballers)(0.00001,X : Nations)(0.00001,Q : Persons)};
    \addplot [black,fill=white] coordinates{(0.00001,O : Footballers)(0.00001,X : Nations)(0.00001,Q : Persons)};
    \legend {\scriptsize{lived in}, \scriptsize{nationality}, \scriptsize{athlete}};
  \end{axis}
\end{tikzpicture}
\hspace{0.5cm}
\begin{tikzpicture}[scale=0.8]
\centering
  \begin{axis}[
        xbar,
        xmode=log, log origin=infty,
        bar width=2mm, 
        xmajorgrids = true,
        enlarge y limits=0.18,
        xlabel={Relation Degree (Log Scale)},
        symbolic y coords={X : Countries,V : Cities,S : Footballers,Q : Painters},
        ytick=data,
        nodes near coords
         align={horizontal},
         legend pos = north east,
         title = Q : Painters
  ]
    \addplot [black,fill=gray] coordinates{(0.00001,Q : Painters)(0.00003,S : Footballers)(0.00025,V : Cities)(0.00529,X : Countries)};
    \addplot [black,fill=gray!50] coordinates{(0.00001,Q : Painters)(0.00003,S : Footballers)(0.05011,V : Cities)(0.00012,X : Countries)};
    \addplot [black,fill=white] coordinates{(0.00180,Q : Painters)(0.00003,S : Footballers)(0.00025,V : Cities)(0.00012,X : Countries)};
    \legend {\scriptsize{citizen of}, \scriptsize{place of birth}, \scriptsize{occupation}};
  \end{axis}
\end{tikzpicture}
\begin{tikzpicture}[scale=0.8]
\centering
  \begin{axis}[
        xbar,
        xmode=log, log origin=infty,
        bar width=2mm, 
        xmajorgrids = true,
        enlarge y limits=0.18,
        xlabel={Relation Degree (Log Scale)},
        symbolic y coords={X : Countries,V : Cities,S : Footballers,Q : Painters},
        yticklabel=\empty,
        nodes near coords
         align={horizontal},
         legend pos = north east,
         title = V : Cities
  ]
    \addplot [black,fill=gray] coordinates{(0.00025,Q : Painters)(0.00012,S : Footballers)(0.00328,V : Cities)(0.00901,X : Countries)};
    \addplot [black,fill=gray!50] coordinates{(0.00025,Q : Painters)(0.00012,S : Footballers)(0.00328,V : Cities)(0.18919,X : Countries)};
    \addplot [black,fill=white] coordinates{(0.00025,Q : Painters)(0.00012,S : Footballers)(0.00328,V : Cities)(0.02703,X : Countries)};
    \legend {\scriptsize{citizen of}, \scriptsize{country}, \scriptsize{located in}};
  \end{axis}
\end{tikzpicture}
\caption[Plots of learned community relations]{Plots of learned community relations for selected outgoing predicates for the FB15k-237 and WikiData datasets. Specifically, we showcased community O (Footballers) and community T (Nations) outgoing relations for the FB15k-237 dataset (top) and community Q (Painters) and community V (Cities) outgoing relations for the WikiData dataset.}
\label{fig:hb_community_relations}
\end{figure}

\begin{figure}[t!]
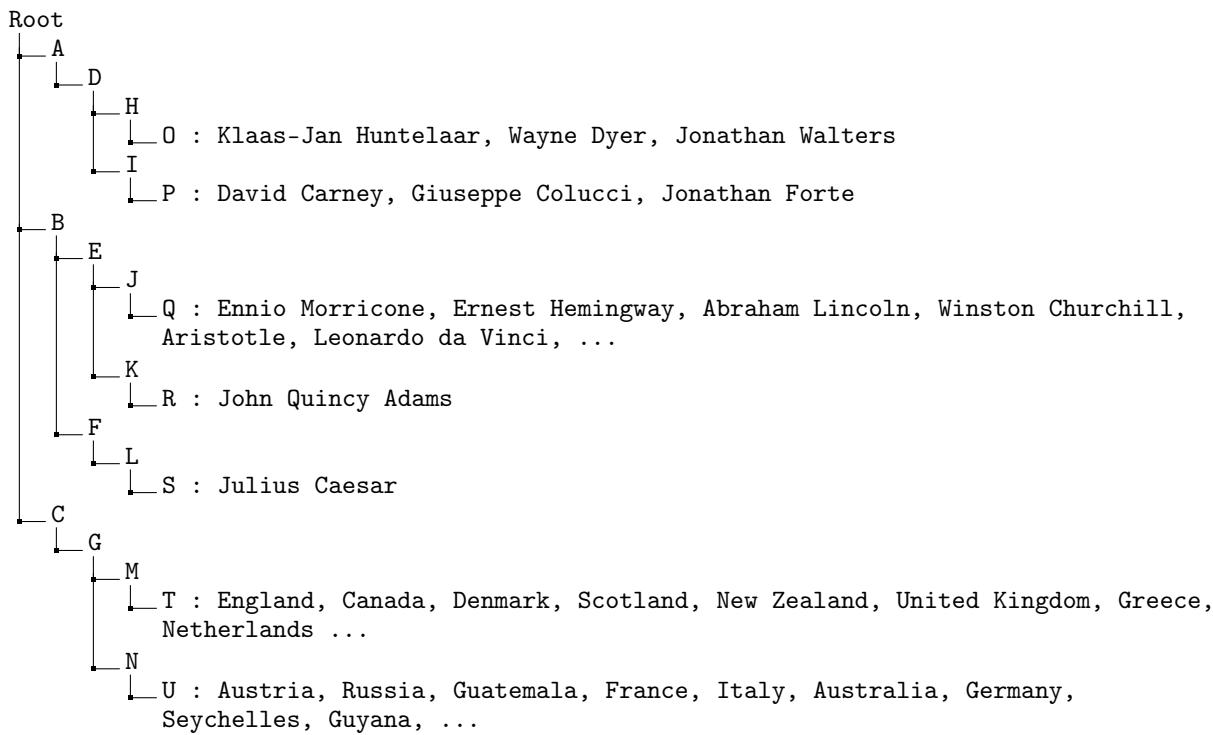

\raggedright
\dirtree{%
.1 Root. 
.2 A.
.3 D.
.4 H.
.5 O : Klaas-Jan Huntelaar, Wayne Dyer, Jonathan Walters.
.4 I.
.5 P : David Carney, Giuseppe Colucci, Jonathan Forte.
.2 B.
.3 E.
.4 J.
.5 Q : Ennio Morricone, Ernest Hemingway, Abraham Lincoln, Winston Churchill, Aristotle, Leonardo da Vinci, {...} .
.4 K.
.5 R : John Quincy Adams.
.3 F.
.4 L.
.5 S : Julius Caesar.
.2 C.
.3 G.
.4 M.
.5 T : England, Canada, Denmark, Scotland, New Zealand, United Kingdom, Greece, Netherlands {...}.
.4 N.
.5 U : Austria, Russia, Guatemala, France, Italy, Australia, Germany, Seychelles, Guyana, {...}.
}
\caption{Excerpt of our induced hierarchy on the FB15k-237 dataset. Entities in communities \texttt{O} and \texttt{P} are footballers. Ennio Morricone is a pianist. Ernest Hemmingway is a journalist. Abraham Lincoln, Winston Churchill, John Quincy Adams, and Julius Caesar are politicians. Aristotle and Leonardo da Vinci are scientists. The entities in communities \texttt{T} and \texttt{U} are countries.}
\label{fig:hb_tree}
\end{figure}

\begin{figure}[t!]
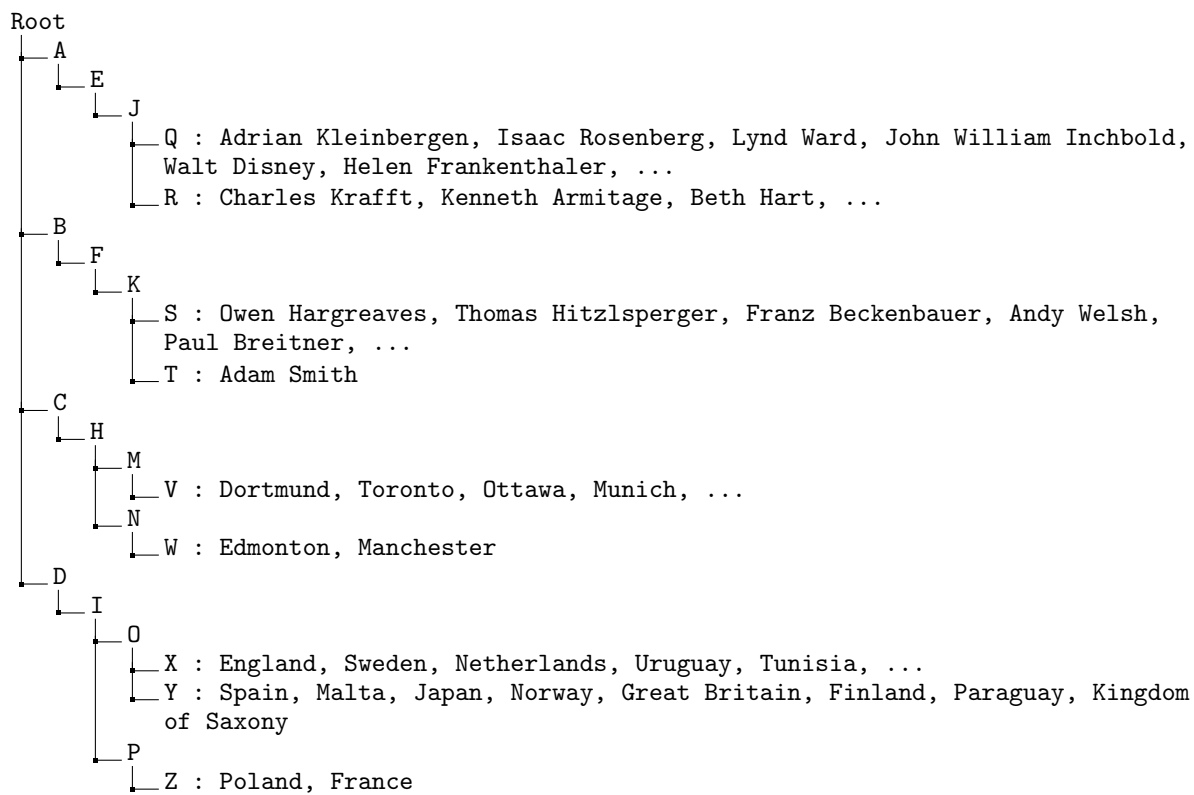

\raggedright
\dirtree{%
.1 Root. 
.2 A.
.3 E.
.4 J.
.5 Q : Adrian Kleinbergen, Isaac Rosenberg, Lynd Ward, John William Inchbold, Walt Disney, Helen Frankenthaler, {...} .
.5 R : Charles Krafft, Kenneth Armitage, Beth Hart, \texttt{...} .
.2 B.
.3 F.
.4 K.
.5 S : Owen Hargreaves, Thomas Hitzlsperger, Franz Beckenbauer, Andy Welsh, Paul Breitner, {...} .
.5 T : Adam Smith.
.2 C.
.3 H.
.4 M.
.5 V : Dortmund, Toronto, Ottawa, Munich, {...}.
.4 N.
.5 W : Edmonton, Manchester.
.2 D.
.3 I.
.4 O.
.5 X : England, Sweden, Netherlands, Uruguay, Tunisia, {...} .
.5 Y : Spain, Malta, Japan, Norway, Great Britain, Finland, Paraguay, Kingdom of Saxony .
.4 P.
.5 Z : Poland, France.
}
\caption{Excerpt of our induced hierarchy on the WikiData dataset. Entities in communities \texttt{Q} and \texttt{R} are painters, in communities \texttt{S} and \texttt{T} are footballers, in communities \texttt{V} and \texttt{W} are cities, and in communities \texttt{X}, \texttt{Y}, and \texttt{Z} are countries as defined in WikiData.}
\label{fig:hb_tree1}
\end{figure}

\begin{itemize}
    \item A concise clustering is limited in both the number of obtained clusters and the semantic diversity of the entities that constitute each cluster. In our method, this is largely regulated by the hyperparameters $\gamma$, allowing for control over the number of clusters obtained, and the interplay between $\lambda$ and $\sigma$, regulating the intracluster density. In practice, however, the hyperparameter tuning required is fickle. Figure \ref{fig:hb_tree} demonstrates this through an excerpt of the results obtained on the FB15k-237 dataset. Singleton communities \texttt{R} and \texttt{S} are both composed of politician entities and, in keeping with the conciseness attribute, warrant a merger. Community \texttt{Q} on the other hand is composed of a heterogeneous set of entities and requires splitting. It is concise, but, as we shall elaborate upon later, it is not robust. Thus the direction in which to adjust the hyperparameters is not clear. The induced hierarchy also suffers in terms of conciseness at higher levels in the hierarchy. Communities \texttt{A} and \texttt{B} are both comprised of people and require a merger at their level. This issue can largely be explained by the data itself. Footballers in our dataset have structural properties which differ them from the persons clustered in the community \texttt{B} subtree. In addition to sharing the same profession, they belong to football teams, have football specific triples such as the position they play, and are more likely than non-footballer persons in the dataset to have information relating to physical characteristics such as height and weight. In contrast, non-footballer persons have less structural similarities, even within members of their own professions. For instance, even though all scientists have their scientific contributions, this isn't reflected in the data as uniformly. The WikiData dataset -- an excerpt of which is displayed in Figure \ref{fig:hb_tree1} -- is simpler than FB15k-237 due to there being only two professions with largely disjoint neighbourhoods. The induced hierarchy is not concise, however, in the splitting of persons and places. Specifically, this happens at the first level where persons are split into communities \texttt{A} and \texttt{B} and places into communities \texttt{C} and \texttt{D}. We note, however, the high ARI and NMI scores at this level despite this error. A closer inspection reveals that not all runs of our model encounter this issue, thus the scores are higher than they would be had they been measured only on the hierarchy presented in the excerpt.
    \item Robustness in clustering refers to ``maximizing both within-group homogeneity and between-group heterogeneity, [making] groups that are as distinct (nonoverlapping) as possible, with all members within a group being as alike as possible \cite{bailey1994typologies}.'' We see robustness as an issue when examining the non-footballer professions in Figure \ref{fig:hb_tree}. Pianists, journalists, politicians, and scientists are not sufficiently semantically homogenous to warrant their inclusion in community \texttt{Q}. Another issue is the splitting of nations into communities \texttt{O} and \texttt{P} as there is no evident geographic, social, political, or economic distinction made in the clustering. A deeper look into the data also reveals no apparent structural differences between communities \texttt{M} and \texttt{N}. It is likely, therefore, that this issue can be due to model inaccuracy and arrival at a suboptimal state after inference. The WikiData dataset also suffers from unsubstantiated splits at the lowest levels as seen in communities \texttt{R}, \texttt{T}, and \texttt{Y}. This issue may be the result of the small size of the data and the model's resultant sensitivity to the relational information provided. For instance, close to half of the entities in this dataset have first order neighbourhoods of one or two entities, giving our model little to learn from.
    \item A taxonomy is comprehensive if it can classify all entities in the domain. Clustering, including the hierarchical clustering obtained by our method, is induced empirically from the data and thus necessarily classifies all entities into communities. As such, our generative model is comprehensive.
    \item An extendable clustering is one that allows for the dynamic inclusion of additional information and, in the hierarchical case, the structural change and adaptability to incoming information. In these two respects, our method is highly extendable. Indeed, the Gibbs sampling process itself requires the removal of entities from the hierarchy before they are resampled. Each resampling not only has the potential to change the community assignment of the entity but also to change the structure of the hierarchy itself. Due to the infinite formulation of the nCRP and stick breaking process, there is no prior constraint on this structure. 
    \item The final qualitative attribute identified in a useful taxonomy is that it is explanatory. In this regard, the taxonomy should provide useful explanations about the objects it is organizing. In the context of hierarchical clustering, these explanations could, for instance, take the form of community labels. Although our model does not assign labels to communities, they can be ascertained by examining the type information of their constituent entities. For instance in Figure \ref{fig:hb_tree} the FB15k-237 communities \texttt{O}, \texttt{X}, and \texttt{P} correspond to footballers, nations, and persons and the WikiData communities \texttt{Q}, \texttt{S}, \texttt{V}, and \texttt{X} correspond to painters, footballers, cities, and countries. An explanatory advantage of our method is that it infers community relations as part of the generative process. A fragment of these relations is conveyed in Figure \ref{fig:hb_community_relations}. The results indicate community relations which are largely expected. For instance, on the FB15k-237 dataset, footballers are much more likely to be related to nations by predicates \textit{nationality} and \textit{lived in} than \textit{athlete}. Furthermore, we see that nations are equally unlikely to be the subjects with the other communities or predicates such as lived in, nationality, and athlete. On the WikiData dataset, we likewise see explainable results. Painters, for instance, are likely to be related to cities by place of birth and countries by nationality. They are unlikely to relate to other painters and footballers on these predicates.
\end{itemize}

\section{Conclusion}\label{sec:conclusion}

In this paper, we demonstrated the use of stochastic blockmodels for learning hierarchies from knowledge graphs in an approach that is, to the best of our knowledge, the first to marry these two fields in an academic setting. To this end, we proposed a model which leverages the nCRP and stick breaking process to generate a community hierarchy composed of a knowledge graph's constituent entities. The model is defined non-parametrically and thus makes no assumptions about its structure, allowing it to adapt to the knowledge graph and potentially induce an infinite number of communities on an infinite number of levels. In addition to the model itself, a Markov chain Monte Carlo scheme leveraging collapsed Gibbs sampling was devised for posterior inference of the model's parameters. The model was evaluated on three datasets using quantitative and qualitative analysis to demonstrate its effectiveness in learning coherent community hierarchies on both synthetic and real-world data. The qualitative analysis made use of attributes commonly employed in taxonomy evaluation, presenting a novel and principled for qualitative analysis of hierarchical clusterings. 
Future work will investigate scalability when applying our model -- and indeed stochastic blockmodels more generally -- to knowledge graphs. As discussed earlier, the time complexities of the inference schemes for these models usually do not allow for scaling to the types of large knowledge bases that are encountered in real-world applications.
The inference scheme proposed in this work is one such instance however several methods exist in the literature more scalable than collapsed Gibbs sampling.
An example of such a method is variational inference, a scheme that uses the evidence lower bound to guide the inference process and obtain the posterior distribution. This method is generally faster than Gibbs sampling and, although not asymptotically exact, produces similar results \cite{salimans2015markov}. The challenge with this approach is that its optimization equations are more difficult to solve as compared to Markov chain Monte Carlo methods.
Despite this, several works have already successfully applied variational inference to probabilistic graphical models. Indeed, the original inference scheme for the Mixed Membership Stochastic Blockmodel leveraged variational inference.
Furthermore, Blei and Jordan \cite{blei2006variational} provided a variational inference scheme for Dirichlet processes and a variational inference scheme for the nCRP was proposed in Wang and Blei \cite{wang2009variational}. 
Departing from the variational approach, Chen et al. \cite{chen2017scalable} propose an evolution of the Gibbs sampling algorithm for the nCRP with partially collapsed Gibbs sampling. This approach resulted in a time increase of over two magnitudes over the classic Gibbs sampling approach.
Another line of approach to increase the scalability of stochastic blockmodels is to devise a model which does not require sampling all $|\mathcal{E}|^2|\mathcal{R}|$ relations in the knowledge graph directly. To this end, the Bernoulli-Poisson link function has been applied successfully to simple graphs \cite{zhou2015infinite,rai2015large,fan2019scalable}. These methods eliminate the need for quadratic time relation sampling and instead rely on density based sampling which is less computationally demanding, especially on sparse networks. Given that most knowledge graphs are highly sparse, applying such an approach appears promising. The work presented in this paper provides evidence for further research in this direction in the Semantic Web community.

\nocite{*}
\bibliographystyle{abbrv}
\bibliography{references} 

\appendix 
\numberwithin{equation}{section}
\makeatletter 
\newcommand{\section@cntformat}{Appendix \thesection:\ }
\makeatother

\section{Probability Mass and Density Functions}\label{adx:a}

In this appendix, we provide the probability mass and density functions for the distributions used in our paper. Probability mass functions capture the probability of a discrete random variable realizing a value, denoted $x$: 
\begin{align}
    \text{Bernoulli}(p, q) & =  p^x q^{(n-x)} \label{eqn:bernoulli}\\
    \text{Multinomial}(\mathbf{p}) & = \dfrac{\Gamma(\sum_i x_i + 1))}{\prod_i\Gamma(x_i + 1))}\prod_i^L p_i^{x_i} \label{eqn:multinomial}
\end{align}
Where $p$, $q$, and $\mathbf{p}$ are the parameters of their respective distributions. Probability density functions capture the relative likelihood of a continuous random variable realizing the value $x$:
\begin{align}
    \text{Beta}(\alpha, \beta) & = \dfrac{x^{\alpha - 1} (1-x)^{\beta - 1}}{\text{B}(\alpha, \beta)} \ , \text{B}(\alpha, \beta) = \dfrac{\Gamma(\alpha)\Gamma(\beta)}{\Gamma(\alpha + \beta)} \label{eqn:beta}\\
    \text{Dirichlet}(\pmb{\alpha}, L) & = \dfrac{\Gamma(\sum_{l=1}^L \alpha_l)}{\prod_{l=1}^L \Gamma(\alpha_l)} \prod_{l=1}^L x_l^{\alpha_l - 1} \label{eqn:dirichlet} \\
\end{align}
Where $\alpha$, $\beta$, $\pmb{\alpha}$, and $L$ are the parameters of their respective distributions.

\section{Integral Form of the Beta Function} \label{adx:integral_form_beta_function}

In this appendix, we provide the derivation to obtain the integral form of the Beta function. We do this by leveraging the definition of the Beta distribution. Specifically, we begin with the identity that the integral of a probability density function with respect to its support is equal to 1 and proceed with simple integral calculus:
\begin{align}
    \bigintssss_{\ 0}^1 \text{Beta}(\alpha, \beta) \ \text{d}x & = 1 \nonumber \\
    \bigintssss_{\ 0}^1 \dfrac{x^{\alpha - 1} (1-x)^{\beta - 1}}{\text{B}(\alpha, \beta)} \ \text{d}x & = 1 \nonumber \\
    \dfrac{1}{\text{B}(\alpha, \beta)} \bigintssss_{\ 0}^1 x^{\alpha - 1} (1-x)^{\beta - 1} \ \text{d}x & = 1 \nonumber \\
    \bigintssss_{\ 0}^1 x^{\alpha - 1} (1-x)^{\beta - 1} \ \text{d}x & = \text{B}(\alpha, \beta) \label{eqn:normalizingconstant}
\end{align}

\section{Simplifying Level Likelihood} \label{adx:level_likelihood}

In this appendix, we provide the simplification of level likelihood by eliminating the Gamma function.
Recall from Equation \ref{eqn:zlikelihood} that the level likelihood, $\mathbb{P}(\mathbf{g}_{ij*} \ | \  \mathbf{G}_{-(ij*)},  \mathbf{P}, \mathbf{Z},  \lambda, \eta)$, is expressed as follows:
\begin{align}
    \mathbb{P}(\mathbf{g}_{ij*} \ | \  \mathbf{G}_{-(ij*)},  \mathbf{P}, & \mathbf{Z},  \lambda, \eta) \nonumber \\
    & = \prod_{g_{ijr} \in \mathbf{g}_{ij*}}  \dfrac{\Gamma(\#^{c_{pqr}=1}_{-(ijr)}  + g_{ijr} + \lambda)\Gamma(\#^{c_{pqr}=0}_{-(ijr)} + (1 - g_{ijr}) + \eta )\Gamma(\#^{c_{pqr}=1}_{-(ijr)}  + \#^{c_{pqr}=0}_{-(ijr)} + \lambda + \eta)}{\Gamma(\#^{c_{pqr}=1}_{-(ijr)}  + \#^{c_{pqr}=0}_{-(ijr)} + 1 + \lambda + \eta)\Gamma(\#^{c_{pqr}=1}_{-(ijr)}  + \lambda)\Gamma(\#^{c_{pqr}=0}_{-(ijr)} + \eta)}  \nonumber \\
    & = \prod_{g_{ijr} \in \mathbf{g}_{ij*}} \dfrac{1}{\#^{c_{pqr}=1}_{-(ijr)}  + \#^{c_{pqr}=0}_{-(ijr)} + \lambda + \eta}  \dfrac{\Gamma(\#^{c_{pqr}=1}_{-(ijr)}  + g_{ijr} + \lambda)\Gamma(\#^{c_{pqr}=0}_{-(ijr)} + (1 - g_{ijr}) + \eta )}{\Gamma(\#^{c_{pqr}=1}_{-(ijr)}  + \lambda)\Gamma(\#^{c_{pqr}=0}_{-(ijr)} + \eta)} \label{eqn:level_likelihood}
\end{align}
We can leverage the fact that $g_{ijr} \in \{0,1\}$ to define the second term as a piecewise function with respect to the value of $g_{ijr}$. Doing so allows us to cancel out terms which appear in both the numerator and denominator after expanding the Gamma function. This process leads to the following simplification:
\begin{align}
    \dfrac{\Gamma(\#^{c_{pqr}=1}_{-(ijr)}  + g_{ijr} +  \lambda)\Gamma(\#^{c_{pqr}=0}_{-(ijr)} + (1 - g_{ijr}) + \eta )}{\Gamma(\#^{c_{pqr}=1}_{-(ijr)}  + \lambda)\Gamma(\#^{c_{pqr}=0}_{-(ijr)} + \eta)}  = \begin{dcases}  \#^{c_{pqr}=1}_{-(ijr)}  + \lambda & g_{ijr} = 1 \\
     \#^{c_{pqr}=0}_{-(ijr)} + \eta & g_{ijr} = 0 \label{eqn:level_likelihood_second_term}
    \end{dcases}
\end{align}
This allows us to put together Equations \ref{eqn:level_likelihood} and \ref{eqn:level_likelihood_second_term} to get the following, ass seen in Equation \ref{eqn:zlikelihood}:
\begin{align}
    \mathbb{P}(\mathbf{g}_{ij*} \ | \ \mathbf{G}_{-(ij*)}, \mathbf{P}, \mathbf{Z}, \lambda, \eta) & = \prod_{g_{ijr} \in \mathbf{g}_{ij*}} \dfrac{g_{ijr}(\#^{c_{pqr}=1}_{-(ijr)}  + \lambda) + (1-g_{ijr})(b + \eta)}{\#^{c_{pqr}=1}_{-(ijr)}  + b + \lambda + \eta}
\end{align}

\section{Marginalizing Finite Level Memberships} \label{adx:level_mixtures}

In order to marginalize finite level memberships, we begin with the definition of its posterior, $\mathbbm{P}(\mathbf{a}_i \ | \ \mathbf{A}_{-i}, \mathbf{Z}, \pmb{\alpha})$, which is defined analogously to Equation \ref{eqn:level_mixtures} with the exception that the Dirichlet prior is used in this case. Formally, we obtain the following through Bayes' rule:
\begin{align}
    \mathbbm{P}(\mathbf{a}_i \ | \ \mathbf{A}_{-i}, \mathbf{Z}, \pmb{\alpha}) & = \dfrac{\mathbbm{P}(\mathbf{Z} \ | \ \mathbf{A, \pmb{\alpha}) \mathbbm{P}(\mathbf{a}_i \ | \ \mathbf{A}_{-i}, \pmb{\alpha})}}{\bigintssss_{\mathbf{a}_{i}} \mathbbm{P}(\mathbf{Z} \ | \ \mathbf{A, \pmb{\alpha}) \mathbbm{P}(\mathbf{a}_i \ | \ \mathbf{A}_{-i}, \pmb{\alpha}) \text{d}\mathbf{a}_i}}\\
    & = \dfrac{\text{Multinomial}(\mathbf{a}_i)\text{Dirichlet}(\pmb{\alpha}, L)}{\bigintssss_{\mathbf{a}_i} \text{Multinomial}(\mathbf{a}_i)\text{Dirichlet}(\pmb{\alpha}, L) \ \text{d}\mathbf{a}_i } 
\end{align}
Where $\pmb{\alpha}$ is a vector of $L$ concentration parameters for each level in the distribution, namely $\pmb{\alpha} = [\alpha_1, \alpha_2, ..., \alpha_L]$ such that $\alpha_l > 0$ and $L$ is the finite number of levels in the hierarchy.
In our marginalization, we adopt the notation from Equation \ref{eqn:level_membership_marginalization_definitions} to indicate the number of indicators in $\mathbf{z}_{i*}$. Furthermore, we define the following vector of concentration parameters to aid in readability:
$\pmb{\alpha'} = [\alpha_1 + \#^{\mathbf{z}_{i*}=1}, \alpha_2 + \#^{\mathbf{z}_{i*}=2}, ..., \alpha_L + \#^{\mathbf{z}_{i*}=L}]$.
With these variables in place, we can derive the Dirichlet posterior for finite level indicators:
\begin{align}
   \mathbbm{P}(\mathbf{a}_i \ | \ \mathbf{A}_{-i}, & \mathbf{Z}_{i *}, \pmb{\alpha})  \nonumber \\
  & =\dfrac{\text{Multinomial} \big(  \mathbf{a}_i \big)\text{Dirichlet}(\pmb{\alpha})}{\bigintssss_{\mathbf{a}_{i}} \text{Multinomial} \big(\mathbf{a}_i \big) \text{Dirichlet}(\pmb{\alpha}) \ \text{d}\mathbf{a}_i} \nonumber \\
    & \overset{\mathrm{(1)}}{=} \dfrac{\Bigg( \dfrac{\Gamma(\sum_{l=1}^L \#^{\mathbf{z}_{i*}=l} + 1)}{\prod_{l=1}^L \Gamma(\#^{\mathbf{z}_{i*}=l} + 1)} \prod_{l=1}^L (a_i^l)^{\#^{\mathbf{z}_{i*}=l}} \Bigg) \Bigg( \dfrac{\Gamma(\sum_{l=1}^L \alpha_l)}{\prod_{l=1}^L\Gamma(\alpha_l)} \prod_{l=1}^L(a_i^l)^{\alpha_l - 1} \Bigg)}{\bigint_{\mathbf{a}_i} \Bigg( \dfrac{\Gamma(\sum_{l=1}^L \#^{\mathbf{z}_{i*}=l} + 1)}{\prod_{l=1}^L \Gamma(\#^{\mathbf{z}_{i*}=l} + 1)} \prod_{l=1}^L (a_i^l)^{\#^{\mathbf{z}_{i*}=l}} \Bigg) \Bigg( \dfrac{\Gamma(\sum_{l=1}^L \alpha_l)}{\prod_{l=1}^L\Gamma(\alpha_l)} \prod_{l=1}^L(a_i^l)^{\alpha_l - 1} \Bigg) \ \text{d}\mathbf{a}_i} \nonumber \\
    & \overset{\mathrm{(2)}}{=} \dfrac{\Bigg( \dfrac{\Gamma(\sum_{l=1}^L \#^{\mathbf{z}_{i*}=l} + 1)}{\prod_{l=1}^L \Gamma(\#^{\mathbf{z}_{i*}=l} + 1)}  \dfrac{\prod_{l=1}^L\Gamma(\alpha_l)}{\Gamma(\sum_{l=1}^L \alpha_l)} \Bigg)\Bigg( \dfrac{\prod_{l=1}^L\Gamma(\alpha'_l)}{\Gamma(\sum_{l=1}^L \alpha'_l)}  \dfrac{\Gamma(\sum_{l=1}^L \alpha'_l)}{\prod_{l=1}^L \Gamma(\alpha'_l)} \Bigg) \prod_{l=1}^L(a_i^l)^{\alpha'_l - 1}}{\bigint_{\mathbf{a}_i} \Bigg( \dfrac{\Gamma(\sum_{l=1}^L \#^{\mathbf{z}_{i*}=l} + 1)}{\prod_{l=1}^L \Gamma(\#^{\mathbf{z}_{i*}=l} + 1)}  \dfrac{\prod_{l=1}^L\Gamma(\alpha_l)}{\Gamma(\sum_{l=1}^L \alpha_l)} \Bigg)\Bigg( \dfrac{\prod_{l=1}^L\Gamma(\alpha'_l)}{\Gamma(\sum_{l=1}^L \alpha'_l)}  \dfrac{\Gamma(\sum_{l=1}^L \alpha'_l)}{\prod_{l=1}^L \Gamma(\alpha'_l)} \Bigg) \prod_{l=1}^L(a_i^l)^{\alpha'_l - 1} \ \text{d}\mathbf{a}_i} \nonumber\\
        & \overset{\mathrm{(3)}}{=} \dfrac{\Bigg( \dfrac{\Gamma(\sum_{l=1}^L \#^{\mathbf{z}_{i*}=l} + 1)}{\prod_{l=1}^L \Gamma(\#^{\mathbf{z}_{i*}=l} + 1)}  \dfrac{\prod_{l=1}^L\Gamma(\alpha_l)}{\Gamma(\sum_{l=1}^L \alpha_l)}  \dfrac{\prod_{l=1}^L\Gamma(\alpha'_l)}{\Gamma(\sum_{l=1}^L \alpha'_l)} \Bigg) \text{Dirichlet}(\pmb{\alpha'}) }{\bigint_{\mathbf{a}_i}\Bigg( \dfrac{\Gamma(\sum_{l=1}^L \#^{\mathbf{z}_{i*}=l} + 1)}{\prod_{l=1}^L \Gamma(\#^{\mathbf{z}_{i*}=l} + 1)}  \dfrac{\prod_{l=1}^L\Gamma(\alpha_l)}{\Gamma(\sum_{l=1}^L \alpha_l)}  \dfrac{\prod_{l=1}^L\Gamma(\alpha'_l)}{\Gamma(\sum_{l=1}^L \alpha'_l)} \Bigg) \text{Dirichlet}(\pmb{\alpha'}) \ \text{d}\mathbf{a}_i } \nonumber \\
       & \overset{\mathrm{(4)}}{=} \dfrac{\Bigg( \dfrac{\Gamma(\sum_{l=1}^L \#^{\mathbf{z}_{i*}=l} + 1)}{\prod_{l=1}^L \Gamma(\#^{\mathbf{z}_{i*}=l} + 1)}  \dfrac{\prod_{l=1}^L\Gamma(\alpha_l)}{\Gamma(\sum_{l=1}^L \alpha_l)}  \dfrac{\prod_{l=1}^L\Gamma(\alpha'_l)}{\Gamma(\sum_{l=1}^L \alpha'_l)} \Bigg) \text{Dirichlet}(\pmb{\alpha'}) }{\Bigg( \dfrac{\Gamma(\sum_{l=1}^L \#^{\mathbf{z}_{i*}=l} + 1)}{\prod_{l=1}^L \Gamma(\#^{\mathbf{z}_{i*}=l} + 1)}  \dfrac{\prod_{l=1}^L\Gamma(\alpha_l)}{\Gamma(\sum_{l=1}^L \alpha_l)}  \dfrac{\prod_{l=1}^L\Gamma(\alpha'_l)}{\Gamma(\sum_{l=1}^L \alpha'_l)} \Bigg) \bigint_{\mathbf{a}_i} \text{Dirichlet}(\pmb{\alpha'}) \ \text{d}\mathbf{a}_i } \nonumber \\
       & \overset{\mathrm{(5)}}{=} \text{Dirichlet}(\pmb{\alpha'})
\end{align}
Where (1) is obtained by applying the definitions of the Multinomial and Dirichlet distributions as per Equations \ref{eqn:multinomial} and \ref{eqn:dirichlet}, respectively; (2) leverages the definition of $\pmb{\alpha}'$ to group level memberships and introduces cancelling numerator and denominator terms using $\pmb{\alpha}'$ to obtain a Dirichlet probability density function as shown by replacement in (3); (4) groups terms constant with respect to $\mathbf{a}_i$ in integration; and (5) leverages the law of total probability.

\end{document}